\documentclass[10pt,journal,compsoc]{IEEEtran}

\usepackage{amsmath,amsfonts,bm}

\def\eqref#1{equation~\ref{#1}}

\def\1{\bm{1}}

\DeclareMathAlphabet{\mathsfit}{\encodingdefault}{\sfdefault}{m}{sl}
\SetMathAlphabet{\mathsfit}{bold}{\encodingdefault}{\sfdefault}{bx}{n}

\usepackage{adjustbox}
\usepackage{amssymb}
\usepackage{algorithm,algorithmicx}
\usepackage{algpseudocode}
\usepackage{multirow}
\usepackage{times}
\usepackage{soul}
\usepackage{ragged2e}
\usepackage{url}
\usepackage{hyperref}
\usepackage{booktabs}
\usepackage{graphicx}
\usepackage{subfigure}
\usepackage{bm}
\usepackage{xcolor}
\usepackage{amsmath}
\usepackage{amsfonts}
\usepackage{stmaryrd}
\usepackage{stfloats}
\usepackage{xspace}
\usepackage{threeparttable}
\usepackage{wrapfig}
\usepackage{centernot}
\usepackage{mathtools}
\usepackage{latexsym}
\usepackage{epsfig}
\usepackage{pifont}
\usepackage{enumitem}
\usepackage[normalem]{ulem}
\usepackage{ulem}
\usepackage{titlesec}
\titleformat{\paragraph}
{\normalfont\normalsize\itshape}{\theparagraph}{1em}{}
\titlespacing*{\paragraph}
{0pt}{2.25ex plus 1ex minus .2ex}{0.5ex plus .1ex}

\renewcommand{\raggedright}{\leftskip=0pt \rightskip=0pt plus 0cm}

\newcommand{\ie}{\emph{i.e.,}\xspace}
\newcommand{\eg}{\emph{e.g.,}\xspace}
\newcommand{\aka}{\emph{a.k.a.,}\xspace}
\newcommand{\etal}{\emph{et al.}\xspace}

\ifCLASSOPTIONcompsoc
\usepackage{cite}
\else
  \usepackage{cite}  
\fi

\ifCLASSINFOpdf
\else
\fi
\begin{document}

\title{Temporal Sentence Grounding in Videos: A Survey and Future Directions}

\author{Hao Zhang, Aixin Sun, Wei Jing, and Joey Tianyi Zhou
\IEEEcompsocitemizethanks{
\IEEEcompsocthanksitem H.~Zhang is with School of Computer Science and Engineering, Nanyang Technological University, Singapore, 639798. 
\IEEEcompsocthanksitem A.~Sun is with S-Lab, Nanyang Technological University, Singapore, 639798. 
\IEEEcompsocthanksitem W.~Jing is with Alibaba Group, China, 311121.
\IEEEcompsocthanksitem J.T.~Zhou is with Centre for Frontier AI Research, A*STAR, Singapore, 138632.
\IEEEcompsocthanksitem Corresponding author: A.~Sun (Email: axsun@ntu.edu.sg).
}
}

\markboth{Accepted by IEEE TRANSACTIONS ON PATTERN ANALYSIS AND MACHINE INTELLIGENCE}%
{Zhang \MakeLowercase{\textit{et al.}}: Temporal Sentence Grounding in Videos: A Survey and Future Directions}

\IEEEtitleabstractindextext{
\begin{abstract}
\raggedright{Temporal sentence grounding in videos (TSGV), \aka natural language video localization (NLVL) or video moment retrieval (VMR), aims to retrieve a temporal moment that semantically corresponds to a language query from an untrimmed video. Connecting computer vision and natural language, TSGV has drawn significant attention from researchers in both communities. This survey attempts to provide a summary of fundamental concepts in TSGV and current research status, as well as future research directions. As the background, we present a common structure of functional components in TSGV, in a tutorial style: from feature extraction from raw video and language query, to answer prediction of the target moment.  Then we review the techniques for multimodal understanding and interaction, which is the key focus of TSGV for effective alignment between the two modalities. We construct a taxonomy of TSGV techniques and elaborate the methods in different categories with their strengths and weaknesses. Lastly, we discuss issues with the current TSGV research and share our insights about  promising research directions.
}
\end{abstract}

}

\maketitle

\IEEEdisplaynontitleabstractindextext
\IEEEpeerreviewmaketitle

\IEEEraisesectionheading{\section{Introduction}\label{sec:intro}}

\IEEEPARstart{V}{ideo} has gradually become a major type of information transmission media, thanks to the fast development and innovation in communication and media creation technologies. A video is formed from a sequence of continuous image frames possibly accompanied by audio and subtitles. Compared to image and text, video conveys richer semantic knowledge, as well as more diverse and complex activities. Despite the strengths of video, searching for content from the video is challenging. Thus, there is a high demand for techniques that could quickly retrieve video segments of user interest, specified in natural language.

\subsection{Definition and History}
\label{ssec:definition}
\begin{figure}[t]
    \centering
    \includegraphics[trim={0cm 0cm 0cm 0cm},clip,width=0.9\linewidth]{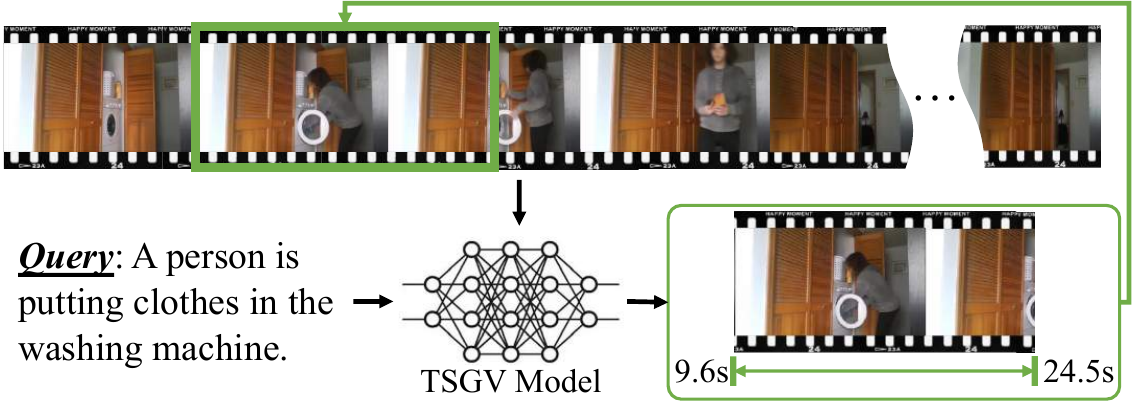}
    \caption{An illustration of temporal sentence grounding in videos (TSGV).}
	\label{fig:example}
\end{figure}

Given an untrimmed video, temporal sentence grounding in videos (TSGV) is to retrieve a video segment, also known as a temporal moment, that semantically corresponds to a query in natural language \ie sentence. As illustrated in Fig.~\ref{fig:example}, for the query ``\textit{A person is putting clothes in the washing machine.}'', TSGV needs to return the start and end timestamps (\ie $9.6s$ and $24.5s$) of a video moment from the input video as the answer. The answer moment should contain actions or events described by the query. 

As a fundamental vision-language problem, TSGV also serves as an intermediate step for various downstream vision-language tasks, such as video question answering and video-grounded dialogue\footnote{For detailed relations between TSGV and other vision-language tasks, please refer to Appendix Section~\ref{appd:ssec:tsgv_vs_others}.}. These tasks require localizing relevant moments about questions, then discovering or generating answers to the input questions by analyzing the retrieved moments. Naturally, TSGV connects computer vision (CV) and natural language processing (NLP) and benefits from the advancements made in both areas. 

TSGV also shares similarities with some classical tasks in both CV and NLP. For instance, video action recognition (VAR) ~\cite{tran2015learning,feichtenhofer2016convolutional,carreira2017quo,xu2020g} in CV is to detect video segments, which perform specific actions in video. Although VAR localizes temporal segments with activity information, it is constrained by the predefined action categories. TSGV is more flexible and aims to retrieve complicated and diverse activities from video via arbitrary language queries. In this sense, TSGV needs a semantic understanding of both video and language, as well as the multimodal interaction between them. TSGV is similar to the reading comprehension (RC) task in NLP~\cite{wang2017gated,seo2017bidaf,wei2018fast,huang2018fusionnet}, which is to retrieve a span of words from the text to answer a question. The core of RC is the interaction between text passages and query. TSGV models the interaction between two different modalities, making it more arduous and challenging.

\begin{figure}[t]
    \centering
    \includegraphics[trim={0cm 0cm 0cm 0cm},clip,width=0.9\linewidth]{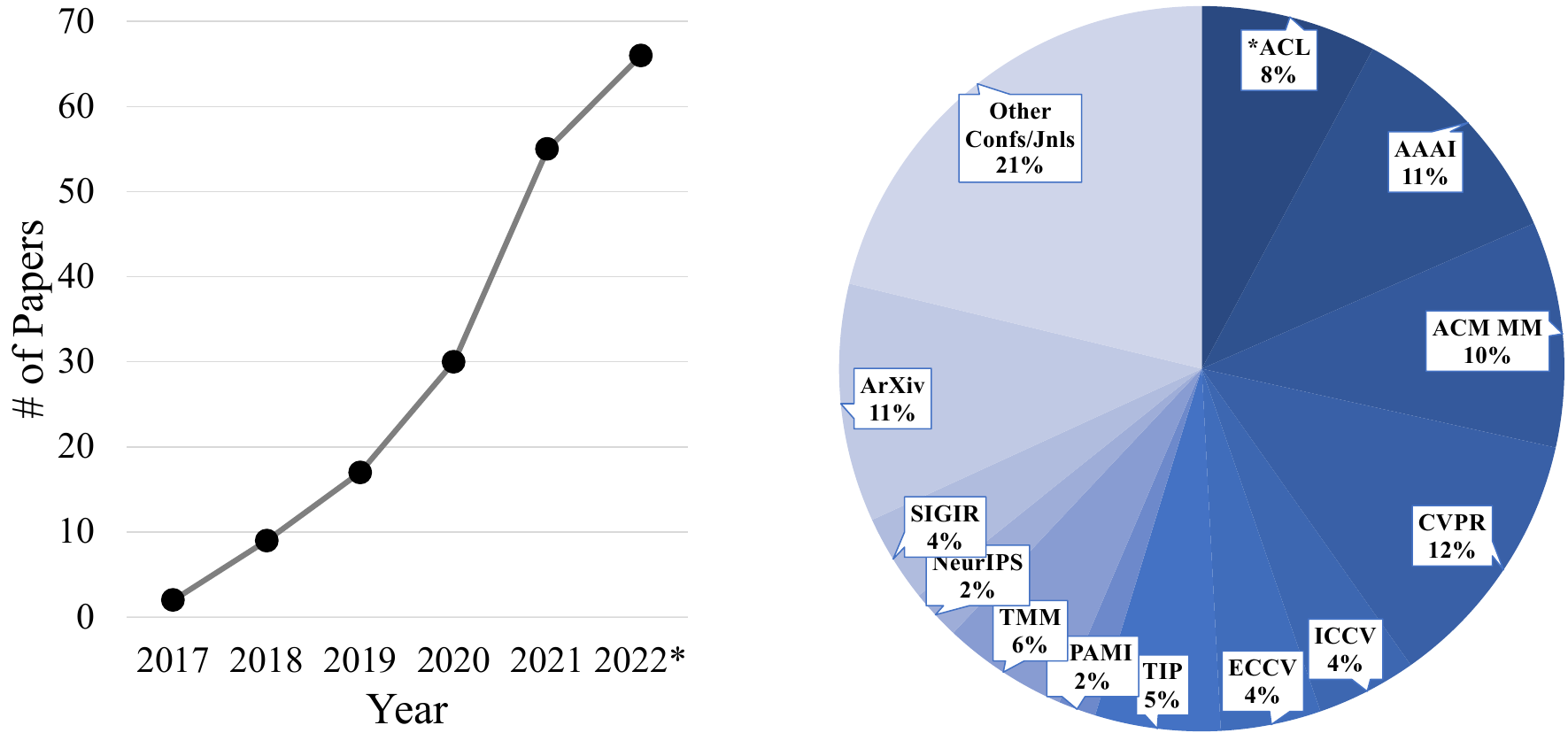}
    \caption{Statistics of the collected papers in this survey. Left:  number of papers published each year (till September 2022). Right: distribution of papers by venue, where *ACL denotes the series of conferences hosted by the Association for Computational Linguistics.}
	\label{fig:paper_stat}
\end{figure}

TSGV was proposed in 2017~\cite{Gao2017TALLTA,hendricks17iccv}; the task immediately drew significant attention from researchers. Early solutions mainly adopt an ineffective two-stage approach, first to sample moments as candidate answers, then to score these candidates~\cite{Gao2017TALLTA,hendricks17iccv,Liu2018CML,Liu2018AMR,ge2019mac}. Subsequent solutions focus more on effective and efficient multimodal interactions between video and query. A lot of methods are developed, including proposal-based~\cite{Xu2018TexttoClipVR,chen2018temporally,zhang2019man,yuan2019semantic,zhang2020learning}, proposal-free~\cite{yuan2019to,ghosh2019excl,chen2019localizing,lu2019debug,zhang2020vslnet}, reinforcement learning-based~\cite{he2019read,wang2019language,wu2020tree}, and weakly-supervised~\cite{mithun2019weakly,gao2019wslln,duan2018weakly,lin2020weakly,chen2021towards_cvpr} methods, etc. 

In this survey, we aim to provide a comprehensive and systematic review of TSGV research. We collect papers from reputable conferences and journals in CV, NLP, MM, IR, and machine learning areas, \eg CVPR, ECCV, ICCV, WACV, BMVC, ACL, EMNLP, NAACL, SIGIR, ACM MM, NeurIPS, AAAI, IJCAI, and TPAMI, TMM, TIP, etc. The papers were mainly published from 2017 to 2022\footnote{The paper collection was conducted lastly on 2022-09-18.}. For the paper collection, we primarily rely on academic search engines and digital libraries, such as IEEE Xplore, ACM Digital Library, ScienceDirect, Springer, ACL Anthology, CVF Open Access, etc. We also adopt Google Scholar to collect papers in other conferences/journals, and open-sourced articles.\footnote{A number of keywords and their combinations are utilized for paper searching, including moment, grounding, localization, language query, video retrieval, moment retrieval, video grounding, temporal grounding, moment localization, video localization, temporal localization, temporal language grounding, temporal sentence grounding, etc.} Fig.~\ref{fig:paper_stat} summarizes the statistics of the collected papers.

\subsection{The User's Dilemma and the Role of Expertise}
\label{ssec:related_survey}
The availability of a vast collection of TSGV methods easily confounds a researcher or practitioner attempting to select or design an algorithm suitable for a specific problem at hand. Existing surveys~\cite{yang2020asurvey,liu2021survey,Lan2021ASO} summarize the progress of TSGV research and establish taxonomies of methods based on their task formulation and architecture. Being the first survey, the taxonomy presented in Yang \etal~\cite{yang2020asurvey} is relatively incomplete and coarse. Liu \etal~\cite{liu2021survey} propose a pipeline of the TSGV model by partitioning it into three components and categorizing the existing solutions into supervised and weakly-supervised groups. However, their taxonomy is unable to cover various TSGV approaches as well. Lan \etal~\cite{Lan2021ASO} present a more complete taxonomy, with detailed illustrations and comparison between different categories of methods. Benchmark datasets and evaluation metrics are also covered. The most recent survey by Liu \etal~\cite{liu2022vmlsurvey} covers more TSGV methods and provides an efficiency comparison among methods. Similar to prior work, this survey lists current research but does not provide an in-depth critical analysis of methods and insights into future directions.

Our survey covers more recent developments in TSGV research. By abstracting common generalities in all methods, we summarize different types of TSGV methodologies and reveal a common pipeline of the TSGV model. We also establish a more comprehensive taxonomy and conclude more concrete and promising future research directions. All existing surveys focus on summarizing existing TSGV methods and stating future research directions. However, they do not provide a critical analysis of existing TSGV methods. More importantly, common questions from researchers/practitioners are not well addressed in existing surveys: (i) How should TSGV data be processed? (ii) How should the data be used in a particular TSGV method? (iii) What does a TSGV method generally look like and how it works? and (iv) Which model assessment is appropriate for a particular TSGV method? Our aim is to provide these perspectives on the composition of TSGV methods and state-of-the-art TSGV research. With such perspectives, an informed practitioner is able to confidently assess the trade-offs of various TSGV methods and make a competent decision on designing a TSGV solution with a suite of techniques. 

This survey is organized as follows. In Section~\ref{sec:background}, we present a general pipeline of TSGV methods and interpret the technical details in a tutorial style. It provides readers with background on what a TSGV model generally looks like, its I/O, and functional components. Section~\ref{sec:dataset_measure} summarizes the major benchmark datasets and  evaluation metrics. Section~\ref{sec:method_overview} classifies TSGV solutions into  categories, elaborates on the methods in each category, and discusses their pros and cons. Section~\ref{sec:evaluation} summarizes the current research progress. 
Section~\ref{sec:challenges_directions} discusses open issues and further research directions. Section~\ref{sec:conclusion} concludes this paper.

\section{Background}
\label{sec:background}
\begin{figure*}[!t]
    \centering
    \begin{minipage}{0.6\linewidth}
        \centering
        \includegraphics[trim={0cm 0cm 0cm 0cm},clip,width=0.95\linewidth]{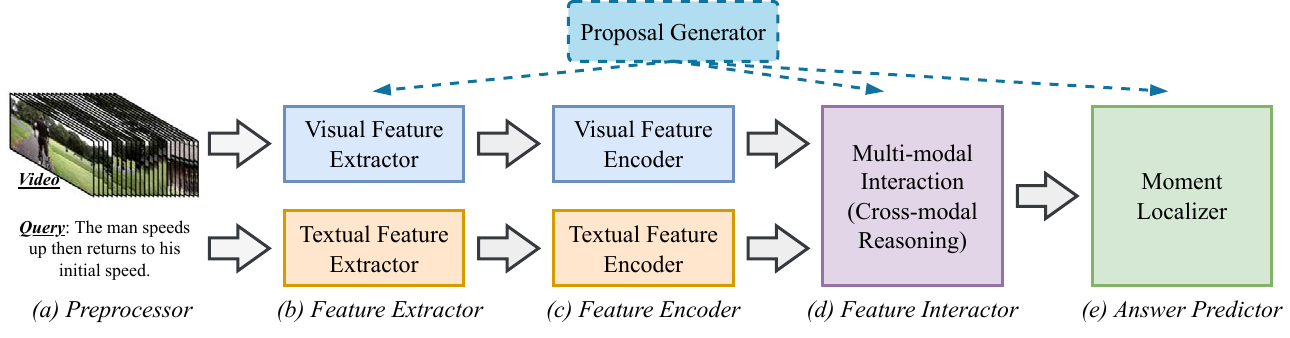}
        \makeatletter\def\@captype{figure}\makeatother\caption{A general pipeline for temporal sentence grounding in videos.}
        \label{fig:tsgv_pipeline}
    \end{minipage}
    \hspace*{\fill}
   \begin{minipage}{0.38\linewidth}
       \centering
       \includegraphics[width=0.9\textwidth]{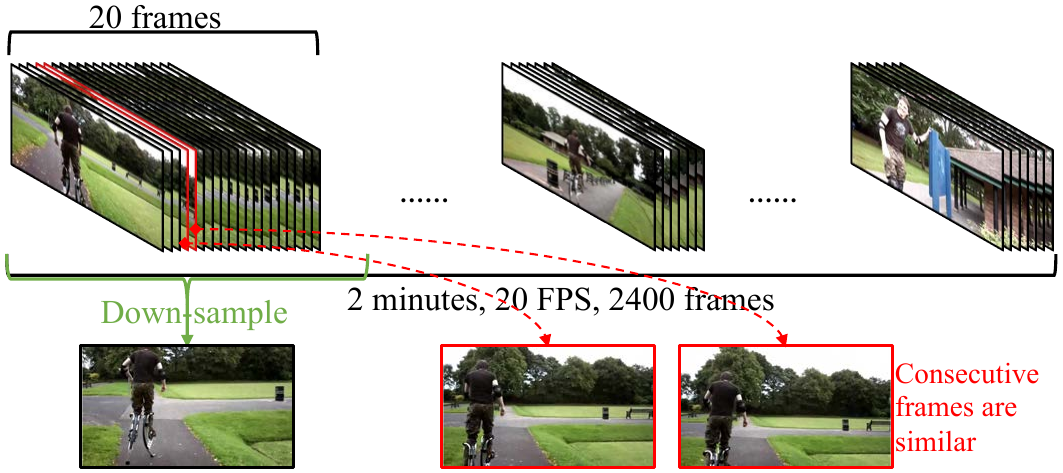}
       \makeatletter\def\@captype{figure}\makeatother\caption{An example of video frames down-sampling.}
       \label{fig:video_inputs}
    \end{minipage}
\end{figure*}

There are no theoretical guidelines that reveal a common structure or pipeline of a TSGV method. Despite various sophisticated architectures in different methods, conceptually, a TSGV method generally contains  six components shown in Fig.~\ref{fig:tsgv_pipeline}. The dotted line in the figure indicates that the proposal generator is an optional component, and it may be placed at different stages. We brief these main components to provide the necessary background to the readers before we zoom into the technical details in Section~\ref{sec:method_overview}.

A TSGV method takes a video-query pair as input, where the video is a collection of consecutive image frames, and the query is a sequence of words. The preprocessor prepares inputs for feature extraction, \eg  downsampling and resizing image frames in the video, and tokenizing words in the query sentence. Feature extractor converts the video frames and query words into their corresponding vector feature representations. Then the encoder module maps the video and query features to the same dimension and aggregates contextual information to enhance the feature representation. The interactor module, an essential component in TSGV, learns the multimodal representations by modeling the cross-modal interaction between video and query. Finally, the answer predictor generates moment predictions based on the learned multimodal representations. For proposal-based methods, the answer predictor makes predictions based on the proposals generated by the proposal generator. A proposal can be considered as a candidate answer moment, which can be generated at different stages. An example proposal is a video segment sampled from the input video. Proposal-free methods predict answers directly without the need of generating candidate answers.  

Before we elaborate the details of each component, we define the following notations. Given a TSGV dataset, we denote its video corpus as $\mathcal{V}=\{V^1,V^2,\dots,V^N\}$ and its query set as $\mathcal{Q}=\{Q^1,Q^2,\dots,Q^M\}$, where $N$ and $M$ are the number of videos and queries, respectively. Note that multiple queries can be posed on the same video with its different moments as answers; typically $M\geq N$ in TSGV datasets. Given a video-query pair, a video $V$ contains $T$ frames $V=[f_1,f_2,\dots,f_T]$ and a query $Q$ has $m$ words $Q=[q_1,q_2,\dots,q_m]$, the start and end time of the  ground truth moment are denoted by $\tau_s$ and $\tau_e$, $1\leq\tau_s<\tau_e\leq T$. Here, we use the frame index to represent time points, based on a fixed frame rate or fps. Mathematically, TSGV is to retrieve the target moment starting from $\tau_s$ and ending at $\tau_e$ by giving a video $V$ and query $Q$, \ie $\mathcal{F}_{TSGV}:(V,Q)\mapsto (\tau_s,\tau_e)$.

\subsection{Preprocessor}
\label{ssec:tsgv_inputs}
Video is a series of still images and the number of frames can be very large. For instance, a 2-minute video with $20$ fps has $2,400$ frames in total. Thus, it is infeasible (and often unnecessary) to process every frame in a video due to computational cost. Besides, video is continuous, \ie changes between consecutive frames are usually small and smooth. Hence, it is reasonable to downsample video for efficient computation. As shown in Fig.~\ref{fig:video_inputs}, if we sample $1$ frame from every $20$ consecutive frames, we only need to process $120$ frames instead of $2400$ frames for this $2$-minute video. With a downsample rate $r_{ds}$, the number of video frames becomes $T'=T/r_{ds}$. Downsample rate has a direct impact on video quality and should be carefully selected depending on the dataset.

Language query is discrete and words in a sentence demonstrate syntactic structure. Different word combinations lead to very different semantic meanings. For instance, in a query sentence ``\textit{The man speeds up then returns to his initial speed.}'', the words ``\textit{initial}'' and ``\textit{speed}'' carry different meanings, and their combination describes a specific scene. For preprocessing, a query is typically tokenized into word tokens. If a query contains too many words, a common strategy is truncation, \ie taking a fixed number of words from the beginning and discarding the rest. 

\subsection{Feature Extractor}
\label{ssec:tsgv_feat_extractor}
The feature extractor bridges the raw inputs and the model by converting inputs into feature representations.

\smallskip\noindent\textbf{Textual Feature Extractor} maps a query sentence to an embedding space, which can be categorized into token-level and sentence-level extractors.
Token-level extractor converts each word into its corresponding word embedding by using pre-trained word embeddings (PWE), \eg Word2Vec~\cite{mikolov2013efficient} and GloVe~\cite{pennington2014glove}, or pre-trained language models (PLM), \eg BERT~\cite{devlin2019bert} and RoBERTa~\cite{liu2019roberta}. We represent token-level extraction as:
\begin{equation}
    Q=[q_1,\dots,q_m] \xmapsto{\text{PWE/PLM}} \mathbf{Q}=[\mathbf{q}_1, \dots,\mathbf{q}_m]\in\mathbb{R}^{m\times d_q},
\end{equation}
where $d_q$ denotes the word embedding dimension. 

Sentence-level extractor encodes the entire query into a sentence feature in  $d_s$ dimension, by using pre-trained sentence encoder (PSE), \eg Skip-Thought~\cite{kiros2015skip}, InferSent~\cite{conneau2017supervised},  Sentence-BERT~\cite{reimers2019sentence}, or PWE/PLM with a trainable sentence encoder (TSE). We represent the process as:
\begin{equation}
\begin{aligned}
    Q & =[q_1,\dots,q_m] \xmapsto{\text{PSE}} \mathbf{q}_s\in\mathbb{R}^{d_s}, \text{ or } \\
    Q & =[q_1,\dots,q_m] \xmapsto{\text{PWE/PLM}} \mathbf{Q}\in\mathbb{R}^{m\times d_q} \xmapsto{\text{TSE}} \mathbf{q}_s\in\mathbb{R}^{d_s}
\end{aligned}
\end{equation}

\smallskip\noindent\textbf{Visual Feature Extractor} converts video frames to a sequence of visual features. Depending on whether proposals are generated directly on the input video, there are two types of feature extraction. 

Recall that a proposal is a candidate answer. A straightforward approach is to sample video segments from the input video as proposals. Proposals may contain a different number of frames. Suppose there are $n_{\text{seg}}$ video segments as proposals, the feature extraction process is described as:
\begin{equation}
\begin{aligned}
    & V\in\mathbb{R}^{T'\times frame} \xmapsto{\text{proposals}} \{\text{segment}_i\in\mathbb{R}^{\chi\times frame}\}_{i=1}^{n_{\text{seg}}} \\
    & \xmapsto[\text{extractor}]{\text{visual feature}} \mathbf{V}=\{\mathbf{v}_{p,i}\in\mathbb{R}^{d_v}\}_{i=1}^{n_{\text{seg}}},
\end{aligned}
\end{equation}
where $\chi$ is the number of frames in a proposal, and $d_v$ denotes the dimension of extracted features. The task becomes to determine whether a proposal represented by $\mathbf{v}_{p,i}$ is the correct answer. 

If proposals are not generated directly from the input video, then the video is uniformly decomposed into a sequence of non-overlapping snippets. Suppose there are $n_{\text{snp}}$ video snippets and each snippet contains $\xi$ frames, the extraction process is:
\begin{equation}
\begin{aligned}
    & V\in\mathbb{R}^{T'\times frame} \xmapsto{\text{decompose}} [\text{snippet}_i]_{i=1}^{n_{\text{snp}}}\in\mathbb{R}^{n_{\text{snp}}\times\xi\times frame} \\
    & \xmapsto[\text{extractor}]{\text{visual feature}} \mathbf{V}=[\mathbf{v}_i]_{i=1}^{n_{\text{snp}}}\in\mathbb{R}^{n_{\text{snp}}\times d_v}.
\end{aligned}
\end{equation}

Here we distinguish ``video snippet'' from ``video segment''. A video segment is sampled as a proposal to match the target moment, and a video snippet is a very short clip that only contains a few frames, \ie $\xi<<\chi$ in general. Furthermore, as each video segment is one candidate answer, the video segments are irrelevant to each other and they are further processed separately in TSGV. As very short clips, video snippets are maintained in sequence, and are jointly processed in later stages. 

Each frame is a still image. 
From frames to features, the commonly used pre-trained visual feature extractors are (i) 3D-ConvNet for action recognition, \eg C3D~\cite{tran2015learning} or I3D~\cite{carreira2017quo}, and (ii) 2D-ConvNet for object detection, \eg VGG~\cite{simonyan2014very} or ResNet~\cite{he2016deep}.

\begin{figure*}[t]
    \centering
    \subfigure[Sliding window (SW) strategy]
    {
        \label{fig:proposal_sw}    
        \includegraphics[trim={0.2cm 0.2cm 0.2cm 0cm},clip,width=0.22\textwidth]{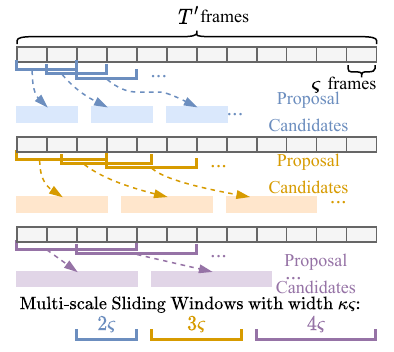}
    }
    \hfill
    \subfigure[Proposal generated (PG) strategy]
    {
        \label{fig:proposal_pg}  
       \includegraphics[trim={0cm 0.3cm 0.1cm 0.2cm},clip,width=0.22\textwidth]{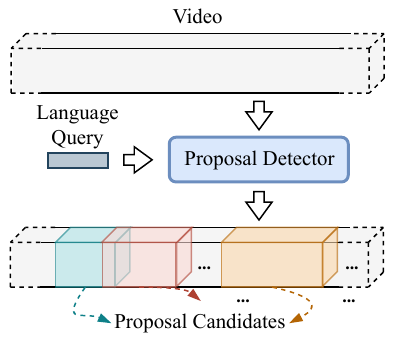}
    }
    \hfill
     \subfigure[Anchor-based strategy]
    {
        \label{fig:proposal_anchor}  
      \includegraphics[trim={0.2cm 0.4cm 0.2cm 0.1cm},clip,width=0.22\textwidth]{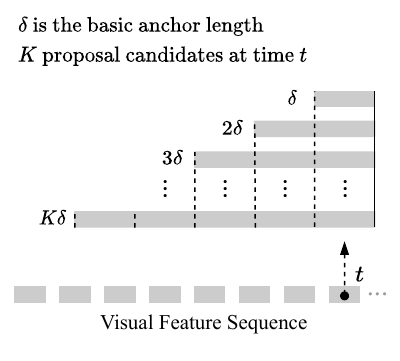}
    }
    \hfill
     \subfigure[2D-Map strategy]
    {
        \label{fig:proposal_2d}  
       \includegraphics[trim={0cm 0.4cm 0.1cm 0cm},clip,width=0.22\textwidth]{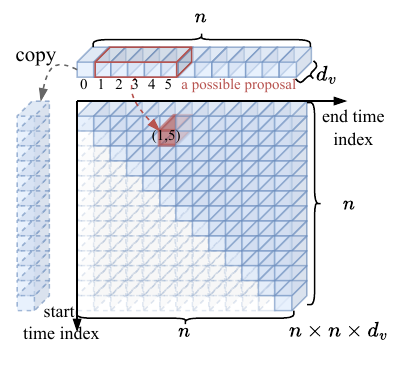}
    }
    \caption{Illustration of sliding window, proposal generated, anchor-based, and 2D-Map strategies.}
    \label{fig:fourStrategies}
\end{figure*}

\subsection{Feature Encoder and Feature Interactor}
\label{ssec:tsgv_feat_encoder}
Feature encoder maps visual and textual features to the same dimension, and refines their feature representations by encoding their corresponding contextual information. Existing TSGV methods use various feature encoders, from simple multi-layer perceptrons to complex transformers and graph neural networks. The design of the feature encoder highly depends on the model architecture. 

Briefed in Section~\ref{ssec:tsgv_inputs}, there are token-level and sentence-level query features. There are also two types of visual features, depending on whether a proposal generator is applied on input video, \ie proposal feature and video snippet feature sequence. Let $d$ be the target dimension for both visual and textual features. Mapping of sentence-level and token-level query features is defined as:
\begin{equation}
\begin{aligned}
    \mathbf{q}_s\in\mathbb{R}^{d_s} & \xmapsto[\text{encoder}]{\text{textual feature}} \mathbf{q}'_s\in\mathbb{R}^{d}, \text{ and } \\ \mathbf{Q}\in\mathbb{R}^{m\times d_q} & \xmapsto[\text{encoder}]{\text{textual feature}} \mathbf{Q'}\in\mathbb{R}^{m\times d}.
\end{aligned}
\end{equation}
For the proposal feature and video snippet feature sequence, the mapping is written as:
\begin{equation}
\begin{aligned}
    \mathbf{v}_p\in\mathbb{R}^{d_v} & \xmapsto[\text{encoder}]{\text{visual feature}} \mathbf{v}'_p\in\mathbb{R}^{d}, \text{ and } \\ \mathbf{V}\in\mathbb{R}^{n\times d_v} & \xmapsto[\text{encoder}]{\text{visual feature}} \mathbf{V'}\in\mathbb{R}^{n\times d},
\end{aligned}
\end{equation}
where we simply use $\mathbf{v}_p\in\mathbb{R}^{d_v}$ to represent the visual feature of a proposal, and $n$ to replace $n_{snp}$.

\begin{figure}[t]
    \centering
    \includegraphics[trim={0cm 0cm 0cm 0cm},clip,width=\linewidth]{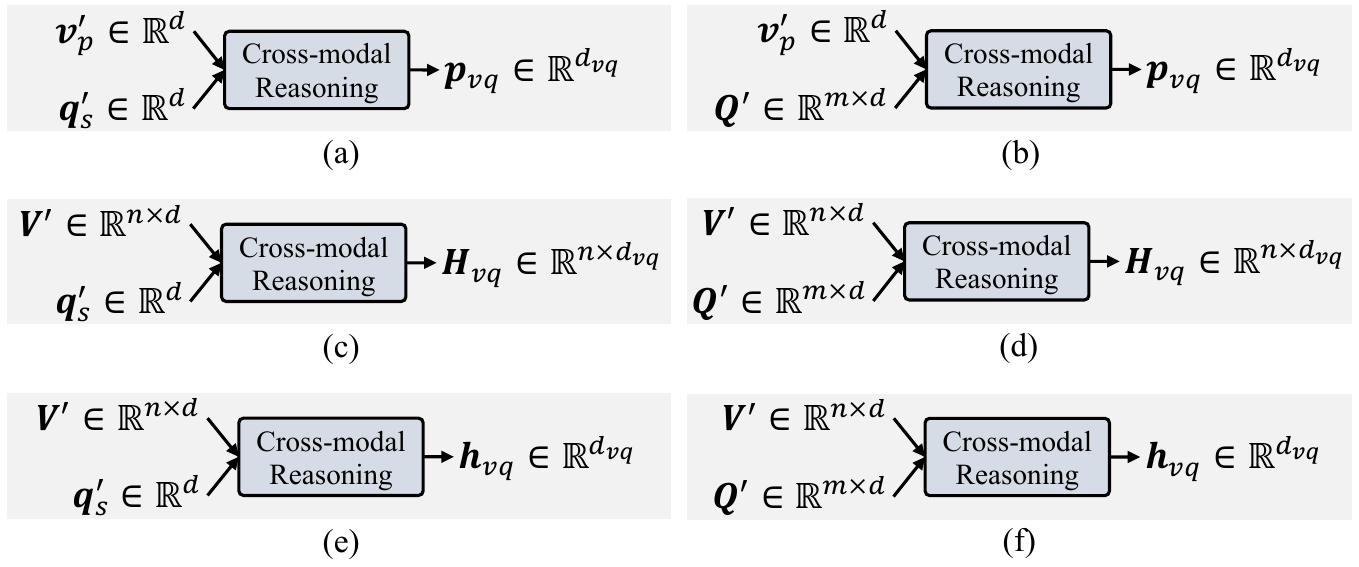}
    \caption{The common input/output feature formats of feature interactor in TSGV. $\mathbf{p}_{vq}\in\mathbb{R}^{d_{vq}}$ denotes the learned multimodal proposal feature; $\mathbf{H}_{vq}=[\mathbf{h}_{vq}^1,\dots,\mathbf{h}_{vq}^n]\in\mathbb{R}^{n\times d_{vq}}$ is the multimodal snippet feature sequence; $\mathbf{h}_{vq}\in\mathbb{R}^{d_{vq}}$ is the pooled multimodal snippet feature.  $d_{vq}$ denotes the dimension of output multimodal feature.}
	\label{fig:feat_interactor_io}
\end{figure}

Feature interactor, an essential component in any TSGV method, aims to learn cross-interaction between video and query. Recall the goal of TSGV is to retrieve a target moment from the video that \textit{semantically corresponds} to the query. Thus, the feature interactor requires to understand the semantic meaning of query and to recognize various activities in the video simultaneously. It then performs to fuse query and video representations by emphasizing the portion of the video content that is most relevant to the query semantics. In general, the quality of the feature interactor determines the performance of a TSGV method to a large extent. 

Fig.~\ref{fig:feat_interactor_io} summarizes the various input and output formats of different feature interactors among existing TSGV methods. The input is determined by the types of query features (token-level or sentence-level), and the types of visual features (proposal or snippet sequence). The common output feature formats include (i) the learned multimodal proposal feature $\mathbf{p}_{vq}\in\mathbb{R}^{d_{vq}}$, (ii) the multimodal snippet feature sequence  $\mathbf{H}_{vq}=[\mathbf{h}_{vq}^1,\dots,\mathbf{h}_{vq}^n]\in\mathbb{R}^{n\times d_{vq}}$, and (iii) the pooled multimodal snippet feature $\mathbf{h}_{vq}\in\mathbb{R}^{d_{vq}}$. Here,  $d_{vq}$ is the dimension of the multimodal feature.

Output format of the feature interactor is highly correlated with the answer predictor in a TSGV method. Answer predictor may depend on proposals that can be generated at different stages. We next brief the proposal generation before the answer predictor.

\subsection{Proposal Generation}
\label{ssec:tsgv_proposal}
Depending on whether a proposal generation module is used, existing TSGV methods can be roughly categorized into \textit{proposal-based} and \textit{proposal-free} methods.
As shown in Fig.~\ref{fig:tsgv_pipeline}, the proposal generator can be integrated into the model at various positions. For instance, proposals can be directly sampled from the input video. Proposals can also be generated before or after the feature interactor based on the visual features. Anchor-based methods generate proposals during answer prediction. A method may also engage multiple proposal generation strategies.

Sliding window-based (SW)  strategy~\cite{hendricks17iccv,Gao2017TALLTA,hendricks2018localizing,Liu2018AMR,Liu2018CML,wu2018multi,ge2019mac,zhang2019exploiting,jiang2019cross,ning2020asst,zeng2021multi,ning2021interaction} generates proposal candidates by densely sampling fixed-length video segments on the input video, using pre-defined multi-scale sliding windows. SW strategy is usually performed directly on video frames. Illustrated in Fig.~\ref{fig:proposal_sw}, given a downsampled video with $T'$ frames and a set of sliding windows, each sliding window samples video segments sequentially, with a preset overlap ratio ($r_o$). In our illustration, we use three different sliding windows $sw\in\{\kappa\varsigma\}_{\kappa=2,3,4}$ ($\varsigma$ is a basic window size) and set $r_o=0.5$. Overlap ratio is necessary to increase the chance of covering the target moment. Then we have a set of video segments as proposals.

Proposal-generated (PG) strategy~\cite{xu2017r,Xu2018TexttoClipVR,Xu2019MultilevelLA,chen2019semantic,xiao2021boundary,liu2021adaptive,xiao2021natural,hu2021video} produces proposals by utilizing auxiliary modules, \eg pre-trained segment proposal network (SPN)~\cite{xu2017r} or carefully designed proposal detector. The PG strategy is usually performed on visual features, but it involves the query as input to guide its proposal generation process, illustrated in  Fig.~\ref{fig:proposal_pg}. Hence, the proposals generated are related to the query. Depending on the position of the proposal detector, PG strategy may also involve feature encoder and interactor.

Anchor-based strategy~\cite{chen2018temporally,zhang2019man,zhu2019cross,yuan2019semantic,yuan2020semantic,lin2020moment,Wang2020TemporallyGL,qu2020fine,liu2020jointly,liu2020reasoning,ma2021hierarchical,zhang2021multimodal,liu2021progressively,wang2021dctnet} generates proposals with pre-set multi-scale anchors. Different from the SW strategy, it is performed on the encoded visual features and is integrated in the answer predictor. Suppose we have $K$ different scale anchors, and the length of a basic anchor is $\delta$. Fig.~\ref{fig:proposal_anchor} plots a commonly used anchor-based strategy. This strategy applies $K$ preset anchors to generate proposals, ended at a time step $t$, where $t$ is the index of the multimodal visual feature in the feature sequence.

Another version of anchor-based strategy is 2D-Map strategy~\cite{liu2018temporal,zhang2020learning,zhang2021ms2dtan,zheng2021progressive,wang2021structured,hu2021coarse,soldan2021vlg,gao2021relation,zhang2021multi,huang2020aligned,xu2020g,gao2021fast,wu2021diving,wang2021negative,jia2022stcmnet}. Different from the standard anchor-based strategy above, 2D-Map strategy is usually applied after the feature extractor, \ie before answer predictor. It generates proposals by modeling the temporal relations between video moments through a two-dimensional map. One dimension indicates the starting time of a moment; the other indicates the end time.  Given a visual feature sequence with $n\times d_v$, all possible proposal candidates are computed based on a 2D temporal feature map. Shown in Fig.~\ref{fig:proposal_2d}, a candidate proposal representation can be computed by max-pooling the corresponding visual features across a specific time span, resulting in the 2D feature map with $n\times n\times d_v$. Note the start ($a$) and end ($b$) timestamps of a proposal candidate should satisfy $a\leq b$. Therefore, only proposal candidates that locate in the upper triangular part of the 2D map are valid. 

\subsection{Answer Predictor and Objective}
\label{ssec:tsgv_ans_predictor}
Answer predictor is responsible for predicting the position of a target moment based on the learned multimodal features. Next, we brief the commonly used answer predictors and their corresponding objectives, for both proposal-based and proposal-free methods. Methods may combine multiple answer predictors or incorporate various auxiliary objectives to boost performance. In this background section, we only focus on the main objectives.

For proposal-based methods, the answer predictor computes a score for each proposal. Ideally, a proposal gets a higher score if it is closer to the ground truth moment. Specifically, given a multimodal proposal feature $\mathbf{p}_{vq}$, its score is computed as $s=\sigma(\mathcal{A}(\mathbf{p}_{vq}))\in\mathbb{R}^1$, where $\mathcal{A}$ is answer predictor and $\sigma$ is an (optional) activation function. Then, the proposal with the highest score is selected as the answer. If proposals are generated by anchor-based strategy, the score is computed based on the multimodal snippet feature sequence $\mathbf{H}_{vq}$ by applying multi-scale anchors in the answer predictor.  

Various learning objectives have been developed for proposal-based methods. The alignment loss~\cite{Gao2017TALLTA,hendricks2018localizing,Liu2018AMR,Liu2018CML,wu2018multi,ge2019mac,jiang2019cross,ning2020asst,zeng2021multi,ning2021interaction,chen2019semantic,liu2021adaptive} is commonly used for SW and PG strategies, which is defined as:
\begin{equation}
    \mathcal{L}_{aln} = \gamma\log(1+e^{-s_{i,i}}) + \sum_{j=0,j\neq i}^{N_{neg}}\log(1+e^{s_{i,j}}),
\label{eqn:aln_loss}
\end{equation}
where $s_{i,i}$ is the score of aligned (or positive) proposal-query pair, and $s_{i,j}$ is the score of misaligned (or negative) pair; $\gamma$ is a hyper-parameter to control the weight between positive and negative pairs; $N_{neg}$ is the number of negative pairs. For a given query, a proposal is considered positive if it has a good overlap with the ground truth moment, measured by IoU (intersection area over union area). Otherwise, it is negative. Nevertheless, a negative pair can also be constructed by replacing a random query or pairing random but unmatched proposals and queries. In general, $\mathcal{L}_{aln}$ encourages aligned proposal-query pairs to have positive scores and misaligned pairs to have negative scores. Besides, triple-based ranking loss~\cite{hendricks17iccv,hendricks2018localizing,liu2018temporal,shao2018find,zhang2019exploiting,Xu2018TexttoClipVR,Xu2019MultilevelLA,hu2021video} has also been used for SW and PG strategies:
\begin{equation}
    \mathcal{L}_{triple} = \max(0, \eta+s'-s)
\label{eqn:triple_loss}
\end{equation}
where $s$ denotes the score of matched proposal-query pair and $s'$ is the score of mismatched proposal-query pair. Similarly, $\mathcal{L}_{triple}$ encourages similarities between aligned pairs to be greater than misaligned pairs by some margin $\eta>0$.

For anchor-based and 2D-Map strategies, binary cross-entropy loss~\cite{liu2021adaptive,chen2018temporally,zhang2019man,yuan2019semantic,yuan2020semantic,zhu2019cross,lin2020moment,Wang2020TemporallyGL,qu2020fine,ma2021hierarchical,zhang2021multimodal,liu2020jointly,liu2020reasoning,liu2021progressively,wang2021dctnet,zhang2020learning,zhang2021ms2dtan,zheng2021progressive,wang2021structured,hu2021coarse,jia2022stcmnet,soldan2021vlg,gao2021relation,wu2021diving,zhang2021multi,wang2021negative,gao2021fast,liu2021context,wang2020dual,bao2021dense,ding2021support} is usually adopted, which is defined as:
\begin{equation}
    \mathcal{L}_{bce} = \gamma\cdot y\cdot\log s + (1-y)\cdot\log(1-s)
\label{eqn:bce_loss}
\end{equation}
where $\gamma$ is an optional balance weight, determined based on the number of positive and negative samples. $y$ is the corresponding anchor label for the proposal; $y=1$ if the proposal candidate has IoU with ground truth moment larger than a threshold $\theta$, \ie positive. Otherwise $y=0$.
$y$ may also be defined as the scaled IoU value between the proposal and the ground truth moment.

Proposal-free methods do not generate proposals. Instead, they use a regressor or a span predictor as the answer predictor. Specifically, regression-based predictor aims to regress the start and end times of the  target moment directly. It takes the pooled multimodal snippet feature $\mathbf{h}_{vq}$ as input and predicts the temporal positions $(t_s,t_e)$. Mathematically, we represent this process as $(t_s,t_e)=\sigma(\mathcal{A}(\mathbf{h}_{vq}))\in\mathbb{R}^{2}$, where $\mathcal{A}$ denotes the regressor, and $\sigma$ is (optional) Sigmoid activation to normalize the output to $[0,1]$. Given a predicted $(t_s,t_e)$ and the normalized ground truth $(\tau_s,\tau_e)$, the smoothed $L_1$ loss~\cite{liu2021adaptive,yuan2019semantic,yuan2020semantic,ma2021hierarchical,yuan2019to,zhang2020simple,zeng2020dense,mun2020local,li2021proposal,zhou2021embracing,chen2021end,cao2021pursuit}, MSE loss~\cite{xiao2021boundary,xiao2021natural,lu2019debug,chen2020rethinking,liu2021single,wang2020dual,xu2021boundary} or Huber loss~\cite{chen2020hierarchical,chen2020learning}, \ie $R\in\{\text{smooth}_{L_1}, \text{MSE}, \text{Huber}\}$, is commonly used as learning objectives:
\begin{equation}
    \mathcal{L}_{reg} = R(t_s-\tau_s) + R(t_e-\tau_e),
\label{eqn:reg_loss}
\end{equation}

Span predictor also predicts the start and end boundaries of the target moment directly. Different from repression-based predictor, span predictor computes the probability of each video snippet being the start and end points of the target moment. Specifically, it takes the multimodal snippet feature sequence $\mathbf{H}_{vq}$, and computes the start and end boundary scores as $(\mathbf{S}_s, \mathbf{S}_e)=\mathcal{A}(\mathbf{H}_{vq})\in\mathbb{R}^{n\times 2}$. Then, the probability distributions of boundaries are computed by $\mathbf{P}_s=\text{softmax}(\mathbf{S}_s)\in\mathbb{R}^{n}$ and $\mathbf{P}_e=\text{softmax}(\mathbf{S}_e)\in\mathbb{R}^{n}$, where $\mathbf{P}_{s/e}^t$ denotes the probability of $t$-th snippet be the start/end boundary. Cross-entropy loss~\cite{xiao2021boundary,chen2019localizing,ghosh2019excl,zhang2020vslnet,zhang2021natural,zhang2021parallel,nan2021interventional,yu2021cross,tang2021multi,tang2021frame,zhang2021temporal,qi2021collaborative,zhang2022natural,gou2021sneak,xu2021boundary} and Kullback-Leibler (KL) divergence~\cite{rodriguez2020proposal,zhao2021cascaded,liang2021local,rodriguez2021dori,xiao2021natural,wang2020dual} are both commonly used for span prediction. The cross-entropy objective is defined as:  
\begin{equation}
    \mathcal{L}_{span} = f_{XE}(\mathbf{P}_s, \mathbf{Y}_s) + f_{XE}(\mathbf{P}_e, \mathbf{Y}_e),
\label{eqn:span_xe_loss}
\end{equation}
where $f_{XE}$ is the cross-entropy loss; $\mathbf{Y}_s$ and $\mathbf{Y}_e$ denote the ground truth labels for the start and end boundaries, respectively. $\mathbf{Y}_{s/e}$ is a $n$-dim one-hot vector, which is generated by setting the index of the snippet contains $\tau_{s/e}$ as $1$, and others as $0$. Similarly, the KL-divergence objective is defined as:
\begin{equation}
    \mathcal{L}_{span} = D_{KL}(\mathbf{P}_s||\mathbf{\hat{Y}}_s) + D_{KL}(\mathbf{P}_e||\mathbf{\hat{Y}}_e),
\label{eqn:span_kl_div_loss}
\end{equation}
where $D_{KL}$ denotes KL-divergence; $\mathbf{\hat{Y}}_s$ and $\mathbf{\hat{Y}}_e$ are the ground truth start and end boundary distributions. Not specified in an one-hot $\mathbf{Y}_{s/e}$, the ground truth boundary distribution is formulated as $\mathbf{\hat{Y}}_{s/e}\sim\mathcal{N}(\tau_{s/e}, \sigma_{std}^2)$, where $\mathcal{N}(\mu,\sigma_{std}^2)$ is the normal distribution with expectation $\mu$ and standard deviation $\sigma_{std}$.

To summarize, we brief the main components of the TSGV method from input processing to answer prediction. Although existing TSGV methods may contain more sophisticated structures and diverse ancillary modules, their model frameworks generally follow this pipeline. Among the components, the effectiveness of the feature interactor highly affects TSGV performance. Proposal generation strategies are highly correlated with the design of the answer predictor, and each strategy has its own advantages and drawbacks. Lastly, all methods rely on effective feature extractors, mainly developed in computer vision and natural language processing areas.

\section{Datasets and Measures}
\label{sec:dataset_measure}
Datasets are essential resources for building and evaluating TSGV methods. We  review benchmark datasets and evaluation metrics.

\begin{table*}
    \caption{Statistics of the TSGV benchmark datasets. Different queries may correspond to the same moment.}
    \small
	\centering
	\begin{tabular}{ l |c c c| c c |c }
		\specialrule{.1em}{.05em}{.05em}
		Dataset & DiDeMo & Charades-STA & ActivityNet Captions & TACoS$_{\text{org}}$ & TACoS$_{\text{2DTAN}}$ & MAD  \\
		\hline
        Video Source & Flickr & Homes & YouTube & \multicolumn{2}{c|}{Lab Kitchen} & Movie \\
        Domain & Open & Indoor Activity & Open & \multicolumn{2}{c|}{Cooking} & Open \\
        \hline
        \# Videos & 10,464 & 6,672 & 14,926 & \multicolumn{2}{c|}{127} & 650 \\
        \# Moments & 26,892 & 11,767 & 71,953 & 3,290 & 7,069 & - \\
        \# Queries (or Annotations) & 40,543 & 16,124 & 71,953 & 18,818 & 18,227 & 384,600 \\
        Average \# Annotations per Video & 3.87 & 2.42 & 4.82 & 148.17 & 143.52 & - \\
        \hline
        Vocabulary Size & 7,785 & 1,303 & 15,505 & 2,344 & 2,287 & 61,400 \\
        \hline
        Average Video Length (seconds) & 30.00 & 30.60 & 117.60 & \multicolumn{2}{c|}{286.59} & 6,646.20 \\
        Min / Max Video Length (seconds) & - & 5.50 / 194.33 &  1.58 / 755.11 &  \multicolumn{2}{c|}{48.30 / 1,402.18} & - \\
        Average Moment Length (seconds) & - & 8.09 &  37.14 &  6.10 & 27.88 & 4.10 \\
        Min / Max Moment Length (seconds) & - & 1.68 / 80.80 & 0.05 / 408.80 & 0.31 / 166.97 &  0.48 / 843.20 & - \\
        Average Query Length (words)& - & 7.22 & 14.41 & 10.05 & 9.42 & 12.70 \\
        Min / Max Query Length (words)& - & 3 / 13 & 4 / 91 & 2 / 229 & 2 / 69 & - \\
        \specialrule{.1em}{.05em}{.05em}
	\end{tabular}
	\label{tab:data_stat}
\end{table*}

\subsection{Benchmark Datasets}
\label{ssec:datasets}
A TSGV dataset typically contains a collection of videos. Each video may come with one or more annotations, \ie moment-query pairs. Each annotation has a query corresponding to a moment in the video. A few TSGV datasets have been developed, covering various scenarios with distinct characteristics \eg different scenes, and activity complexities, summarized in 
Table~\ref{tab:data_stat}.\footnote{For DiDeMo and MAD datasets, we directly obtain their statistical results from the original papers. For others, we conduct  statistics on raw datasets. We also filter out or modify some invalid annotations in each dataset.}

\smallskip \noindent \textbf{DiDeMo} has its root in YFCC100M~\cite{thomee2016YFCC100M} dataset, and the latter contains over $100k$ Flickr videos about various human activities. Hendricks \etal~\cite{hendricks17iccv} randomly select over $14,000$ videos, then split and label video segments. Each segment is a five-second video clip, hence the length of ground truth moment is five seconds. DiDeMo dataset consists of $10,464$ videos and $40,543$ annotations in total, on average $3.87$ annotations per video. Note that the videos are released in the form of extracted visual features, hence we cannot provide detailed statistics in Table~\ref{tab:data_stat}.  Hendricks \etal~\cite{hendricks2018localizing} further collect a TEMPO dataset, which is built on top of DiDeMo, by augmenting language queries via the template model (template language) and human annotators (human language). Compared to DiDeMo, TEMPO contains more complex human-language queries.

\smallskip \noindent \textbf{Charades-STA} is built by Gao \etal~\cite{Gao2017TALLTA} from the Charades dataset~\cite{sigurdsson2016hollywood}. The Charades dataset contains $9,848$ annotated videos about human daily indoor activities for video activity recognition. The original dataset provides $27,847$ video-level sentence descriptions, $66,500$ temporally localized intervals for $157$ action categories, and $41,104$ labels for $46$ object categories. Based on Charades, Gao \etal~\cite{Gao2017TALLTA} design a semi-automatic way to construct Charades-STA. They first parse the activity labels from video descriptions using Stanford CoreNLP~\cite{manning2014stanford}, then match the labels with sub-sentences, and finally align sub-sentences with the original label-indicated temporal intervals. A collection of (sentence query, target moment) pairs are generated as annotations. Because the original descriptions are quite short, Gao \etal~\cite{Gao2017TALLTA} further combine consecutive descriptions into a more complex sentence to enhance the description complexity for test. Charades-STA contains $6,672$ videos and $16,124$ annotations. Average video length, moment length, and query length are $30.60$ seconds, $8.09$ seconds and $7.22$ words, respectively.

\smallskip \noindent \textbf{ActivityNet Captions} is developed by Krishna \etal~\cite{krishna2017dense} for dense video captioning task. However, the sentence-moment pairs in this dataset can naturally be adopted for TSGV task. The videos are taken from  ActivityNet~\cite{heilbron2015activitynet} dataset, a human activity understanding benchmark. ActivityNet provides samples from $203$ activity classes, with an average of $137$ untrimmed videos per class and $1.41$ activity instances per video~\cite{heilbron2015activitynet}. The official test set of ActivityNet Captions is withheld for competition, existing TSGV methods mainly use the official ``val1'' and/or ``val2'' development sets as test sets. Thus,  statistics of ActivityNet Captions in Table~\ref{tab:data_stat} does not consider its official test set. In total, there are $14,926$ videos and $71,953$ annotations in ActivityNet Captions, where each video contains $4.82$ temporally localized sentences on average. Average video and moment lengths are $117.60$ and $37.14$ seconds, respectively. Average query length is about $14.41$ words.

\smallskip \noindent \textbf{TACoS} dataset~\cite{regneri2013grounding} is selected from the MPII Cooking Composite Activities dataset~\cite{Rohrbach2012SDA}, originally developed for human activity recognition under specific scene, \ie composite cooking activities in lab kitchen. TACoS contains $127$ videos, and each video is associated with two types of annotations: (1) fine-grained activity labels with temporal location, and (2) natural language descriptions with temporal locations. The natural language descriptions are from crowd-sourcing annotators, who describe the video content by sentences~\cite{regneri2013grounding}. TACoS has $18,818$ moment-query pairs. Average video and moment lengths are $286.59$ and $6.10$ seconds, and average query length is $10.05$ words. Each video in TACoS contains $148.17$ annotations on average. We name this dataset TACoS$_{\text{org}}$ in Table~\ref{tab:data_stat}. A modified version TACoS$_{\text{2DTAN}}$ is made available by Zhang \etal~\cite{zhang2020learning}.  TACoS$_{\text{2DTAN}}$ has $18,227$ annotations. On average, there are $143.52$ annotations per video. The average moment length and query length after modification are $27.88$ seconds and $9.42$ words, respectively.

\smallskip \noindent \textbf{MAD}~\cite{soldan2021mad} is a large-scale dataset containing mainstream movies. Compared to previous datasets, MAD aims to avoid the hidden biases (detailed in Section~\ref{ssec:challenges}) and provide accurate and unbiased annotations for TSGV. Instead of relying on crowd-sourced annotations, Soldan \etal~\cite{soldan2021mad} adopt a scalable data collection strategy. They transcribe the audio description track of a movie and remove sentences associated with actor’s speech, to obtain highly descriptive sentences that are grounded in long-form videos. MAD contains $650$ movies with over $1,200$ hours of video length in total. Average video duration is around $110$ minutes. Each video in MAD is a full movie without pruning. MAD has $348,600$ queries with vocabulary size of $61,400$. Average query length is $12.7$ words. Average length of temporal moment in MAD is merely $4.1$ seconds, making the localization process more challenging.\footnote{At the time of writing, MAD dataset is not publicly available.} 

\begin{figure}[t]
    \centering
    \includegraphics[trim={0cm 0cm 0cm 0cm},clip,width=0.95\linewidth]{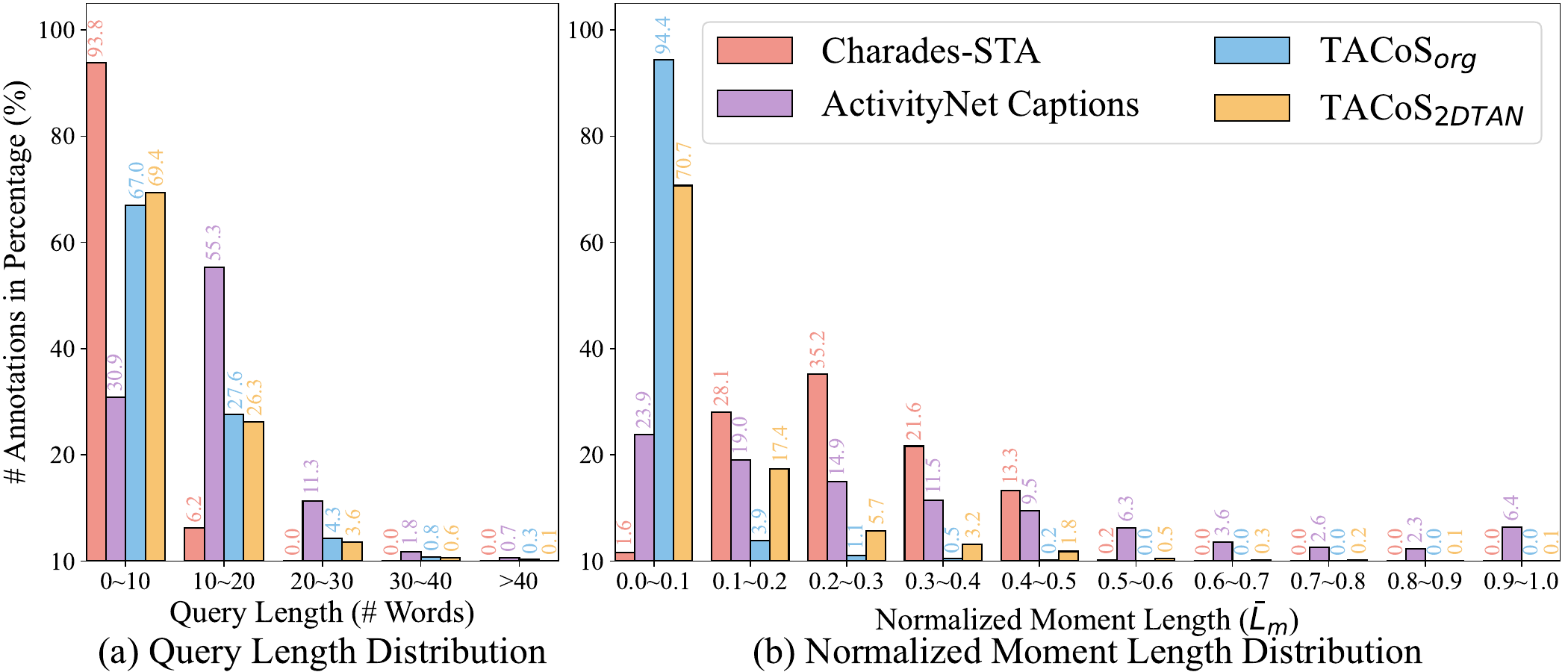}
    \caption{Statistics of query length and normalized moment length over  benchmark datasets.}
	\label{fig:data_stat_plot}
\end{figure}

Videos in the aforementioned datasets may be from open domain or constrained in narrow and specific scenes (see Table~\ref{tab:data_stat}). Open domain videos contain more diverse and complex activities, making them more challenging, but are closer to real-world scenarios. Although DiDeMo videos are from open domain, the answers in this dataset are in fixed-length, \ie five-second. The fixed length considerably reduces the complexity of finding answers in DiDeMo. 

ActivityNet Captions and DiDeMo have a much larger vocabulary size than Charades-STA and TACoS, suggesting that the former two datasets provide rich variations in language queries. From the perspective of query length (see Fig.~\ref{fig:data_stat_plot} (a)), a large portion of queries in Charades-STA ($93.8\%$) and TACoS ($>67.0\%$) has fewer than $10$ words. Query length distribution indicates that ActivityNet Captions contain more queries with complicated expressions. Fig.~\ref{fig:data_stat_plot} (b) depicts the normalized moment length ($\bar{L}_m$) distribution, against the length of its source video. A small $\bar{L}_m$ means the moment is difficult to retrieve due to moment sparsity~\cite{zhang2021natural}. The figure shows more than $70.7\%$ of the moments in TACoS has $\bar{L}_m\leq 0.1$, while $70.1\%$ moments in Charades-STA are in the range of $0.2<\bar{L}_m\leq 0.5$.

\begin{figure}[t]
    \centering
	\subfigure[An illustration of temporal IoU.]
	{
	    \label{fig:tiou}	
	    \includegraphics[trim={0cm 0cm 0cm 0cm},clip,width=0.4\textwidth]{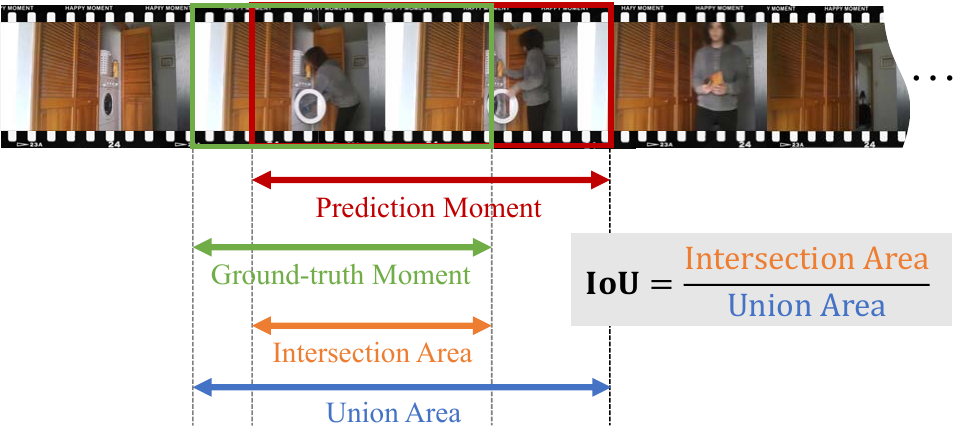}
	}
	\subfigure[An illustration of dR@\textit{n},IoU@\textit{m}.]
	{
	    \label{fig:dr_iou}	
	    \includegraphics[trim={0cm 0cm 0cm 0cm},clip, width=0.4\textwidth]{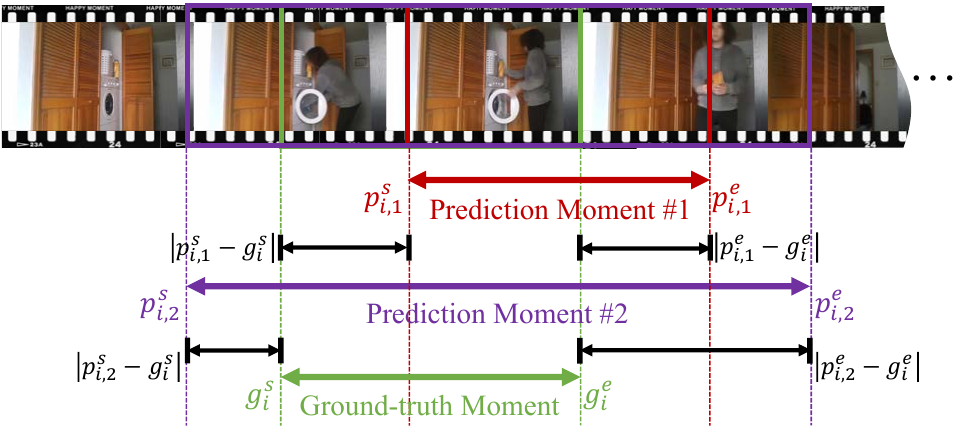}
	}
	\caption{The temporal intersection over union (IoU), and the discounted-R@\textit{n},IoU@\textit{m} (dR@\textit{n},IoU@\textit{m}).  $p_{i}^{s}$ and $p_{i}^{e}$ are start and timestamps of predicted moments, $g^{s/e}$ is start/end timestamp of ground-truth moment. $|\cdot|$ denotes absolute operation.}
	\label{fig:rank_iou}
\end{figure}

\subsection{Evaluation Metrics}
\label{ssec:metrics}
TSGV are generally evaluated by comparing  predictions with ground truth annotations. The widely used measures include: mean IoU (mIoU), $\langle$R@\textit{n}, IoU@\textit{m}$\rangle$, and  $\langle$dR@\textit{n}, IoU@\textit{m}$\rangle$. 

Intersection over Union (IoU) is a metric commonly used in object detection~\cite{girshick2014rich,ren2017faster,lin2020focal} for measuring the similarity between two bounding boxes. Hence, the standard IoU in object detection is defined on a two-dimensional spatial space. TSGV focuses on temporal dimension only. The temporal IoU is adopted to measure similarity between the ground truth and predicted moments in TSGV, illustrated in Fig.~\ref{fig:tiou}.  IoU is computed as the ratio of intersection area over union area between two moments, in the range of $0.0$ to $1.0$. A larger IoU means the two moments match better, and $\text{IoU}=1.0$ denotes the exact match. The mIoU metric is the average temporal IoUs among all annotations in the test set. Mathematically, mIoU is defined as:
\begin{equation}
    \text{mIoU} = \frac{1}{N_q}\sum_{i=1}^{N_q} \text{IoU}_{i}
\end{equation}
where $N_q$ denotes the total number of annotations or query samples, and $\text{IoU}_{i}$ is the IoU value of $i$-th sample.

The mIoU is computed based on the single top-ranked prediction for each query. However, given a query, the top-ranked prediction by a TSGV model may not always have the best match with ground truth.  It is reasonable to relax the evaluation by considering top-$n$ retrieved moments for each query.  The $\langle$R@\textit{n}, IoU@\textit{m}$\rangle$~\cite{hu2016natural} is the percentage of queries, having at least one result whose temporal IoU with ground truth is larger than \textit{m} among the top-\textit{n} retrieved moments. For query $q_i$, among its top-\textit{n} retrieved moments, if there exists at least one moment whose IoU with ground truth is larger than \textit{m}, then $q_i$ is considered as positive, denoted by $r(n,m,q_i)=1$. Otherwise, $r(n,m,q_i)=0$. Thus, $\langle$R@\textit{n}, IoU@\textit{m}$\rangle$ is calculated as:
\begin{equation}
    \text{R@}\textit{n}\text{,IoU@}\textit{m}=\frac{1}{N_q}\sum_{i=1}^{N_q}r(n,m,q_i)
\end{equation}

Yuan \etal~\cite{yuan2021closer} reveal that  $\langle$R@\textit{n}, IoU@\textit{m}$\rangle$ is unreliable for small IoU thresholds. A method tends to generate long predictions if a substantial proportion of ground truth moments are long in a dataset. The method increases its chance of correct prediction under small IoU thresholds. 
Discounted-R@\textit{n}, IoU@\textit{m} $\langle$dR@\textit{n}, IoU@\textit{m}$\rangle$ is proposed to alleviate this problem~\cite{yuan2021closer}. This new measure leverages ``temporal distance'' between the predicted and ground truth moments to discount $r(n,m,q_i)$ value.  $\langle$dR@\textit{n}, IoU@\textit{m}$\rangle$ is calculated as:
\begin{equation}
    \text{dR@}\textit{n}\text{,IoU@}\textit{m} = \frac{1}{N_q}\sum_{i=1}^{N_q}r(n,m,q_i)\cdot\alpha_i^s\cdot\alpha_i^e
\end{equation}
where discounted ratio $\alpha_i^*=1-|p_i^* - g_i^*|$, $*\in\{s,e\}$. $|p_i^* - g_i^*|$ is the absolute distance between the boundaries of the predicted and the ground truth moments (see Fig.~\ref{fig:dr_iou}). Note that both $p_i^*$ and $g_i^*$ are normalized in $0.0$ to $1.0$ by dividing the corresponding whole video length. If the predicted moment exactly matches ground truth, then the discounted ratio $\alpha_i^*=1$, and the metric degrades to $\langle$R@\textit{n}, IoU@\textit{m}$\rangle$. Otherwise, even if IoU threshold is met, $r(n,m,q_i)$ is discounted by $\alpha_i^*$, which helps to restrain overlong predictions. 

In Fig.~\ref{fig:dr_iou}, $(p_{i,1}^s,p_{i,1}^e)$ and $(p_{i,2}^s,p_{i,2}^e)$ are two example predicted moments of query $q_i$, and $(g_i^s,g_i^e)$ is ground truth moment. Suppose both Predictions 1 and 2 in Fig.~\ref{fig:dr_iou} have the same IoU value which satisfies $\text{IoU}\geq m$, ($m\leq 0.5$ here), $\langle$dR@\textit{n}, IoU@\textit{m}$\rangle$ penalizes more on Prediction 2 since its temporal boundaries are farther from ground truth. With respect to $\langle$R@\textit{n},IoU@\textit{m}$\rangle$ and $\langle$dR@\textit{n},IoU@\textit{m}$\rangle$ metrics, community is habituated to set $n\in\{1,5,10\}$ and $m\in\{0.3, 0.5, 0.7\}$.

\section{TSGV Methods}
\label{sec:method_overview}

\begin{figure*}[tb]
    \centering
    \includegraphics[trim={0cm 0cm 0cm 0cm},clip,width=0.98\linewidth]{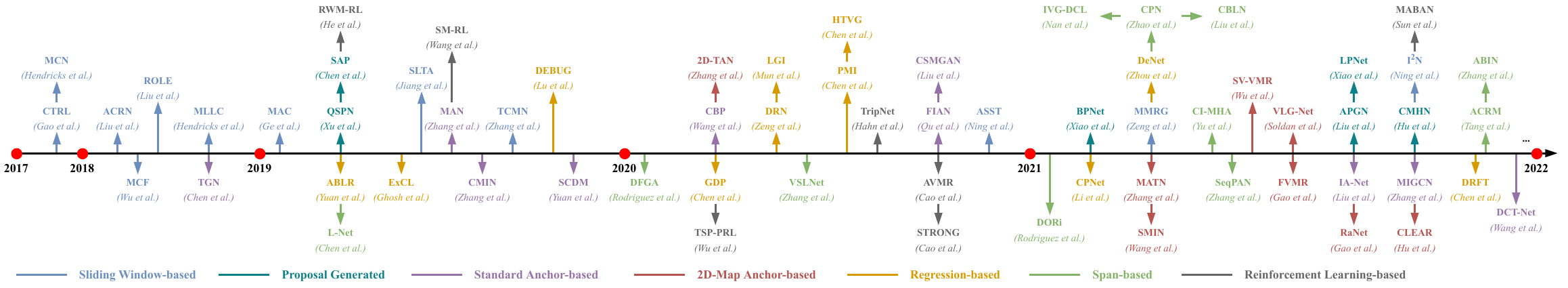}
    \caption{Chronological overview of selected supervised TSGV methods in different categories. The methods plotted at the same position on the timeline are published in the same venue.}
	\label{fig:chronological_overview}
\end{figure*}

The majority of solutions proposed for TSGV belong to the supervised learning paradigm. Early solutions mainly rely on sliding windows or segment proposal networks to pre-sample proposal candidates from the input video. Then, the proposals are paired with the query to generate the best answers through cross-modal matching. However, this two-stage ``\textit{propose-and-rank}'' pipeline is inefficient, because densely sampling candidates with overlap are essential to achieve high accuracy, leading to redundant computation and low efficiency. Meanwhile, the pairwise proposal-query matching may also neglect the contextual information. To overcome these drawbacks, alternative solutions like anchor-based and proposal-free methods are developed to address TSGV in an ``\textit{end-to-end}'' manner. These methods encode the entire video sequence and all video information is maintained in the model, gradually becoming the predominant solution for TSGV. Fig.~\ref{fig:chronological_overview} depicts a chronological overview of the development of supervised learning for TSGV.

Supervised learning requires a large number of annotated samples to train a TSGV method. Considering the difficulty and cost of data annotation, recent studies attempt to solve TSGV with weakly-supervised learning. These methods relieve the annotation burden by learning from video-query pairs without the detailed annotation of the temporal locations of  events in videos.

\begin{figure}[t]
    \centering
    \includegraphics[trim={0cm 0cm 0cm 0cm},clip,width=0.95\linewidth]{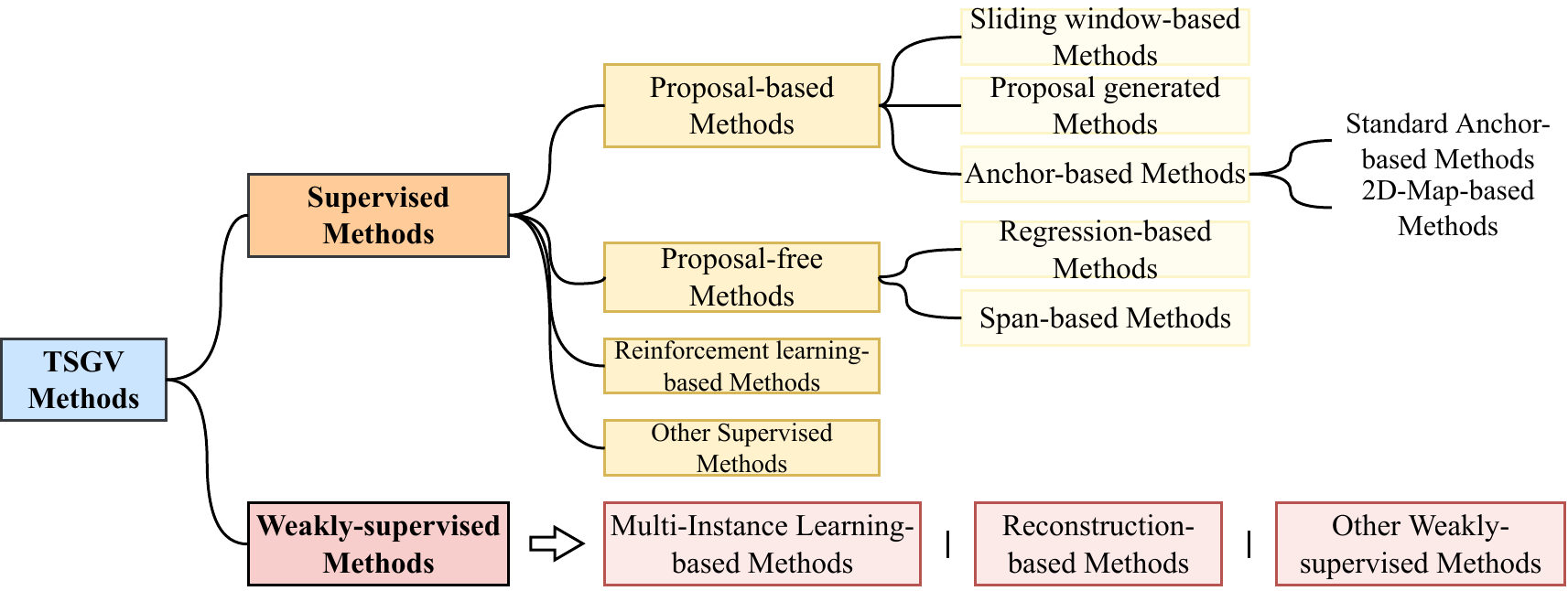}
    \caption{A taxonomy of methods for TSGV.}
	\label{fig:taxonomy}
\end{figure}

Accordingly, the simple classification of proposal-based and proposal-free methods in Section~\ref{sec:background} is incapable of covering all TSGV methods.  Based on the method architecture and learning algorithm, we propose a new taxonomy in Fig.~\ref{fig:taxonomy} to categorize TSGV methods. Next, we review the solutions to TSGV following this taxonomy and discuss the characteristics of each method category. Because the majority are supervised learning solutions, this section is organized mainly based on the categories under supervised learning.  

\subsection{Proposal-based Method}
\label{ssec:proposal_based}
Depending on the ways to generate proposal candidates,  proposal-based methods can be grouped into three categories, \ie sliding window-based, proposal generated, and anchor-based methods. Sliding window-based and some of the proposal-generated methods follow a two-stage propose-and-rank pipeline, where the generation of proposal candidates is separated from the model computation. Anchor-based methods incorporate proposal generation in model computation to achieve end-to-end learning. 

\subsubsection{Sliding Window-based Method}
\label{para:sw_method}
The sliding window-based method adopts multi-scale sliding windows (SW) to generate proposal candidates (ref. Fig.~\ref{fig:proposal_sw}). Then the multimodal matching module finds the best matching proposal for a query. CTRL~\cite{Gao2017TALLTA} and MCN~\cite{hendricks17iccv} are two canonical SW methods, which are also pioneering work in TSGV. They define the task and construct corresponding benchmark datasets. 

\begin{figure}[t]
    \centering
    \includegraphics[trim={0cm 0cm 0cm 0cm},clip,width=0.95\linewidth]{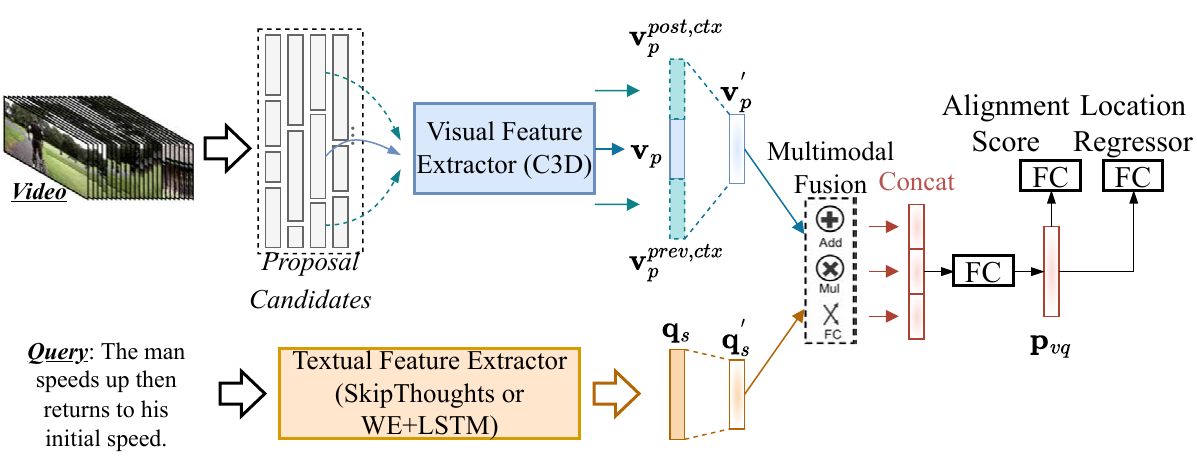}
    \caption{CTRL architecture, reproduced from Gao \etal~\cite{Gao2017TALLTA}.}
	\label{fig:ctrl_model}
\end{figure}

CTRL first produces proposals of various lengths through sliding windows, then encodes these proposals by a visual encoder, shown in Fig.~\ref{fig:ctrl_model}. The query is converted to sentence representation via a textual encoder. For cross-modal reasoning, it builds a relatively simple multimodal processing module with three operators, \ie add, multiply, and fully connected (FC) layer, to fuse  visual and textual features. CTRL designs multi-task objectives by using both an alignment predictor and a regressor. The alignment predictor computes the matching score between the proposal and query (ref. Eqn.~\ref{eqn:aln_loss}). However, for an aligned proposal-query pair, the position of the proposal may not match the ground truth moment exactly. The regressor uses the smoothed $L_1$ loss to compute the corresponding offsets (ref. Eqn.~\ref{eqn:reg_loss}) to better align the proposal.

Different from CTRL, MCN aims to project both proposal and query features to a common embedding space. Then, it encourages the distance between the query and the aligned proposal to be smaller than that of negative proposals. Specifically, MCN minimizes the squared distance between the query and proposals to supervise model learning. Negative proposals can be misaligned proposals within the same video (intra-video), or proposals from other videos (inter-video). Thus, MCN builds both intra- and inter-triple-based ranking losses (ref. Eqn.~\ref{eqn:triple_loss}) as objectives. The intra-loss differentiates subtle differences within a video, and the inter-loss differentiates broad semantic concepts. Based on MCN, Hendricks \etal~\cite{hendricks2018localizing} further proposes MLLC, which treats the video context as a latent variable and unifies MCN and CTRL for moment localization.

The prior methods encode the entire query into one feature vector and apply simple cross-modal reasoning for feature fusion. However, treating queries holistically may obfuscate the keywords that have rich temporal and semantic cues. The simple fusion strategy also leads to inferior cross-modal understanding. Temporal dependencies and reasoning between video events and texts are not fully considered. Spatial-temporal information inside the video or query is also overlooked. A number of methods are proposed to address these issues. Among them, ROLE~\cite{Liu2018CML}, MCF~\cite{wu2018multi}, ACRN~\cite{Liu2018AMR}, TCMN~\cite{zhang2019exploiting}, and ASST~\cite{ning2020asst} mainly focus on refining the multimodal interaction/fusion between visual and textual features, through more sophisticated structures or semantic decomposition of video/query. ACL~\cite{ge2019mac}, built upon CTRL, explicitly mines activity concepts from video and language as prior knowledge, to calibrate the confidence of the proposal to be the target moment. In addition to multimodal interaction refinement, SLTA~\cite{jiang2019cross} and MMRG~\cite{zeng2021multi} also exploit to incorporate appearance knowledge, \ie object-level spatial visual features, to enhance cross-modal reasoning as an additional view of video content. Instead of generating proposals at the initial stage, Ning \etal~\cite{ning2021interaction} equip SW strategy inside their model enabling end-to-end training. CAMG~\cite{hu2022camg} designs a context-aware moment graph method, which utilizes semantic and temporal moment graphs to refine the proposals with semantic and position information.

In general, early SW-based methods have simple architectures. These methods lack both in-depth analyses of semantic knowledge of modalities and fine-grained multimodal fusion mechanisms, leading to inferior performance. The following work attempts to address these weaknesses by devising various techniques to better exploit video content and query, enhancing cross-modal reasoning between them. Despite continuous improvements, the two-stage sliding window-based methods suffer from inevitable drawbacks. Specifically, densely sampling proposals with multi-scale sliding windows result in heavily computational costs, as many overlapped areas are re-computed. These methods are also sensitive to negative samples, where fallacious negative samples may lead to inferior results.

\subsubsection{Proposal Generated Method}
\label{para:pg_method}
The proposal generated (PG) method alleviates the computation burden of SW-based methods by avoiding the dense sampling process. Instead, PG methods generate proposals conditioned on the query. The number of proposals hence reduces remarkably.

\begin{figure}[t]
    \centering
	\subfigure[Query-guided segment proposal network.]
	{
	    \label{fig:qspn_proposal}	
	    \includegraphics[trim={0cm 0cm 2.8cm 0cm},clip,width=0.9\linewidth]{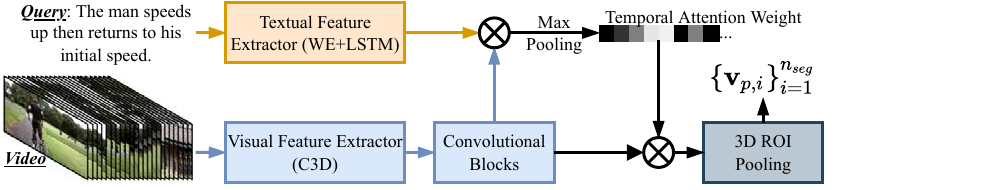}
	}
    \subfigure[The early fusion retrieval model of QSPN.]
	{
	    \label{fig:qspn_model}	
	    \includegraphics[trim={1cm 0cm 0cm 0cm},clip, width=0.85\linewidth]{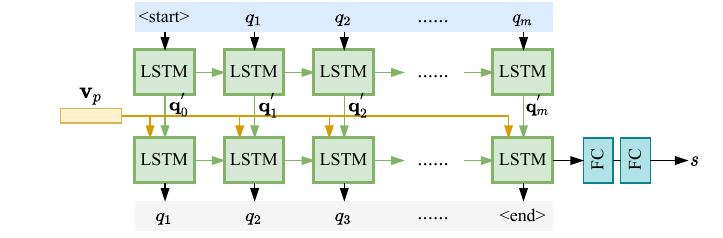}
	}
	\caption{QSPN architecture, reproduced from Xu \etal~\cite{Xu2019MultilevelLA}.}
	\label{fig:qspn}
\end{figure}

Early proposal-generated methods still follow the two-stage propose-and-rank pipeline. Xu \etal~\cite{Xu2018TexttoClipVR} employ a pre-trained segment proposal network (SPN)~\cite{xu2017r} for proposal candidate generation, rather than adopting sliding windows.
Based on Xu \etal~\cite{Xu2018TexttoClipVR}, QSPN~\cite{Xu2019MultilevelLA} further ameliorates SPN to produce query-specific proposal candidates. As illustrated in Fig.~\ref{fig:qspn_proposal}, QSPN interacts query embedding with visual features to derive temporal attention weights and re-weights the visual features to refine proposal generation. With the generated proposal feature, QSPN sequentially encodes the proposal with each token in the query and predicts the similarity score, at last, shown in Fig.~\ref{fig:qspn_model}. QSPN is optimized by triple-based ranking loss (ref. Eqn.~\ref{eqn:triple_loss}), while a captioning loss is adopted to improve performance via query re-generation. Similarly, SAP~\cite{chen2019semantic} directly trains a visual concept detector to generate proposal candidates by measuring visual-semantic correlations between query and video frames.

Although the two-stage PG methods mitigate computation complexity to some degree, they still encounter some ineluctable drawbacks. To achieve good performance, PG methods still need to sample proposal candidates relatively densely, to increase the chance that at least one proposal can cover or is close to the ground truth moment. Similar to SW-based methods, the two-stage PG methods also rely on ranking-based objectives, making them sensitive to negative samples. Besides, proposal candidates are processed separately; hence, individual pairwise proposal-query matching may neglect the contextual information.

\begin{figure}[t]
    \centering
    \includegraphics[trim={0cm 0cm 0cm 0cm},clip,width=0.9\linewidth]{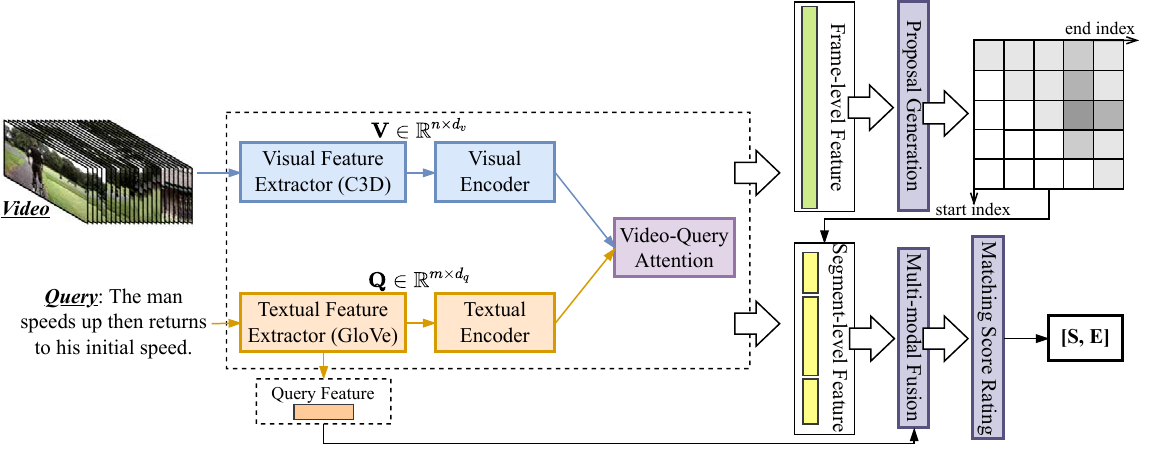}
    \caption{BPNet architecture, reproduced from Xiao \etal~\cite{xiao2021boundary}.}
	\label{fig:bpnet_model}
\end{figure}

To overcome these defects, recent solutions~\cite{xiao2021boundary,liu2021adaptive,xiao2021natural,gao2022efficient,liu2022skimming} reformulate the pipeline of PG methods to a single-pass pattern in an end-to-end manner. Specifically, BPNet~\cite{xiao2021boundary} (see Fig.~\ref{fig:bpnet_model}) and APGN~\cite{liu2021adaptive} replace the separate proposal generator by a proposal-free module (detailed in Section~\ref{ssec:proposal_free}) and jointly train it with the main model. In this case, the proposal generation module is supervised by the ground truth moment, and only a few proposals are required to be generated. Besides, since the whole video is encoded as a feature sequence (ref. Section~\ref{ssec:tsgv_feat_encoder}), visual features are jointly learned and interacted with the query. Thus, the model is able to consider contextual information. LPNet~\cite{xiao2021natural} maintains a boundary-aware predictor and learnable proposal module in parallel, where the boundary-aware predictor could refine predictions of the learnable proposal module. CMHN~\cite{hu2021video} generates proposal candidates with 1D regular convolution and models proposal-query matching in Hamming space through cross-modal hashing. Similar to BPNet, Gao \etal~\cite{gao2022efficient} also adopt a proposal-free module for candidate generation followed by a candidate refinement. Furthermore, SLP~\cite{liu2022skimming} proposes to first select the best-matched frame conditioned on the query. Then it constructs an initial segment based on the frame and dynamically updates it by exploring the adjacent frames with similar semantics.

\subsubsection{Anchor-based Method}
\label{para:anchor_method}
Sliding window and the early proposal-generated methods follow the two-stage propose-and-rank pipeline which suffers from various drawbacks. Researchers then source for alternative structures without pre-cutting proposal candidates at the input stage. One kind of solution is anchor-based methods, which incorporate proposal generation into answer prediction and maintain the proposals with various learning modules. According to how the anchors are produced and maintained, we further classify them into standard anchor-based and 2D-Map methods. 

\begin{figure}[t]
    \centering
    \includegraphics[trim={0cm 0cm 0cm 0cm},clip,width=0.95\linewidth]{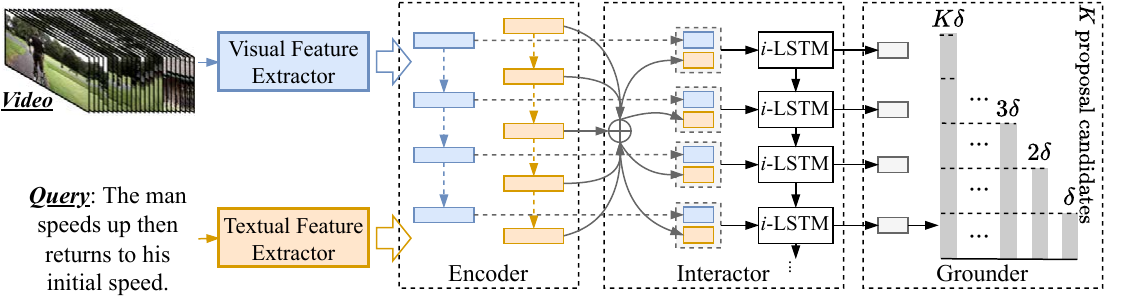}
    \caption{TGN architecture, reproduced from Chen \etal~\cite{chen2018temporally}.}
	\label{fig:tgn_model}
\end{figure}

\smallskip\noindent\textbf{Standard Anchor-based Method.} Methods in this category produce proposal candidates with multi-scale anchors and maintain them sequentially or hierarchically in the model. They aggregate contextual multimodal information and generate the final grounding result in one pass. The first anchor-based method for TSGV is Temporal GroundNet (TGN) by Chen \etal~\cite{chen2018temporally}, shown in Fig.~\ref{fig:tgn_model}. TGN temporally captures the evolving fine-grained frame-by-word interactions between video and query. At each time step, multi-scale proposal candidates ending at the current time are generated using pre-set anchors. Then a sequential LSTM grounder simultaneously scores the group of proposals. TGN adopts weighted binary cross-entropy loss (ref. Eqn.~\ref{eqn:bce_loss}) to optimize the model. In contrast, MAN~\cite{zhang2019man} and SCDM~\cite{yuan2019semantic,yuan2020semantic} adopt temporal convolutional networks to produce proposal candidates hierarchically. That is, proposals with different scales are generated at different levels of the stacked temporal convolution module. SCDM also adopts different multi-scale anchors compared to the standard version.
Specifically, it imposes different scale anchors based on a basic span centered at each time step.

Subsequent work generally follows the strategies of TGN or SCDM with more sophisticated learning modules and/or auxiliary objectives. To be specific, CMIN~\cite{zhu2019cross,lin2020moment}, CBP~\cite{Wang2020TemporallyGL}, FIAN~\cite{qu2020fine}, HDRR~\cite{ma2021hierarchical}, and MIGCN~\cite{zhang2021multimodal} adopt the strategy of TGN, while CSMGAN~\cite{liu2020jointly}, RMN~\cite{liu2020reasoning}, IA-Net~\cite{liu2021progressively}, and DCT-Net~\cite{wang2021dctnet} apply the strategy of SCDM. These solutions design various cross-modal reasoning strategies to perform a more fine-grained and deeper multimodal interaction between video and query for precise moment localization. In addition, CBP~\cite{Wang2020TemporallyGL} introduces an auxiliary boundary module to compute the confidence of the feature at each time step to be the boundary of the target moment. Some works adopt boundary regression modules to refine the generated moments' start and end time points. MIGCN~\cite{zhang2021multimodal} develops a rank module apart from the boundary regression module to distinguish the optimal proposal from a set of similar proposal candidates. ECCL~\cite{liu2021eccl} designs a sliding convolution locator to iteratively predict the best proposal candidates. MA3SRN~\cite{liu2022exploring1} incorporates optical-flow-guided motion-aware, detection-based appearance-aware, and 3D-aware object features to interact with the query for better grounding.

\noindent\textbf{2D-Map Anchor-based Method.} Standard anchor-based method produces proposal candidates with preset multi-scale anchors and maintains them sequentially or hierarchically. These proposals are individually processed  and their temporal dependencies are not well considered. Furthermore, the lengths of proposals are restricted by preset anchors. 2D-Map methods use a two-dimensional map to model the temporal relations between proposal candidates, shown in Fig.~\ref{fig:proposal_2d}. Theoretically, 2D-Map could enumerate all possible proposals at any length, while maintaining their adjacent relations.

\begin{figure}[t]
    \centering
    \includegraphics[trim={0cm 0cm 0.5cm 0cm},clip,width=0.99\linewidth]{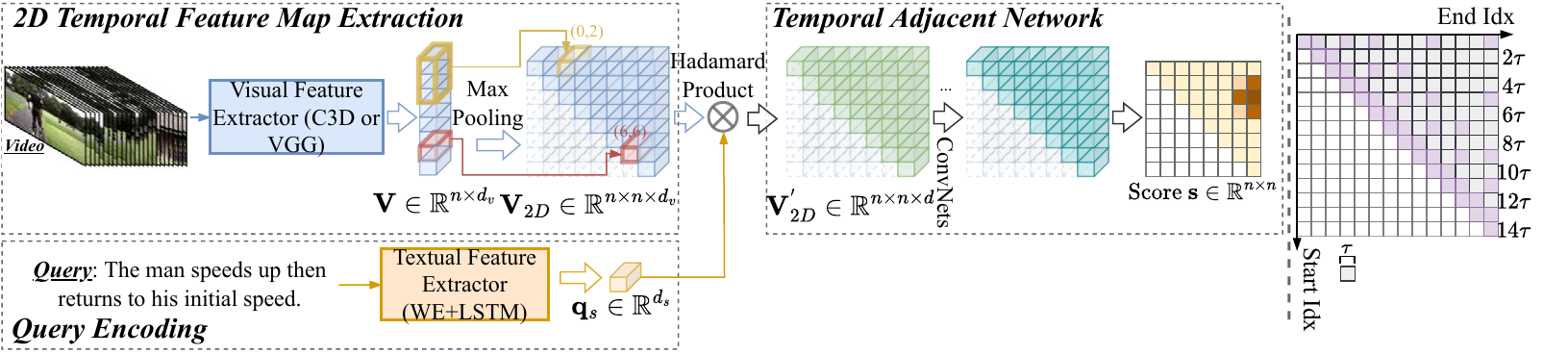}
    \caption{2D-TAN architecture, reproduced from Zhang \etal~\cite{zhang2020learning}.}
	\label{fig:2d-tan_model}
\end{figure}

Before 2D-Map methods, a prior work TMN~\cite{liu2018temporal} first proposes to enumerate all possible consecutive segments as proposals and predict the best-matched proposal as the result through interacting each proposal with the query. However, TMN generates proposals in the answer predictor; its enumeration strategy is more like a variant of a standard anchor-based strategy.

2D-TAN~\cite{zhang2020learning} is the first solution modeling proposal with a 2D temporal map, and its overall architecture is shown in Fig.~\ref{fig:2d-tan_model} left. 2D-TAN first extracts visual features and converts them into a 2D feature map, while the query is encoded in sentence-level representation. A temporal adjacent network is proposed to fuse the query feature with each proposal feature and embed the video context information with a convolutional network. As shown in Fig.~\ref{fig:2d-tan_model} right, 2D-TAN divides the video into evenly spaced video snippets with duration $\tau$, where $(i,j)$ on the 2D map denotes a proposal candidate from time $i\tau$ to $j\tau$\footnote{$i\leq j$, \ie only the upper triangular area of 2D map is valid}. Instead of enumerating all possible consecutive segments as proposals, 2D-MAN proposes a sparse sampling strategy to remove redundant moments which have large overlaps with the selected proposals. The model adopts binary cross-entropy loss for model learning. 2D-TAN is further extended~\cite{zhang2021ms2dtan} with multi-scale modeling to achieve a larger receptive field and obtain richer contexts. The extended version reduces the complexity of proposal generation from quadratic to linear, making dense video prediction more efficient.

Due to its effectiveness, a series of work follows 2D-TAN's proposal generation\footnote{Some methods follow 2D-TAN's proposal generation to produce proposal candidates, but they may not maintain the proposals in a 2D map.} or its overall structure. As illustrated in Fig.~\ref{fig:2d-tan_model}, 2D-TAN directly encodes the query into the sentence-level feature and interacts with proposals via a simple Hadamard product. In this sense, multimodal interaction is overlooked. To remedy, PLN~\cite{zheng2021progressive}, SMIN~\cite{wang2021structured}, CLEAR~\cite{hu2021coarse}, and STCM-Net~\cite{jia2022stcmnet} disentangle video proposals into different temporal granularities~\cite{zheng2021progressive,jia2022stcmnet} or different semantic contents~\cite{wang2021structured,hu2021coarse}, and perform cross-modal reasoning at both coarse- and fine-grained granularities. VLG-Net~\cite{soldan2021vlg} and RaNet~\cite{gao2021relation} maintain query words and video proposals in a graph and adopt GCN~\cite{huang2020aligned,xu2020g} to conduct both intra- and inter-modal interactions. SV-VMR~\cite{wu2021diving} decomposes the query into semantic roles~\cite{shi2019simple} and performs multi-level cross-modal reasoning at the semantic level. MATN~\cite{zhang2021multi} further concatenates proposals and query words into a sequence and encodes them through a single-stream transformer network. It also devises a novel multi-stage boundary regression to refine the predicted moments. Instead of using the simple Hadamard product, DMN~\cite{wang2021negative} proposes to project proposals and query features to a common embedding space and leverage metric learning for cross-modal pair discrimination. FVMR~\cite{gao2021fast} and CCA~\cite{wu2022learning} devise joint semantic embedding module for multimodal interaction to facilitate the cross-modal reasoning. Guo \etal~\cite{guo2022hybird} introduce the Wasserstein distance~\cite{pele2009fast} to match video-text domain. DCLN~\cite{zhang2022dualchannel} and TACI~\cite{shin2022learning} decompose 2D-Map into start and end channels for cross-modal reasoning and then fuses to 2D-Map features to facilitate model training. Xu \etal~\cite{xu2022contrastivela} propose a contrastive language-action pre-training framework for TSGV. Bao \etal~\cite{bao2022learning} address the bias issue via a sample-reweighting-based debiased temporal localizer. Moreover, a series of recent work~\cite{zheng2022teampp,ding2022exploring,wang2022crossmodal,wang2022language,sun2022yount} focuses on developing multi-stage cross-modal fusion module in hierarchy, sequence, or multi-granularity, for better moment prediction.

\subsection{Proposal-free Method}
\label{ssec:proposal_free}
Proposal-based methods perform various proposal generations and essentially depend on the ranking of proposal candidates. In contrast, proposal-free methods directly predict the start and end boundaries of the target moment on fine-grained video snippet sequence, without ranking a vast of proposal candidates. Depending on the format of moment boundaries, proposal-free methods are categorized into regression-based and span-based methods.

\subsubsection{Regression-based Method}
\label{para:regress_method}

\begin{figure}[t]
    \centering
    \includegraphics[trim={0cm 0.2cm 0cm 0cm},clip,width=0.9\linewidth]{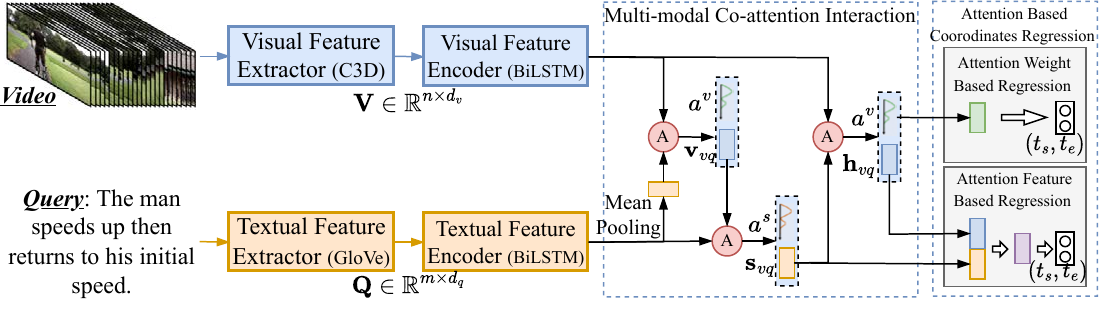}
    \caption{ABLR architecture, reproduced from Yuan \etal~\cite{yuan2019to}.}
	\label{fig:ablr_model}
\end{figure}

Regression-based method computes a time pair $(t_s,t_e)$ and compares the computed pair with ground truth $(\tau_s,\tau_e)$ for model optimization. Attention-based location regression (ABLR)~\cite{yuan2019to} is one of the first regression-based solutions for TSGV. Depicted in Fig.~\ref{fig:ablr_model}, ABLR extracts visual and textual features and encodes them through BiLSTM networks to aggregate contextual information, respectively. Then, a three-stage multimodal co-attention is developed to perform cross-modal reasoning. The multimodal feature is fed to the regressor for moment prediction. ABLR explores two types of regressors. One is attention weight-based regression, which takes video attention weights as input. Another is attended feature-based regression, which fuses the attended visual and textual features as inputs. The model is optimized by the smoothed $L_1$ loss.  ABLR also devises an attention calibration loss to refine video attention, which encourages higher attention weights to video snippets within the ground truth moment.

Concurrently, ExCL~\cite{ghosh2019excl} also addresses TSGV by regression and designs three different answer predictors following ideas from reading comprehension in NLP~\cite{seo2017bidaf,wei2018fast,huang2018fusionnet}. Similar to proposal-based methods, subsequent regression work~\cite{lu2019debug,chen2020rethinking,zhang2020simple,zeng2020dense,mun2020local,li2021proposal,zhou2021embracing,liu2021single,li2022compositional,fang2022hierarchical,xu2022hisa,liu2022memory,li2022phrase,guo2022taohighlight} dives in designing various feature encoding and cross-modal reasoning strategies for superior multimodal interaction and accurate moment localization. From the perspective of regression, DEBUG~\cite{lu2019debug}, GDP~\cite{chen2020rethinking}, and DRN~\cite{zeng2020dense} analyze the data imbalance issue in TSGV: the number of video frames is large, but the positive samples are sparse \ie only two frames for start and end timestamps. They regard all frames within the ground truth moment as positive and densely predict the distances to the boundaries for each frame within the ground truth moment to mitigate the sparsity issue. CMA~\cite{zhang2020simple} and DeNet~\cite{zhou2021embracing} study bias issue in TSGV. Specifically, CMA~\cite{zhang2020simple} rectifies the inevitable annotation bias by moment boundary ambiguities via a two-branch cross-modality attention network and a task-specific regression loss. VISA~\cite{li2022compositional} adopts a variational cross-graph correspondence learning with regression head, to study the generalization ability of the model to queries with novel compositions of seen words. HLGT~\cite{fang2022hierarchical} and MGSL-Net~\cite{liu2022memory} deeply mine the transformer variants~\cite{vaswani2017attention} for TSGV. HiSA~\cite{xu2022hisa} introduces contrastive learning to model intra-video entanglement and inter-video connection as auxiliary objectives. PLPNet~\cite{li2022phrase} further decomposes the query into phrases and localizes each phrase jointly as an ancillary. DeNet~\cite{zhou2021embracing} disentangles query into relations and modified features and devises a debias mechanism to alleviate both query uncertainty and annotation bias issues.

There are also regression methods~\cite{chen2020hierarchical,chen2020learning,chen2021end,liu2022exploring} incorporating additional modalities from video to improve the localization performance. For instance, HVTG~\cite{chen2020hierarchical} and MARN~\cite{liu2022exploring} extract both appearance and motion features from video. In addition to appearance and motion, PMI~\cite{chen2020learning} further exploits audio features from the video extracted by SoundNet~\cite{aytar2016soundnet}.  DRFT~\cite{chen2021end} leverages the visual, optical flow, and depth flow features of video, and analyzes the retrieval results of different feature view combinations.

\subsubsection{Span-based Method}
\label{para:span_method}
Span-based methods aim to predict the probability of each video snippet/frame being the start and end positions of the target moment. Inspired by the reading comprehension (RC) task in NLP~\cite{seo2017bidaf,wei2018fast,huang2018fusionnet}, L-Net~\cite{chen2019localizing} and ExCL~\cite{ghosh2019excl} first formulate TSGV as a span prediction task. In addition to the regression-based  predictors, ExCL also designs corresponding span prediction heads.

\begin{figure}[t]
    \centering
    \includegraphics[trim={0cm 0cm 0.3cm 0cm},clip,width=0.95\linewidth]{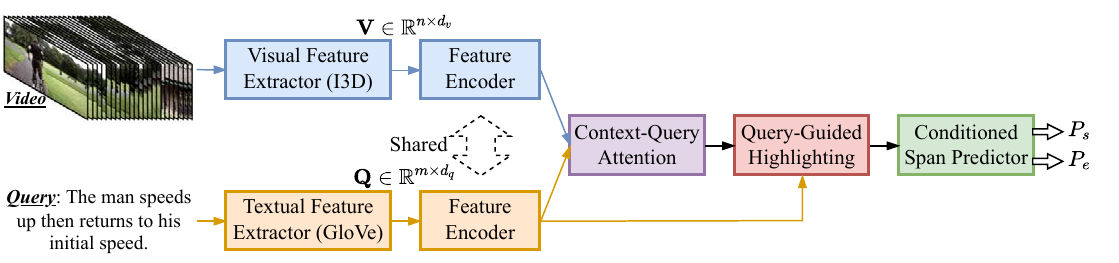}
    \caption{VSLNet architecture, reproduced from Zhang \etal~\cite{zhang2020vslnet}.}
	\label{fig:vslnet_model}
\end{figure}

Based on these two works, Zhang \etal~\cite{zhang2020vslnet} compare differences between RC and TSGV tasks and propose VSLNet. Specifically, video is continuous and causal relations between video events are usually adjacent, while words in query are discrete and demonstrate syntactic structure. Shown in Fig.~\ref{fig:vslnet_model}, VSLNet exploits a context-query attention modified from QANet~\cite{wei2018fast} to perform fine-grained multimodal interaction. Then a conditioned span predictor computes the probabilities of the start/end boundaries of the target moment. To bridge the gap between RC and TSGV, VSLNet introduces a query-guided highlighting module, which effectively narrows down the moment search space to a smaller highlighted region. Existing methods including VSLNet generally perform better on short videos than on long videos. Their follow-up work~\cite{zhang2021natural} extends VSLNet to handle long videos by incorporating concepts from multi-paragraph question answering~\cite{clark2018simple}. Long videos are split into multiple short videos and a hierarchical searching strategy is deployed for moment localization.

In general, the overall frameworks of regression- and span-based methods are similar. Thus, the continuous performance improvements of subsequent work~\cite{rodriguez2020proposal,zhang2021parallel,zhao2021cascaded,nan2021interventional,liang2021local,yu2021cross,tang2021multi,tang2021frame,zhang2021temporal,qi2021collaborative,rodriguez2021dori,liu2021context,zhang2022natural,hao2022cansv,qi2021coarsetofine,yang2022entityaware,shen2022joint,rodriguezOpazo2021locformer,fu2022multiple,zhang2022naturallanguage,zeng2022point,hao2022query,liu2022reducingtv,xu2022stdnet,li2022towards,huang2022videoal} are also achieved by modifying the feature encoding and multimodal interaction modules, introducing auxiliary objectives, and/or augmenting additional features. In particular, SeqPAN~\cite{zhang2021parallel} introduces the concept of named entity recognition~\cite{ma2016end,zhou2019dual,yu2020named} in NLP by splitting the snippet sequence into begin, inside, and end regions of the target moment, and background region. IVG-DCL~\cite{nan2021interventional} introduces a dual contrastive learning mechanism to enhance multimodal interaction and leverages a structured causal model~\cite{pearl2016causal} to address the selection bias of TSGV. CI-MHA~\cite{yu2021cross} proposes to remedy the start/end prediction noise caused by annotator disagreement via an auxiliary moment segmentation task. ABIN~\cite{zhang2021temporal} devises an auxiliary adversarial discriminator network to produce coordinate and frame correlation distributions for moment boundary refinement. DORi~\cite{rodriguez2021dori} incorporates appearance features and captures the relations between objects and actions guided by the query. CBLN~\cite{liu2021context} addresses TSGV from a new perspective. It reformulates TSGV by scoring all pairs of start and end indices simultaneously and predicting moments with a biaffine structure. Hao \etal~\cite{hao2022cansv} and Liu \etal~\cite{liu2022reducingtv} focus on solving the bias issue via video shuffling and contrastive sample generation, respectively. PPT~\cite{zeng2022point} and VPTSL~\cite{li2022towards} introduce  prompts to jointly model video and text with the unified framework. LocFormer~\cite{rodriguezOpazo2021locformer} designs a multimodal transformer for TSGV in BERT-style. CFSTRI~\cite{qi2021coarsetofine}, MS2P~\cite{shen2022joint} and STDNet~\cite{xu2022stdnet} adopt spatio-temporal cues to interact with query for TSGV. Yang \etal~\cite{yang2022entityaware} and Hao~\cite{hao2022query} decompose the query into multiple semantic phrases to interact with video for boundary prediction. MCA~\cite{fu2022multiple} and PEARL~\cite{zhang2022naturallanguage} further incorporate subtitles of the video to assist the TSGV. EMB~\cite{huang2022videoal} solves uncertainties in TSGV with an elastic moment bounding strategy.

\begin{figure}[t]
    \centering
    \includegraphics[trim={0cm 0cm 0cm 0cm},clip,width=0.9\linewidth]{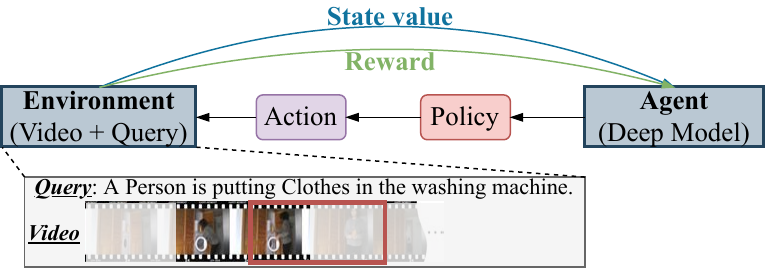}
    \caption{Illustration of sequence decision making formulation in TSGV.}
	\label{fig:seq_decision}
\end{figure}

\subsection{Reinforcement Leaning-based Method}
\label{ssec:rl_based}
From the perspective of proposal usage, reinforcement learning (RL) based methods are also proposal-free methods. However, their task formulation is fundamentally different from the proposal-free methods reviewed earlier. RL-based method formulates TSGV as a sequence decision-making problem and utilizes deep reinforcement learning techniques to solve it. 

Illustrated in Fig.~\ref{fig:seq_decision}, the RL-based method usually maintains a sliding window (the dark red rectangle). The sliding window here is different from that discussed in Section~\ref{ssec:proposal_based}. The RL-based method only adopts a single window, controlled by an agent. An agent, \ie a learnable module, controls the window movement based on a set of handcraft-designed temporal transformations \eg shifting, and scaling.  At each learning step, after each movement, a reward is generated to indicate whether the window is closer or farther away from the target moment. The agent will adapt its action for the next step within pre-defined action space.

\begin{figure}[t]
    \centering
    \includegraphics[trim={0cm 0.2cm 0cm 0cm},clip,width=0.95\linewidth]{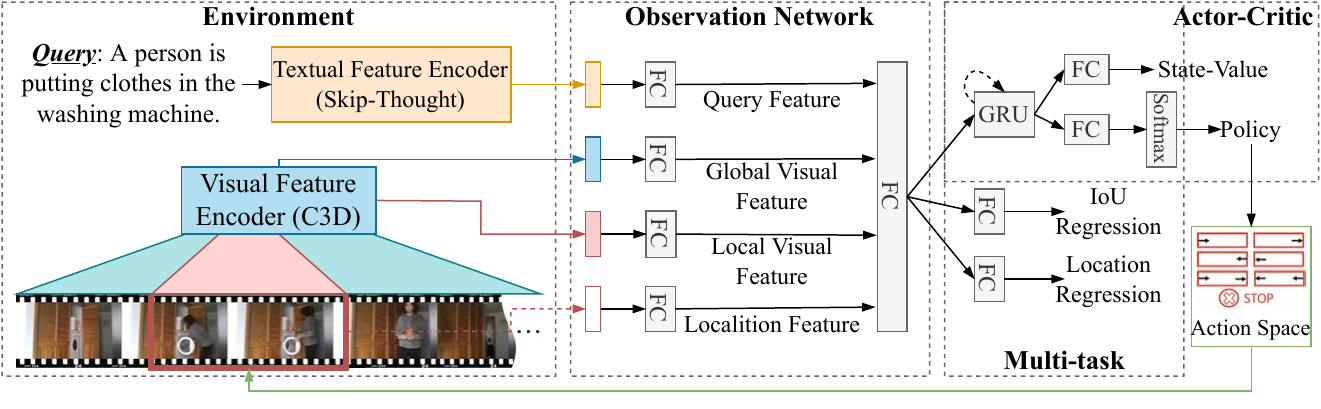}
    \caption{RWM-RL architecture, reproduced from He \etal~\cite{he2019read}.}
	\label{fig:rwm-rl}
\end{figure}

RWM-RL~\cite{he2019read} is one of the first works to define and solve TSGV with an RL framework. Shown in Fig.~\ref{fig:rwm-rl}, it consists of three modules.
The environment module converts the query, global video, and local video segment within the window into corresponding representations. Then the observation network fuses query and video features to output the current state of environment, \ie multimodal representation, at each learning step. In the decision-making module (\ie agent), RWM-RL leverages the actor-critic algorithm~\cite{sutton2018reinforcement} to compute the state-value and an action policy, \ie the probabilistic distribution of all pre-designed actions in the action space. The state-value is used for reward computation, and the action policy determines the movement of the sliding window to adjust the temporal boundaries. RWM-RL defines $7$ actions: moving start/end point ahead/backward by $\delta$ ($4$ scaling actions), shifting both start and end point backward/forward by $\delta$ ($2$ shifting actions), and a STOP action, where $\delta$ denote a basic moving distance. In general, the iterative process ends when encountering the STOP action or reaching the preset maximum number of iteration steps. RWM-RL adopts GRU to model the sequential decision-making process for the actor-critic module. A reward is computed at each step, where the reward is designed to encourage the agent to find a better matching position step by step. All rewards are accumulated for model optimization by utilizing the advantage function~\cite{sutton2018reinforcement} as objective and Monte Carlo sampling~\cite{shapiro2003monte} for policy gradient approximation. To increase the action diversity, RWM-RL further introduces entropy of the policy output as an auxiliary objective following A2-RL~\cite{li2018a2}.

\begin{figure}[t]
    \centering
    \includegraphics[trim={0cm 0cm 0cm 0cm},clip,width=0.8\linewidth]{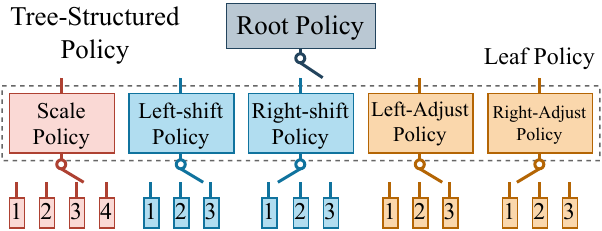}
    \caption{Tree-structured policy, reproduced from Wu \etal~\cite{wu2020tree}.}
	\label{fig:tsp-prl-policy}
\end{figure}

SM-RL~\cite{wang2019language} presents an RNN-based semantic matching RL model to selectively observe proposal candidates produced by a controllable agent. TSP-PRL~\cite{wu2020tree} designs a hierarchical action space with a tree-structured policy, inspired by human's coarse-to-fine decision-making mechanism. The action selection is controlled by a switch over an interface in a tree-structured policy (see Fig.~\ref{fig:tsp-prl-policy}). AVMR~\cite{cao2020adversarial} treats the RL-based module as a generator and devises a Bayesian ranking module as a discriminator to rank proposals. Based on AVMR, Zeng \etal~\cite{zeng2022moment} further deploy continual multi-task learning as the discriminator, which jointly optimizes the ranking and localization subtasks, to boost the performance. STRONG~\cite{cao2020strong} considers appearance and motion features and employs parallel spatial-level and temporal-level RL modules for moment localization. TripNet~\cite{hahn2020tripping} mainly focuses on ameliorating the observation network to boost performance. Instead of using sliding windows, MABAN~\cite{sun2021maban} leverages two individual agents to model start and end points separately. The two agents are conditioned on each other to avoid invalid predictions.

\subsection{Other Supervised Method}
\label{sssec:other_supervised}
In addition to the categories mentioned above, researchers also explore other types of formulations to address TSGV, or under different settings. Shao \etal~\cite{shao2018find} design a unified framework based on TAG~\cite{zhao2017temporal} to perform both video-level retrieval and moment-level localization simultaneously. The two tasks could reinforce each other. Similarly, Jiang \etal~\cite{jiang2022joint} design a cross-task sample transfer to jointly solve video summarization and moment localization. DPIN~\cite{wang2020dual} devises a dual-path interaction network to integrate the benefits of both proposal-based and proposal-free methods. Inspired by Patrick \etal~\cite{patrick2021supportset}, Ding \etal~\cite{ding2021support} propose a support-set based cross-supervision strategy to enhance multimodal interaction, through discriminative contrastive and generative caption objectives. 
Since multiple moments in a video are semantically correlated and temporally coordinated based on their order, several studies~\cite{bao2021dense,shi2021endtoend,jiang2022gtlr,jiang2022semisupervised} explore a novel setting of TSGV, named dense events grounding, which allows jointly localizing multiple moments described in a paragraph, \ie multiple sentences.
SNEAK~\cite{gou2021sneak} studies the adversarial robustness of TSGV models by examining three facets of vulnerabilities, \ie vision, language, and cross-modal interaction, from both attack and defense aspects. Yang \etal~\cite{yang2022videomoment} first explore  neural architecture search for TSGV. Xu \etal~\cite{xu2021boundary} and Cao \etal~\cite{cao2022locvtp} further investigate model pre-training for TSGV, and Xu \etal~\cite{xu2021boundary} construct a large-scale synthesized dataset with annotations and design a boundary-sensitive pretext task. JVTF~\cite{nawaz2022temporal} proposes to solve TSGV by utilizing video question answers as a special variant. Zhang \etal~\cite{zhang2022ama} design a unified framework for VideoQA, TSGV, and VR with global and segment-level alignments. Cao \etal~\cite{cao2021pursuit} reformulate TSGV as a set prediction task and propose a multimodal transformer model inherited from DETR~\cite{carion2020end}, LVTR~\cite{woo2022exploreandmatch} and UMT~\cite{liu2022umt} further improve the DETR-based TSGV framework to boost its performance.

\subsection{Summary of Supervised Method}
\label{sssec:summary_supervised}
We have reviewed different categories of supervised TSGV methods, as well as their advantages and shortcomings. In general, early sliding window-based and proposal-generated methods suffer from low efficiency and flexibility, because of dense and overlapped proposals. These methods also rely on ranking-based loss, making them sensitive to negative samples. Anchor-based methods, another form of the proposal-based solution, learn TSGV in an end-to-end manner. The proposal generation process is  incorporated into the model, abnegating the ineffective SW and PG strategies. Anchor-based methods also enable  contextualized representation learning and fine-grained multimodal interaction. However, the anchor-based methods still need to maintain a mass of proposals during prediction, which hinders model efficiency. 

Proposal-free methods directly learn to predict the boundaries of the target moment, without maintaining any proposals. These methods are more efficient and flexible to adapt to moments with diverse lengths. Nevertheless, compared to proposal-based methods, proposal-free methods overlook the rich information between start and end boundaries and fail to exploit the proposal-level interaction. They also suffer from severe imbalance issues between the positive and negative training samples, \ie only two (start and end) frames are positive in the whole video. Also belonging to a proposal-free category, the design of RL-based methods is intuitive and effective, kind of simulating human's decision-making strategy. However, their performance is unstable due to the difficulty of optimizing RL-based methods.

Despite a vast number of methods in each category, all methods focus on ameliorating cross-modal reasoning, to achieve fine-grained and precise multimodal interaction. Thus, the high-level pipeline of methods in each category is similar in general. Recall that the feature interactor is responsible for understanding the semantic concepts of both query and video and fusing  them to emphasize the video contents that are semantically relevant to the query. In this sense, the quality of the interactor module determines a TSGV model's performance to a great extent.

\subsection{Weakly-supervised TSGV Method}
\label{ssec:weakly_supervised_tsgv}
Supervised learning usually needs a large number of annotations for model training. Annotating temporal boundaries on video with text description is extremely time-consuming and labor-intensive, often not scalable. Furthermore, annotations also suffer from the inaccurate issue, \ie action boundaries in videos are usually subjective and inconsistent across different annotators.

Under the weakly-supervised setting, TSGV methods only need video-query pairs but not the annotations of starting/end time. They explore to find results in a shared multimodal feature space or with a reconstruction-based strategy. In general, the existing weakly-supervised TSGV methods can be roughly grouped into multi-instance learning and reconstruction-based models.

\subsubsection{Multi-Instance Learning Method}
\label{sssec:mil_based}
Multi-instance learning method generally regards the input video as a bag of instances with bag-level annotations. The prediction of instance, \ie proposals, is aggregated as the bag-level prediction.

\begin{figure}[t]
    \centering
    \includegraphics[trim={0cm 0cm 0cm 0cm},clip,width=0.95\linewidth]{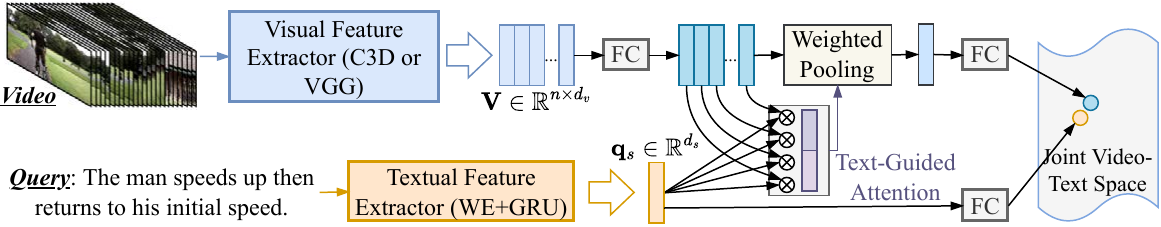}
    \caption{TGA architecture, reproduced from Mithun \etal~\cite{mithun2019weakly}.}
	\label{fig:tga}
\end{figure}

TGA~\cite{mithun2019weakly} first solves TSGV under the multi-instance learning setting. As shown in Fig.~\ref{fig:tga}, TGA first encodes video and query features and presents text-guided attention to learn text-specific global video representations. Then both visual and textual features are projected to a joint space. TGA regards the video and its corresponding query descriptions as positive pairs, while the video with other queries and the query with other videos as negative pairs. TGA learns visual-text alignment at the video level by maximizing the matching scores of  positive samples while minimizing the scores of negative samples.

To achieve good performance, MIL-based methods have to perform precise semantics alignment between video and query. Thus, subsequent solutions~\cite{gao2019wslln,chen2020look,ma2020vlanet,wu2020reinforcement,zhang2020counterfactual,da2021asynce,wang2021visual,wang2021finegrained,huang2021cross,yang2021local,teng2021regularized,wang2021weakly,tan2021logan,chen2022explore,mo2022multiscale,wang2022siamese} mainly focus on devising a sophisticated cross-modal alignment module, designing a robust proposal selection strategy, and/or building effective learning objectives. WSLLN~\cite{gao2019wslln} models alignment and detection modules in parallel to perform proposal selection and video-level alignment simultaneously. VLANet~\cite{ma2020vlanet} designs a surrogate proposal selection module to prune irrelevant proposal candidates. Chen \etal~\cite{chen2020look} and Teng \etal~\cite{teng2021regularized} perform video-query alignment at multiple granularities. CCL~\cite{zhang2020counterfactual}, VCA~\cite{wang2021visual} and MSCL~\cite{mo2022multiscale} introduce contrastive learning mechanisms to effectively distinguish positive and negative (or counterfactual positive) proposals. BAR~\cite{wu2020reinforcement} involves an additional RL module to progressively refine the retrieved proposals. FSAN~\cite{wang2021finegrained}, WSTAN~\cite{wang2021weakly}, and LoGAN~\cite{tan2021logan} focus on mining video and query contents and their correlations. Then they design a fine-grained cross-modal alignment module for accurate moment localization. Da \etal~\cite{da2021asynce} study the uncertain false-positive frame issue, \ie an object might appear sparsely across multiple frames and devise an AsyNCE loss to mitigate the issue by disentangling positive pairs from negative ones. CRM~\cite{huang2021cross} uses a cross-sentence relation mining strategy to explicitly model cross-sentence relations in the paragraph and explore cross-moment relations in the video. LCNet~\cite{yang2021local} further deploys self-supervised cycle consistent loss to guide video-query matching. SAN~\cite{wang2022siamese} designs a multi-scale Siamese module to progressively reduce the semantic gap between the visual and textual modalities. Chen \etal~\cite{chen2022explore} explore the inter-contrast between videos via composition and design a single-stream framework with multi-task learning.

\subsubsection{Reconstruction-based Method}
\label{sssec:reconstruction_based}
Reconstruction-based method tackles TSGV in an indirect way. Methods in this category first take video and query as inputs to produce desired proposals matched to the query. Then the proposals are used to reconstruct the query, where the intermediate proposals are treated as localization results.

\begin{figure}[t]
    \centering
    \includegraphics[trim={0cm 0cm 0cm 0cm},clip,width=0.9\linewidth]{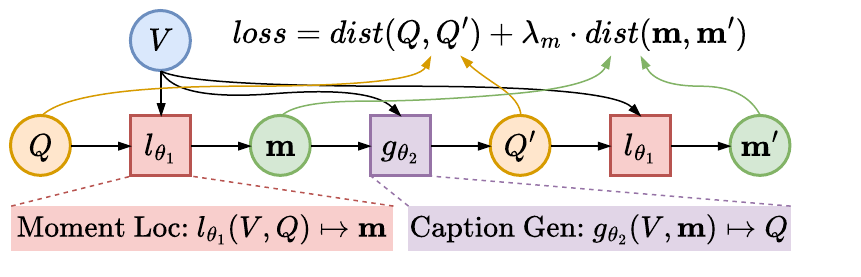}
    \caption{WS-DEC architecture, reproduced from Duan \etal~\cite{duan2018weakly}.}
	\label{fig:ws-dec}
\end{figure}

The idea of reconstruction is first explored by Duan \etal~\cite{duan2018weakly}. They propose a method to solve weakly supervised dense event captioning (WS-DEC), where  moment localization is an auxiliary sub-task to assist model training. The authors indicate that moment localization and event captioning are a pair of dual tasks. Moment localization is to learn a mapping $l_{\theta_{1}}:(V,Q)\mapsto \mathbf{m}$, \ie retrieving a moment $\mathbf{m}$ corresponded to the caption $C_i$ from video $V$. Event captioning inversely generates caption $Q$ for the given $\mathbf{m}$, \ie $g_{\theta_2}:(V,\mathbf{m})\mapsto Q$. Since  $Q$ and $\mathbf{m}$ are a one-to-one correspondence, the dual problems exist simultaneously, and $Q$ and $\mathbf{m}$ are tired together. By nesting the dual functions,  caption-moment pair $(Q,\mathbf{m})$ becomes a fixed-point solution as:
\begin{equation}
        Q  = g_{\theta_2}(V, l_{\theta_1}(V, Q)), \quad
        \mathbf{m}  = l_{\theta_1}(V, g_{\theta_2}(V, \mathbf{m})),
\end{equation}
where $l_{\theta_1}$ and $g_{\theta_2}$ are the localization and captioning modules, respectively. As shown in Fig.~\ref{fig:ws-dec}, WS-DEC first retrieves moment $\mathbf{m}$ by giving video $V$ and caption $Q$; Then the retrieved $\mathbf{m}$ and $V$ are used to reconstruct the caption, denoted by $Q'$; Finally, the reconstructed $Q'$ and $V$ are utilized to relocate the moment $\mathbf{m}'$ again. The objective of WS-DEC is to minimize the distances of $\mathbf{m}$-$\mathbf{m}'$ and $Q$-$Q'$ pairs simultaneously.

\begin{figure}[t]
    \centering
    \includegraphics[trim={0cm 0cm 0cm 0cm},clip,width=0.95\linewidth]{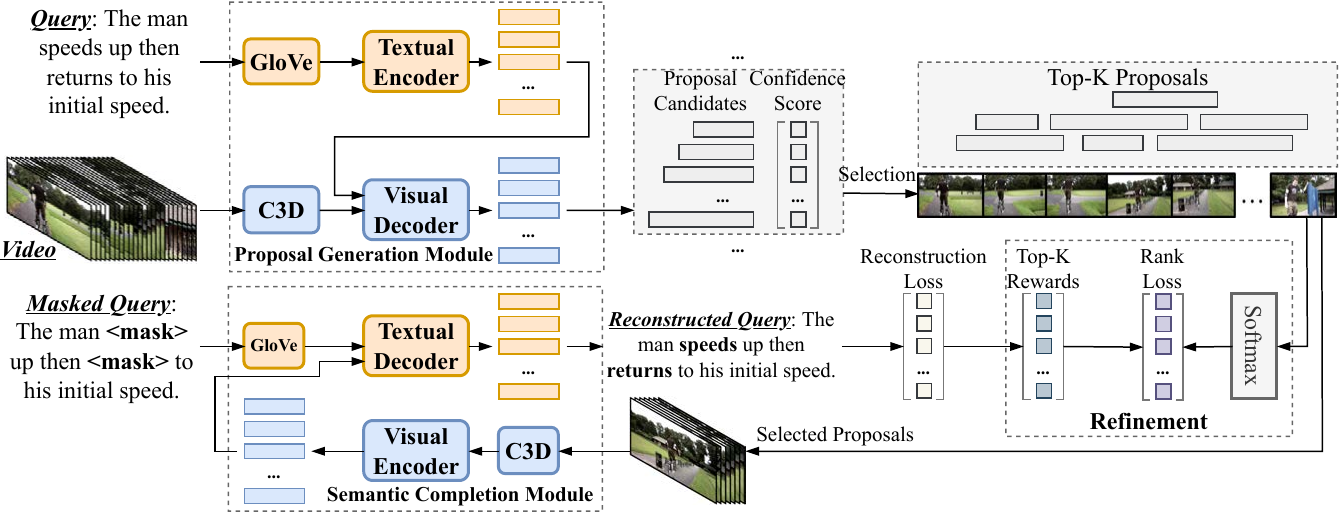}
    \caption{SCN architecture, reproduced from Lin \etal~\cite{lin2020weakly}.}
	\label{fig:scn}
\end{figure}

SCN~\cite{lin2020weakly} adopts a similar idea as WS-DEC. However, SCN is designed for solving weakly supervised TSGV directly; it does not use a specific caption generation module, but switches to reconstruct the masked query. As depicted in Fig.~\ref{fig:scn}, SCN first retrieves a set of proposals from the video. The model then selects top-$K$ proposals as input to reconstruct masked queries, and compute rewards based on reconstruction loss. The rewards further act as feedback to refine proposal generation. CMLNet~\cite{han2022weakly} utilizes a similar structure as SCN and introduces a punishment loss in the candidate generation module. MARN~\cite{song2020weakly} leverages both proposal-level and clip-level video features to produce more accurate proposal candidates. The proposal-level and clip-level features are generated by 2D-Map strategy~\cite{zhang2020learning} and BMN~\cite{lin2019bmn}, respectively. EC-SL~\cite{chen2021towards_cvpr,chen2021towards} improves  WS-DEC by introducing a concept learner and an induced set attention block. Both CPL~\cite{zheng2022weaklysupervised} and CNM~\cite{zheng2022wsvml} introduce contrastive learning into their models.  CPL devises a Gaussian-based contrastive proposal learning module, and CNM explores the contrastive negative sample mining strategy.

\subsubsection{Other Weakly-supervised Method}
\label{sssec:other_weakly_supervised}
In addition to MIL and reconstruction methods, Zhang \etal~\cite{zhang2020regularized} consider both inter- and intra-sample confrontments to address the drawbacks of standard MIL-based methods. The latter generally ignores intra-sample confrontment between moments with semantically similar contents. Luo \etal~\cite{luo2021self} solve the TSGV task in a semi-supervised way. They construct a teacher-student network. The teacher module produces instant pseudo labels for unlabeled samples based on predictions. The student module learns from pseudo labels via self-supervised learning. SVPTR~\cite{jiang2022semisupervised} explores  dense events grounding via contrastive learning under the semi-supervised setting. Nam \etal~\cite{nam2021zero} further propose to learn a TSGV model in zero-shot manner to eliminate the annotation cost. In the zero-shot setting, video-query pairs are not provided. They utilize an off-the-shelf object detector and pseudo-query generation module fine-tuned on RoBERTa~\cite{liu2019roberta} to produce proposals and queries, and simulate the standard TSGV learning. Gao \etal~\cite{gao2021learning} also explores leveraging an off-the-shelf visual concept detector and a pre-trained image-sentence embedding space to perform TSGV without using text annotations on video. Liu \etal~\cite{liu2022unsupervised} design a deep semantic clustering network for unsupervised TSGV. Paul \etal~\cite{paul2022textbased} define the task of localizing novel moments for unseen queries to investigate the ability of TSGV models to novel events. PS-VTG~\cite{xu2022pointsupervised} and ViGA~\cite{cui2022videomr} further explore to utilize single frame/point annotation for TSGV.

\section{Performance Comparison}
\label{sec:evaluation}
We now summarize the reported performance of TSGV methods over the years, by category. Due to the page limit, the  detailed results are listed in supplementary materials.

\begin{table*}[p]
   \small
    \caption{$\text{R@}1, \text{IoU=}m$ of  supervised methods. SW: Sliding Window-based, PG: Proposal Generated, AN: standard Anchor-based, 2D: 2D-Map, RG: Regression-based, SN: Span-based, RL: Reinforcement Learning-based methods.}
    \setlength{\tabcolsep}{2.7pt}
	\centering
	\begin{tabular}{c | c | c | c c c | c c c | c c c | c c c}
		\specialrule{.1em}{.05em}{.05em}
		\multirow{2}{*}{Category} & \multirow{2}{*}{Method} & \multirow{2}{*}{Venue} & \multicolumn{3}{c |}{Charades-STA} & \multicolumn{3}{c |}{ActivityNet Captions} & \multicolumn{3}{c |}{TACoS$_{\text{org}}$} & \multicolumn{3}{c}{TACoS$_{\text{2DTAN}}$} \\
        & & & $m$=0.3 & $m$=0.5 & $m$=0.7 & $m$=0.3 & $m$=0.5 & $m$=0.7 & $m$=0.3 & $m$=0.5 & $m$=0.7 & $m$=0.3 & $m$=0.5 & $m$=0.7 \\
        
        \hline
        
        \multirow{10}{*}{SW} & CTRL~\cite{Gao2017TALLTA} & ICCV'17 & - & 23.63 & 8.89 & - & - & - & 18.32 & 13.30 & - & - & - & - \\
        & {MCN~\cite{hendricks17iccv}} & {ICCV'17} & {13.57} & {4.05} & {-} & {-} & {-} & {-} & {-} & {-} & {-} & {-} & {-} & {-} \\
        & {MCF~\cite{wu2018multi}} & {IJCAI'18} & {-} & {-} & {-} & {-} & {-} & {-} & {18.64} & {12.53} & {-} & {-} & {-} & {-} \\
        & ROLE~\cite{Liu2018CML} & ACM MM'18 & 25.26 & 12.12 & - & - & - & - & - & - & - & - & - & - \\
        & ACRN~\cite{Liu2018AMR} & SIGIR'18 & - & - & - & - & - & - & 19.52 & 14.62 & - & - & - \\
        & SLTA~\cite{jiang2019cross} & ICMR'19 & 38.96 & 22.81 & 8.25 & - & - & - & 17.07 & 11.92 & - & - & - & - \\
        & ACL-K~\cite{ge2019mac} & WACV'19 & 30.48 & 12.20 & - & - & - & - & 24.17 & 20.01 & - & - & - & - \\
        & {ASST~\cite{ning2020asst}} & {TMM'20} & {-} & {37.04} & {18.04} & {-} & {-} & {-} & {-} & {-} & {-} & {-} & {-} & {-} \\
        & MMRG~\cite{zeng2021multi} & CVPR'21 & 71.60 & 44.25 & - & - & - & - & - & - & - & 57.83 & 39.28 & - \\
        & {I2N~\cite{ning2021interaction}} & {TIP'21} & {-} & {56.61} & {34.14} & {-} & {-} & {-} & {-} & {-} & {-} & {31.47} & {29.25} & {-} \\
        & {CAMG~\cite{hu2022camg}} & {ArXiv'22} & {62.10} & {48.33} & {26.53} & {64.58} & {46.68} & {26.64} & {-} & {-} & {-} & {-} & {-} & {-} \\
        
        \hline
        
        \multirow{7}{*}{PG} & QSPN~\cite{Xu2019MultilevelLA} & AAAI'19 & 54.70 & 35.60 & 15.80 & 45.30 & 27.70 & 13.60 & - & - & - & - & - & - \\
        & SAP~\cite{chen2019semantic} & AAAI'19 & - & 27.42 & 13.36 & - & - & - & - & 18.24 & - & - & - & - \\
        & BPNet~\cite{xiao2021boundary} & AAAI'21 & 65.48 & 50.75 & 31.64 & 58.98 & 42.07 & 24.69 & 25.96 & 20.96 & 14.08 & - & - & - \\
        & {LPNet~\cite{xiao2021natural}} & {EMNLP'21} & {66.59} & {54.33} & {34.03} & {64.29} & {45.92} & {25.39} & {-} & {-} & {-} & {-} & {-} & {-} \\
        & APGN~\cite{liu2021adaptive} & EMNLP'21 & - & 62.58 & 38.86 & - & 48.92 & 28.64 & - & - & - & 40.47 & 27.86 & - \\
        & {CMHN~\cite{hu2021video}} & {TIP'21} & {-} & {-} & {-} & {62.49} & {43.47} & {24.02} & {30.04} & {25.58} & {18.44} & {-} & {-} & {-} \\
        & {SLP~\cite{liu2022skimming}} & {ACM MM'22} & {-} & {64.35} & {40.43} & {-} & {52.89} & {32.04} & {-} & {-} & {-} & {42.73} & {32.58} & {-} \\
        
        \hline
        
        \multirow{12}{*}{AN} & TGN~\cite{chen2018temporally} & EMNLP'18 & - & - & - & 45.51 & 28.47 & - & 21.77 & 18.90 & - & - & - & - \\
        & CMIN~\cite{zhu2019cross} & SIGIR'19 & - & - & - & 63.61 & 43.40 & 23.88 & 24.64 & 18.05 & - & - & - & - \\
        & {MAN~\cite{zhang2019man}} & {CVPR'19} & {-} & {46.53} & {22.72} & {-} & {-} & {-} & {-} & {-} & {-} & {-} & {-} & {-} \\
        & SCDM~\cite{yuan2019semantic} & NeurIPS'19 & - & 54.44 & 33.43 & 54.80 & 36.75 & 19.86 & 26.11 & 21.17 & - & - & - & - \\
        & CBP~\cite{Wang2020TemporallyGL} & AAAI'20 & - & 36.80 & 18.87 & 54.30 & 35.76 & 17.80 & 27.31 & 24.79 & 19.10 & - & - & - \\
        & FIAN~\cite{qu2020fine} & ACM MM'20 & - & 58.55 & 37.72 & 64.10 & 47.90 & 29.81 & 33.87 & 28.58 & - & - & - & - \\
        & {CSMGAN~\cite{liu2020jointly}} & {ACM MM'20} & {-} & {-} & {-} & {68.52} & {49.11} & {29.15} & {33.90} & {27.09} & {-} & {-} & {-} & {-} \\
        & {RMN~\cite{liu2020reasoning}} & {COLING'20} & {-} & {59.13} & {36.98} & {67.01} & {47.41} & {27.21} & {32.21} & {25.61} & {-} & {-} & {-} & {-} \\
        & IA-Net~\cite{liu2021progressively} & EMNLP'21 & - & 61.29 & 37.91 & 67.14 & 48.57 & 27.95 & - & - & - & 37.91 & 26.27 & - \\
        & {MIGCN~\cite{zhang2021multimodal}} & {TIP'21} & {-} & {57.10} & {34.54} & {60.03} & {44.94} & {-} & {-} & {-} & {-} & {-} & {-} & {-} \\
        & {DCT-Net~\cite{wang2021dctnet}} & {TIVC'21} & {-} & {-} & {-} & {66.00} & {47.06} & {27.63} & {-} & {-} & {-} & {43.25} & {33.31} & {24.74} \\
        & {MA3SRN~\cite{liu2022exploring1}} & {ArXiv'22} & {-} & {68.98} & {47.79} & {-} & {53.72} & {32.30} & {-} & {-} & {-} & {49.41} & {39.11} & {-} \\
        
        \hline
        
        \multirow{9}{*}{2D} & 2D-TAN~\cite{zhang2020learning} & AAAI'20 & - & 39.81 & 23.31 & 59.45 & 44.51 & 27.38 & - & - & - & 37.29 & 25.32 & - \\
        & MATN~\cite{zhang2021multi} & CVPR'21 & - & - & - & - & 48.02 & 31.78 & - & - & - & 48.79 & 37.57 & - \\
        & SMIN~\cite{wang2021structured} & CVPR'21 & - & 64.06 & 40.75 & - & 48.46 & 30.34 & - & - & - & 48.01 & 35.24 & - \\
        & {PLN~\cite{zheng2021progressive}} & {ArXiv'21} & {68.60} & {56.02} & {35.16} & {59.65} & {45.66} & {29.28} & {-} & {-} & {-} & {43.89} & {31.12} & {-} \\
        & RaNet~\cite{gao2021relation} & EMNLP'21 & - & 60.40 & 39.65 & - & 45.59 & 28.67 & - & - & - & 43.34 & 33.54 & - \\
        & FVMR~\cite{gao2021fast} & ICCV'21 & - & 55.01 & 33.74 & 60.63 & 45.00 & 26.85 & - & - & - & 41.48 & 29.12 & - \\
        & {VLG-Net~\cite{soldan2021vlg}} & {ICCV'21} & {-} & {-} & {-} & {-} & {46.32} & {29.82} & {-} & {-} & {-} & {45.46} & {34.19} & {-} \\
        & CLEAR~\cite{hu2021coarse} & TIP'21 & - & - & - & 59.96 & 45.33 & 28.05 & - & - & - & 42.18 & 30.27 & 15.54 \\
        & {Sun \etal~\cite{sun2022yount}} & {SIGIR'22} & {-} & {60.82} & {41.16} & {-} & {47.92} & {30.47} & {-} & {-} & {-} & {48.81} & {36.74} & {-} \\
        
        \hline
        
        \multirow{9}{*}{RG} & ABLR~\cite{yuan2019to} & AAAI'19 & - & - & - & 55.67 & 36.79 & - & 19.50 & 9.40 & - & - & - & - \\
        & ExCL~\cite{ghosh2019excl} & NAACL'19 & 61.50 & 44.10 & 22.40 & 63.00 & 43.60 & 24.10 & 45.50 & 28.00 & 13.80 & - & - & - \\
        & DEBUG~\cite{lu2019debug} & EMNLP'19 & 54.95 & 37.39 & 17.92 & 55.91 & 39.72 & - & 23.45 & - & - & - & - & - \\
        & {GDP~\cite{chen2020rethinking}} & {AAAI'20} & {54.54} & {39.47} & {18.49} & {56.17} & {39.27} & {-} & {24.14} & {-} & {-} \\
        & DRN~\cite{zeng2020dense} & CVPR'20 & - & 53.09 & 31.75 & - & 45.45 & 24.36 & - & 23.17 & - & - & - & - \\
        & LGI~\cite{mun2020local} & CVPR'20 & 72.96 & 59.46 & 35.48 & 58.52 & 41.51 & 23.07 & - & - & - & - & - & - \\
        & CPNet~\cite{li2021proposal} & AAAI'21 & - & 60.27 & 38.74 & - & 40.56 & 21.63 & - & - & - & 42.61 & 28.29 & - \\
        & {HiSA~\cite{xu2022hisa}} & {TIP'22} & {74.84} & {61.10} & {39.70} & {64.58} & {45.36} & {27.68} & {-} & {-} & {-} & {53.31} & {42.14} & {29.32} \\

        \hline
        
        \multirow{9}{*}{SN} & VSLNet~\cite{zhang2020vslnet} & ACL'20 & 70.46 & 54.19 & 35.22 & 63.16 & 43.22 & 26.16 & 29.61 & 24.27 & 20.03 & - & - & - \\
        & CPN~\cite{zhao2021cascaded} & CVPR'21 & 75.53 & 59.77 & 36.67 & 62.81 & 45.10 & 28.10 & - & - & - & 48.29 & 36.58 & 21.58 \\
        & CI-MHA~\cite{yu2021cross} & SIGIR'21 & 69.87 & 54.68 & 35.27 & 61.49 & 43.97 & 25.13 & - & - & - & - & - & - \\
        & {IVG-DCL~\cite{nan2021interventional}} & {CVPR'21} & {67.63} & {50.24} & {32.88} & {63.22} & {43.84} & {27.10} & {38.84} & {29.07} & {19.05} & {-} & {-} & {-} \\
        & SeqPAN~\cite{zhang2021parallel} & ACL'21 & 73.84 & 60.86 & 41.34 & 61.65 & 45.50 & 28.37 & 31.72 & 27.19 & 21.65 & 48.64 & 39.64 & 28.07 \\
        & ACRM~\cite{tang2021frame} & TMM'21 & 73.47 & 57.93 & 38.33 & - & - & - & - & - & - & 51.26 & 39.34 & 26.94 \\
        & {ABDIN~\cite{zhang2021temporal}} & {TMM'21} & {-} & {-} & {-} & {63.19} & {44.02} & {24.23} & {23.63} & {20.16} & {-} & {-} & {-} & {-} \\
        & VSLNet-L~\cite{zhang2021natural} & TPAMI'21 & 70.46 & 54.19 & 35.22 & 62.35 & 43.86 & 27.51 & 32.04 & 27.92 & 23.28 & 47.66 & 36.34 & 26.42 \\
        & {PEARL~\cite{zhang2022natural}} & {WACV'22} & {71.90} & {53.50} & {35.40} & {-} & {-} & {-} & {-} & {-} & {-} & {42.94} & {32.07} & {18.37} \\
        
        \hline
        
        \multirow{6}{*}{RL} & RWM-RL~\cite{he2019read} & AAAI'19 & - & 36.70 & - & - & 36.90 & - & - & - & - & - & - & - \\
        & SM-RL~\cite{wang2019language} & CVPR'19 & - & 24.36 & 11.17 & - & - & - & 20.25 & 15.95 & - & - & - & - \\
        & TSP-PRL~\cite{wu2020tree} & AAAI'20 & - & 45.45 & 24.75 & 56.02 & 38.82 & - & - & - & - & - & - & - \\
        & TripNet~\cite{hahn2020tripping} & BMVC'20 & 54.64 & 38.29 & 16.07 & 48.42 & 32.19 & 13.93 & 23.95 & 19.17 & 9.52 & - & - & - \\
        & MBAN~\cite{sun2021maban} & TIP'21 & - & 56.29 & 32.26 & - & 42.42 & 24.34 & - & - & - & - & - & - \\
        & {URL~\cite{zeng2022moment}} & {TMCCA'22} & {77.88} & {55.69} & {-} & {-} & {76.88} & {50.11} & {-} & {-} & {-} & {73.26} & {50.53} & {-} \\
        
        \hline 
        
        \multirow{5}{*}{Other} & DPIN~\cite{wang2020dual} & ACM MM'20 & - & 47.98 & 26.96 & 62.40 & 47.27 & 28.31 & - & - & - & 46.74 & 32.92 & - \\
        & {DepNet~\cite{bao2021dense}} & {AAAI'21} & {-} & {-} & {-} & {72.81} & {55.91} & {33.46} & {-} & {-} & {-} & {41.34} & {27.16} & {-} \\
        & GTR~\cite{cao2021pursuit} & EMNLP'21 & - & 62.58 & 39.68 & - & 50.57 & 29.11 & - & - & - & 40.39 & 30.22 & - \\
        & BSP~\cite{xu2021boundary} & ICCV'21 & 68.76 & 53.63 & 29.27 & - & - & - & - & - & - & - & - & - \\
        & {CMAS~\cite{yang2022videomoment}} & {TIP'22} & {-} & {48.37} & {29.44} & {-} & {46.23} & {29.48} & {-} & {-} & {-} & {31.37} & {16.85} & {-} \\
        
        \specialrule{.1em}{.05em}{.05em}
	\end{tabular}
	\label{tab:all_supervised_result}
\end{table*}

\begin{table}
   \small
    \caption{Results of supervised methods on DiDeMo dataset.}
    \setlength{\tabcolsep}{2.5pt}
	\centering
	\begin{tabular}{c | c | c | c c c | c}
		\specialrule{.1em}{.05em}{.05em}
		\multirow{2}{*}{Category} & \multirow{2}{*}{Method} & \multirow{2}{*}{Venue} & \multicolumn{3}{c |}{$\text{R@}1, \text{IoU=}m$} & \multirow{2}{*}{mIoU} \\
        & & & 0.5 & 0.7 & 1.0 & \\
        \hline
        \multirow{7}{*}{SW} & MCN~\cite{hendricks17iccv} & ICCV'17 & - & - & 28.10 & 41.08 \\
        & MLLC~\cite{hendricks2018localizing} & EMNLP'18 & - & - & 27.46 & 41.20 \\
        & ROLE~\cite{Liu2018CML} & ACM MM'18 & 29.40 & 15.68 & - & - \\
        & ACRN~\cite{Liu2018AMR} & SIGIR'18 & 27.44 & 16.65 & - & - \\
        & SLTA~\cite{jiang2019cross} & ICMR'19 & 30.92 & 17.16 & - & - \\
        & {TCMN~\cite{zhang2019exploiting}} & {ACM MM'19} & {-} & {-} & {28.90} & {41.03} \\
        & {ASST~\cite{ning2020asst}} & {TMM'20} & {-} & {-} & {32.38} & {47.49} \\
        & {I2N~\cite{ning2021interaction}} & {TIP'21} & {-} & {-} & {29.00} & {44.32} \\
        \hline
        {PG} & {EFRC~\cite{Xu2018TexttoClipVR}} & {ArXiv'18} & {11.9} & {5.5} & {13.23} & {27.57} \\
        \hline
        \multirow{2}{*}{AN} & TGN~\cite{chen2018temporally} & EMNLP'18 & - & - & 28.23 & 42.97 \\
        & MAN~\cite{zhang2019man} & CVPR'19 & - & - & 27.02 & 41.16 \\
        \hline
        \multirow{2}{*}{2D} & {TMN~\cite{liu2018temporal}} & {ECCV'18} & {-} & {-} & {22.92} & {35.17} \\
        & VLG-Net~\cite{soldan2021vlg} & ICCV'21 & 33.35 & 25.57 & 25.57 & - \\
        \hline 
        {SN} & {L-Net~\cite{chen2019localizing}} & {AAAI'19} & {-} & {-} & {-} & {41.43} \\
        \hline
        RL & SM-RL~\cite{wang2019language} & CVPR'19 & - & - & 31.06 & 43.94 \\
        \specialrule{.1em}{.05em}{.05em}
	\end{tabular}
	\label{tab:didemo_supervised_result}
\end{table}

\begin{table*}
   \small
    \caption{Results of weakly-supervised methods, where MIL is Multi-Instance Learning-based method, REC denotes Reconstruction-based method, and * denotes the zero-shot setting.}
    \setlength{\tabcolsep}{5pt}
	\centering
	\begin{tabular}{c | c | c | c c c | c c c c | c c c}
		\specialrule{.1em}{.05em}{.05em}
		\multirow{3}{*}{Category} & \multirow{3}{*}{Method} & \multirow{3}{*}{Venue} & \multicolumn{3}{c |}{Charades-STA} & \multicolumn{4}{c |}{ActivityNet Captions} & \multicolumn{3}{c}{DiDeMo} \\
		& & & \multicolumn{3}{c |}{$\text{R@}1, \text{IoU=}m$} & \multicolumn{4}{c |}{$\text{R@}1, \text{IoU=}m$} & \multicolumn{2}{c}{$\text{R@}n, \text{IoU=}1.0$} & \multirow{2}{*}{mIoU} \\
        & & & $m$=0.3 & $m$=0.5 & $m$=0.7 & $m$=0.1 & $m$=0.3 & $m$=0.5 & $m$=0.7 & $n$=1 & $n$=5 & \\
        
        \hline
        
        \multirow{17}{*}{MIL} & TGA~\cite{mithun2019weakly} & CVPR'19 & 32.14 & 19.94 & 8.84 & - & - & - & - & 12.19 & 39.74 & 24.92 \\
        & WSLLN~\cite{gao2019wslln} & EMNLP'19 & - & - & - & 75.40 & 42.80 & 22.70 & - & 19.40 & 54.40 & 27.40 \\
        & {Chen \etal~\cite{chen2020look}} & {ArXiv'20} & {39.80} & {27.30} & {12.90} & {74.20} & {44.30} & {23.60} & {-} & {-} & {-} & {-} \\
        & VLANet~\cite{ma2020vlanet} & ECCV'20 & 45.24 & 31.83 & 14.17 & - & - & - & - & 19.32 & 65.68 & 25.33 \\
        & CCL~\cite{zhang2020counterfactual} & NeurIPS'20 & - & 33.21 & 15.68 & - & 50.12 & 31.07 & - & - & - & - \\
        & {BAR~\cite{wu2020reinforcement}} & {ACM MM'20} & {51.64} & {33.98} & {15.97} & {-} & {53.41} & {33.12} & {-} & {-} & {-} & {-} \\
        & {MS-2DTN~\cite{Li2021Multiscale2R}} & {ICPR'21} & {-} & {30.38} & {17.31} & {-} & {49.79} & {29.68} & {-} & {-} & {-} & {-} \\
        & LoGAN~\cite{tan2021logan} & WACV'21 & 51.67 & 34.68 & 14.54 & - & - & - & - & 39.20 & 64.04 & 38.28 \\
        & VCA~\cite{wang2021visual} & ACM MM'21 & 58.58 & 38.13 & 19.57 & 67.96 & 50.45 & 31.00 & - & - & - & - \\
        & FSAN~\cite{wang2021finegrained} & EMNLP'21 & - & - & - & 78.45 & 55.11 & 29.43 & - & 19.40 & 57.85 & 31.92 \\
        & CRM~\cite{huang2021cross} & ICCV'21 & 53.66 & 34.76 & 16.37 & 81.61 & 55.26 & 32.19 & - & - & - & - \\
        & LCNet~\cite{yang2021local} & TIP'21 & 59.60 & 39.19 & 18.87 & 78.58 & 48.49 & 26.33 & - & - & - & - \\
        & WSTAN~\cite{wang2021weakly} & TMM'21 & 43.39 & 29.35 & 12.28 & 79.78 & 52.45 & 30.01 & - & 19.40 & 54.64 & 31.94 \\
        & {Teng \etal~\cite{teng2021regularized}} & {TMM'21} & {-} & {-} & {-} & {65.99} & {44.49} & {24.33} & {-} & {17.00} & {64.80} & {29.59} \\
        & {Chen \etal~\cite{chen2022explore}} & {AAAI'22} & {43.31} & {31.02} & {16.53} & {71.86} & {46.62} & {29.52} & {-} & {-} & {-} & {-} \\
        & {MSCL~\cite{mo2022multiscale}} & {ArXiv'22} & {58.92} & {43.15} & {23.49} & {75.61} & {55.05} & {38.23} & {-} & {-} & {-} & {-} \\
        & {SAN~\cite{wang2022siamese}} & {TMM'22} & {51.02} & {31.02} & {13.12} & {-} & {48.44} & {30.54} & {13.85} & {-} & {-} & {-} \\
        
        \hline 
        
        \multirow{7}{*}{REC} & WS-DEC~\cite{duan2018weakly} & NeurIPS'18 & - & - & - & 62.71 & 41.98 & 23.34 & - & - & - & - \\
        & SCN~\cite{lin2020weakly} & AAAI'20 & 42.96 & 23.58 & 9.97 & 71.48 & 47.23 & 29.22 & - & - & - & - \\
        & MARN~\cite{song2020weakly} & ArXiv'20 & 48.55 & 31.94 & 14.81 & - & 47.01 & 29.95 & - & - & - & - \\
        & EC-SL~\cite{chen2021towards_cvpr} & CVPR'21 & - & - & - & 68.48 & 44.29 & 24.16 & - & - & - & - \\
        & {CMLNet~\cite{han2022weakly}} & {TIVC'22} & {48.99} & {11.24} & {-} & {84.08} & {49.39} & {22.58} & {-} & {-} & {-} & {-} \\
        & {CNM~\cite{zheng2022wsvml}} & {AAAI'22} & {60.39} & {35.43} & {15.45} & {78.13} & {55.68} & {33.33} & {-} & {-} & {-} & {-} \\
        & {CPL~\cite{zheng2022weaklysupervised}} & {CVPR'22} & {65.99} & {49.05} & {22.61} & {71.23} & {50.07} & {30.14} & {-} & {-} & {-} & {-} \\
        
        \hline
        
        \multirow{7}{*}{Other} & RTBPN~\cite{zhang2020regularized} & ACM MM'20 & 60.04 & 32.36 & 13.24 & 73.73 & 49.77 & 29.63 & - & 20.79 & 60.26 & 29.81 \\
        & U-VMR~\cite{gao2021learning} & TCSVT'21 & 46.69 & 20.14 & 8.27 & 69.63 & 46.15 & 26.38 & 11.64 & - & - & - \\
        & PSVL~\cite{nam2021zero}* & ICCV'21 & 46.47 & 31.29 & 14.17 & - & 44.74 & 30.08 & 14.74 & - & - & - \\
        & {PS-VTG~\cite{xu2022pointsupervised}} & {TMM'22} & {60.40} & {39.22} & {20.17} & {-} & {59.71} & {39.59} & {21.98} & {-} & {-} & {-} \\
        & {SVPTR~\cite{jiang2022semisupervised}} & {CVPR'22} & {55.14} & {32.44} & {15.53} & {-} & {78.07} & {61.70} & {38.36} & {-} & {-} & {-} \\
        & {DSCNet~\cite{liu2022unsupervised}} & {AAAI'22} & {44.15} & {28.73} & {14.67} & {-} & {47.29} & {28.16} & {-} & {-} & {-} & {-} \\
        & {ViGA~\cite{cui2022videomr}} & {SIGIR'22} & {71.21} & {45.05} & {20.27} & {-} & {59.61} & {35.79} & {16.96} & {-} & {-} & {-} \\
        
        \specialrule{.1em}{.05em}{.05em}
	\end{tabular}
	\label{tab:all_weakly_result}
\end{table*}

\begin{table}
   \small
    \caption{Result of different visual features on Charades-STA.}
    \setlength{\tabcolsep}{2pt}
	\centering
	\begin{tabular}{c | c | c | c | c c c}
		\specialrule{.1em}{.05em}{.05em}
		\multirow{2}{*}{Method} & \multirow{2}{*}{Category} & \multirow{2}{*}{Venue} & \multirow{2}{*}{Feature} & \multicolumn{3}{c}{$\text{R@}1, \text{IoU=}m$} \\
        & & & & $m$=0.3 & $m$=0.5 & $m$=0.7 \\
        
        \hline
        
        \multirow{3}{*}{DRN~\cite{zeng2020dense}} & \multirow{3}{*}{RG} & \multirow{3}{*}{CVPR'20} & VGG & - & 42.90 & 23.68 \\
        & & & C3D & - & 45.40 & 26.40 \\
        & & & I3D & - & 53.09 & 31.75 \\
        
        \hline
        
        \multirow{2}{*}{CPNet~\cite{li2021proposal}} & \multirow{2}{*}{RG} & \multirow{2}{*}{AAAI'21} & C3D & - & 40.32 & 22.47 \\
        & & & I3D & - & 60.27 & 38.74 \\
        
        \hline
        
        \multirow{2}{*}{BPNet~\cite{xiao2021boundary}} & \multirow{2}{*}{PG} & \multirow{2}{*}{AAAI'21} & C3D & 55.46 & 38.25 & 20.51 \\
        & & & I3D & 65.48 & 50.75 & 31.64 \\
        
        \hline
        
        \multirow{2}{*}{LPNet~\cite{xiao2021natural}} & \multirow{2}{*}{PG} & \multirow{2}{*}{EMNLP'21} & C3D & 59.14 & 40.94 & 21.13 \\
        & & & I3D & 66.59 & 54.33 & 34.03 \\
        
        \hline

        \multirow{2}{*}{RaNet~\cite{gao2021relation}} & \multirow{2}{*}{2D} & \multirow{2}{*}{EMNLP'21} & VGG & - & 43.87 & 26.83 \\
        & & & I3D & - & 60.40 & 39.65 \\

        \hline

        \multirow{3}{*}{FVMR~\cite{gao2021fast}} & \multirow{3}{*}{2D} & \multirow{3}{*}{ICCV'21} & VGG & - & 42.36 & 24.14 \\
        & & & C3D & - & 38.16 & 18.22 \\
        & & & I3D & - & 55.01 & 33.74 \\

        \hline

        \multirow{3}{*}{HDRR~\cite{ma2021hierarchical}} & \multirow{3}{*}{AN} & \multirow{3}{*}{ACMMM'21} & C3D & 62.37 & 43.04 & 21.32 \\
        & & & TS & 68.33 & 54.06 & 27.31 \\
        & & & I3D & 73.44 & 59.46 & 34.11 \\

        \hline

        \multirow{3}{*}{{MIGCN~\cite{zhang2021multimodal}}} & \multirow{3}{*}{{AN}} & \multirow{3}{*}{{TIP'21}} & {C3D} & {-} & {42.26} & {22.04} \\
        & & & {TS} & {-} & {51.80} & {29.33} \\
        & & & {I3D} & {-} & {57.10} & {34.54} \\
        
        \specialrule{.1em}{.05em}{.05em}
	\end{tabular}
	\label{tab:charades_feat_result}
\end{table}

\begin{table}
   \small
    \caption{\small Results of DRFT with different visual modality features on Charades-STA and ActivityNet Captions datasets, where R, F and D denote RGB, flow, and depth modalities, respectively.}
    \setlength{\tabcolsep}{3pt}
	\centering
	\begin{tabular}{c | c | c | c c c | c}
		\specialrule{.1em}{.05em}{.05em}
		\multirow{2}{*}{Method} & \multirow{2}{*}{Venue} & \multirow{2}{*}{Feature} & \multicolumn{3}{c |}{$\text{R@}1, \text{IoU=}m$} & \multirow{2}{*}{mIoU} \\
        & & & $m$=0.3 & $m$=0.5 & $m$=0.7 & \\
        
        \hline
        
        \multirow{8}{*}{DRFT~\cite{chen2021end}} & \multirow{8}{*}{NeurIPS'21} & \multicolumn{5}{c}{Charades-STA} \\
        
        \cline{3-7}
        
        & & R & 73.85 & 60.79 & 36.72 & 52.64 \\
        & & R+F & 74.26 & 61.93 & 38.69 & 53.92 \\
        & & R+F+D & 76.68 & 63.03 & 40.15 & 54.89 \\
        
        \cline{3-7}
        
        & & \multicolumn{5}{c}{ActivityNet Captions} \\
        
        \cline{3-7}
        
        & & R & 60.25 & 42.37 & 25.23 & 43.18 \\
        & & R+F & 61.80 & 43.71 & 26.43 & 44.82 \\
        & & R+F+D & 62.91 & 45.72 & 27.79 & 45.86 \\
        
        \specialrule{.1em}{.05em}{.05em}
	\end{tabular}
	\label{tab:drft_result}
\end{table}

\begin{table}
   \small
    \caption{\small Result of PMI with different modality features on the ActivityNet Captions. IRV2 is Inception-ResNet v2~\cite{szegedy2017inception} visual feature and A is SoundNet~\cite{aytar2016soundnet} audio feature.}
    \setlength{\tabcolsep}{3.5pt}
	\centering
	\begin{tabular}{c | c | c | c c c}
		\specialrule{.1em}{.05em}{.05em}
		\multirow{2}{*}{Method} & \multirow{2}{*}{Venue} & \multirow{2}{*}{Feature} & \multicolumn{3}{c}{$\text{R@}1, \text{IoU=}m$} \\
        & & & $m$=0.3 & $m$=0.5 & $m$=0.7 \\
        
        \hline
        
        \multirow{3}{*}{PMI~\cite{chen2020learning}} & \multirow{3}{*}{ECCV'20} & C3D & 59.69 & 38.28 & 17.83 \\
        & & C3D+IRV2 & 60.16 & 39.16 & 18.02 \\
        & & C3D+IRV2+A & 61.22 & 40.07 & 18.29 \\
        
        \specialrule{.1em}{.05em}{.05em}
	\end{tabular}
	\label{tab:pmi_result}
\end{table}

\smallskip \noindent \textbf{Performance Overview.}
For supervised methods, as summarized in Table~\ref{tab:all_supervised_result} and Table~\ref{tab:didemo_supervised_result}, in general, anchor-based (ANchor and 2D-map) and proposal-free (ReGression and SpaN) methods are superior to sliding window-based (SW) and proposal-generated (PG) methods. Within the SW category, MMRG~\cite{zeng2021multi} introduces a graph structure to model the visual-textual relations and adds a boundary regression auxiliary objective to guide moment retrieval, outperforming early SW methods by a large margin. A similar observation holds in the PG category. Compared to early anchor-based and proposal-free work, recent methods incorporate more sophisticated multimodal interaction strategies to refine the cross-modal reasoning between video and query. They also introduce various auxiliary objectives to enhance the feature representation learning and steer the model for more precise moment localization. In the RL category, recent solutions mainly focus on designing more powerful agents or refined action space (policy) to achieve accurate sequence decisions~\cite{wu2020tree,sun2021maban}. Despite the improvements of recent RL-based methods, the performance gap between RL-based methods and anchor-based/proposal-free methods remains distinct. Two possible reasons for the inferior results are: (i) the RL learning process is not very stable, and (ii) the multimodal interaction between the two modalities is not fully exploited in RL methods. Among other methods, GTR~\cite{cao2021pursuit} and BSP~\cite{xu2021boundary} provide a new perspective to solve TSGV. BSP proposes a pre-training paradigm for TSGV by designing a boundary-sensitive pretext task and collecting a synthesized dataset with temporal boundaries. GTR builds an end-to-end framework to learn TSGV from raw videos directly. Although their results are slightly inferior to other solutions, both open up new directions for TSGV.

For weakly-supervised methods (Table~\ref{tab:all_weakly_result}), in general, MIL-based methods are superior to reconstruction-based methods. Other than cross-modal reasoning, the learning objective also plays a key role in MIL-based methods. Recent solutions adopt more effective strategies or introduce auxiliary objectives, such as contrastive learning~\cite{zhang2020counterfactual}, pseudo supervision~\cite{gao2019wslln,wang2021weakly}, and boundary adjustment~\cite{chen2020look}. For other methods, PSVL~\cite{nam2021zero} solves TSGV under the zero-shot setting, which assumes that the video-query pairs are inaccessible, \ie only the text corpora and video collection are given. Although the zero-shot setting is more challenging, arguably the setting is closer to real-world scenarios.

\smallskip \noindent \textbf{Impact of Features.} Improvements in model performance may come from various sources. In particular, various visual (\eg VGG~\cite{simonyan2014very}, ResNet~\cite{szegedy2017inception} C3D~\cite{tran2015learning}, I3D~\cite{carreira2017quo}) and textual (\eg GloVe~\cite{pennington2014glove}, BERT~\cite{devlin2019bert}) feature extractors have been utilized in different models. Among visual feature extractors, VGG~\cite{simonyan2014very} and ResNet~\cite{szegedy2017inception} are pre-trained on image datasets, and they are more effective in extracting appearance features like objects and visual concepts from video frames. In contrast, C3D~\cite{tran2015learning} and I3D~\cite{carreira2017quo} are pre-trained on video action recognition datasets, for extracting  motion features like actions or activities, from video snippets or segments. In general, feature analysis conducted by different methods on the Charades-STA dataset shows that VGG$<$C3D$<$I3D, with respect to model performance (able~\ref{tab:charades_feat_result}). Because TSGV is mainly for activity retrieval, video-based feature extractors are more effective than their image-based counterparts. I3D contains a more sophisticated structure and is trained on larger datasets than C3D, leading to a more powerful representation ability.

Visual features are mostly extracted from RGB frames. Chen \etal~\cite{chen2021end} explore incorporating optical flow and depth map information in frames as complementary visual features. Optical flow focuses on large motion, and depth maps reflect the scene configuration when the action is related to objects recognizable by their shapes. As shown in Table~\ref{tab:drft_result}, prominent improvements are obtained by adding more visual modality features on Charades-STA and ActivityNet Captions. A query may contain descriptions of both objects and actions. Thus, some methods~\cite{chen2020learning,rodriguez2021dori} exploit both appearance and motion features to represent a video. In addition to the motion features extracted by C3D model, Chen \etal~\cite{chen2020learning} introduce appearance features by IRV2~\cite{szegedy2017inception} and audio features from video by SoundNet~\cite{aytar2016soundnet}. In general, as summarized in Table~\ref{tab:pmi_result}, improving the feature extractor or exploiting more diverse features leads to better accuracy.

Moreover, we also conduct the efficiency comparison among different method categories by selecting one representative model from each category. The results and discussion are presented in Section~\ref{appd:ssec:efficiency} in the Appendix.

\section{Challenges and Future Directions}
\label{sec:challenges_directions}

\subsection{Critical Analysis}
\label{ssec:challenges}

\begin{figure}[t]
    \centering
    \includegraphics[trim={0cm 0cm 0cm 0cm},clip,width=0.8\linewidth]{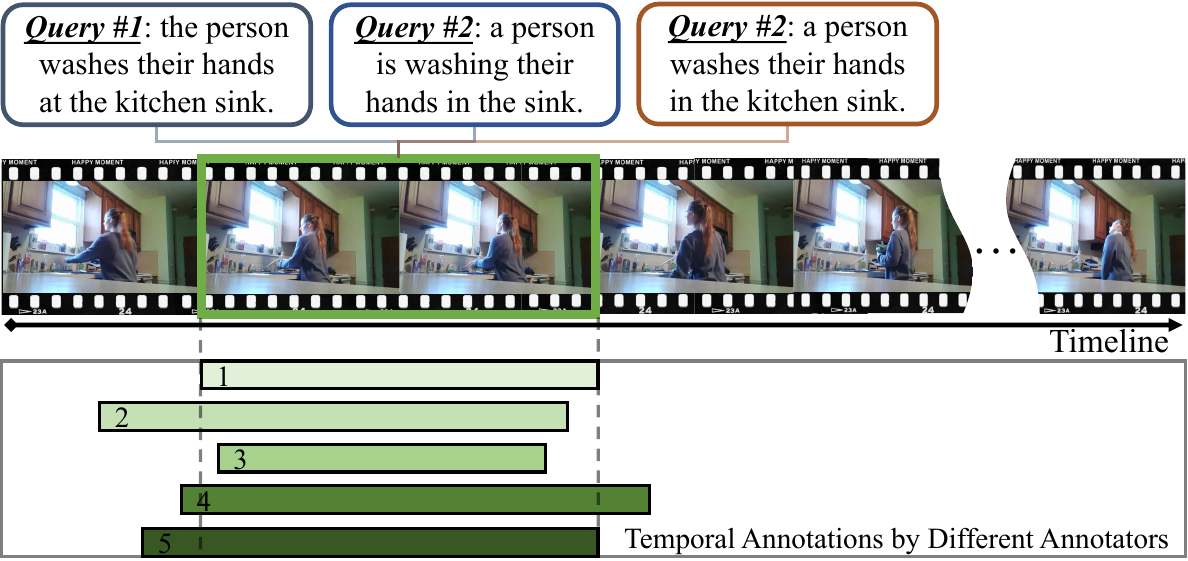}
    \caption{Illustration of data uncertainty in TSGV benchmarks. Annotation uncertainty means disagreements of annotated ground-truth moment across different annotators. Query uncertainty means various query expressions for one ground-truth moment.}
	\label{fig:data_uncertainty}
\end{figure}

\smallskip \noindent \textbf{Data Uncertainty.} Recent studies~\cite{zhou2021embracing,otani2020challengesmr} observe that data samples among current benchmark datasets are ambiguous and inconsistent. First, the annotated moments of a query may be discrepant across annotators, \ie \textit{annotation uncertainty}~\cite{otani2020challengesmr}. As shown in Fig.~\ref{fig:data_uncertainty}, for the same query \#1, temporal boundaries annotated on the same video by different annotators are different. Such an issue is inevitable due to the subjectivity of annotators. Second, multiple queries may be used to describe the same event/moment, \ie \textit{query uncertainty}. Fig.~\ref{fig:data_uncertainty} also shows that the three queries are attached to the same moment.

For annotation uncertainty, existing methods usually apply single-style annotations, \ie each data sample is labeled by one annotator, because of the potentially expensive cost of multiple labeling. The inherent uncertainty in moment localization is ignored. Consequently, models may capture single-style prediction bias during training, leading to inferior generalization performance. For query uncertainty, similarly,  methods only take one query as input for a moment and encode the  query as a deterministic vector. In this case, the variety of query expressions cannot be learned by the model. The model may not well handle queries in different expressions for the same event.

To mitigate annotation uncertainty, Otani \etal~\cite{otani2020challengesmr} propose to re-annotate the Charades-STA and ActivityNet Captions datasets on Amazon Mechanical Turk. They also present two alternative evaluation metrics by considering the issues of multiple ground truths and potential miss-labeled samples. Specifically, the first metric evaluates the predicted moments with respect to the nearest-neighbor reference, which is based on the fact that a video may have multiple positive moments for a single query sentence. When a predicted moment is close to at least one of the reference moments, \ie its IoU with reference moment is larger than a threshold, it is counted as positive. The second metric considers the reliability of human annotations. When a reference moment largely overlaps with the majority of other reference moments, the reference moment is more reliable. A reference moment that is different from others is likely miss-labeled. Zhou \etal~\cite{zhou2021embracing} address both annotation and query uncertainties from a model design perspective. For query uncertainty, they introduce a decoupling method to disentangle each query into a relation feature and a modified feature. The relation feature encodes the discriminative and consistent information; the modified feature encodes the personalized information. Then the modified feature is encoded as Gaussian distribution and a sampling operation is adopted in the latent space to obtain multiple query representations. For annotation uncertainty, they propose a debias mechanism modified from multiple choice learning to generate diverse predictions. Recently, Zhou \etal~\cite{zhou2022thinking} devise a framework to achieve diverse moment localization with only single-label annotations, by constructing soft multi-labels through semantic similarity of multiple video-query pairs. Huang \etal~\cite{huang2022videoal} introduce  elastic moment bounding to accommodate flexible and adaptive moment boundaries. The goal is to model a universally interpretable video-text correlation with tolerance to underlying uncertainties in pre-fixed annotations. Despite many efforts made, the uncertainty issue remains far from being solved.


\begin{figure}[t]
    \centering
	\subfigure[Charades-STA]
	{
	    \label{fig:charades_action_stat}	
	    \includegraphics[trim={0cm 0cm 0cm 0cm},clip,width=0.85\linewidth]{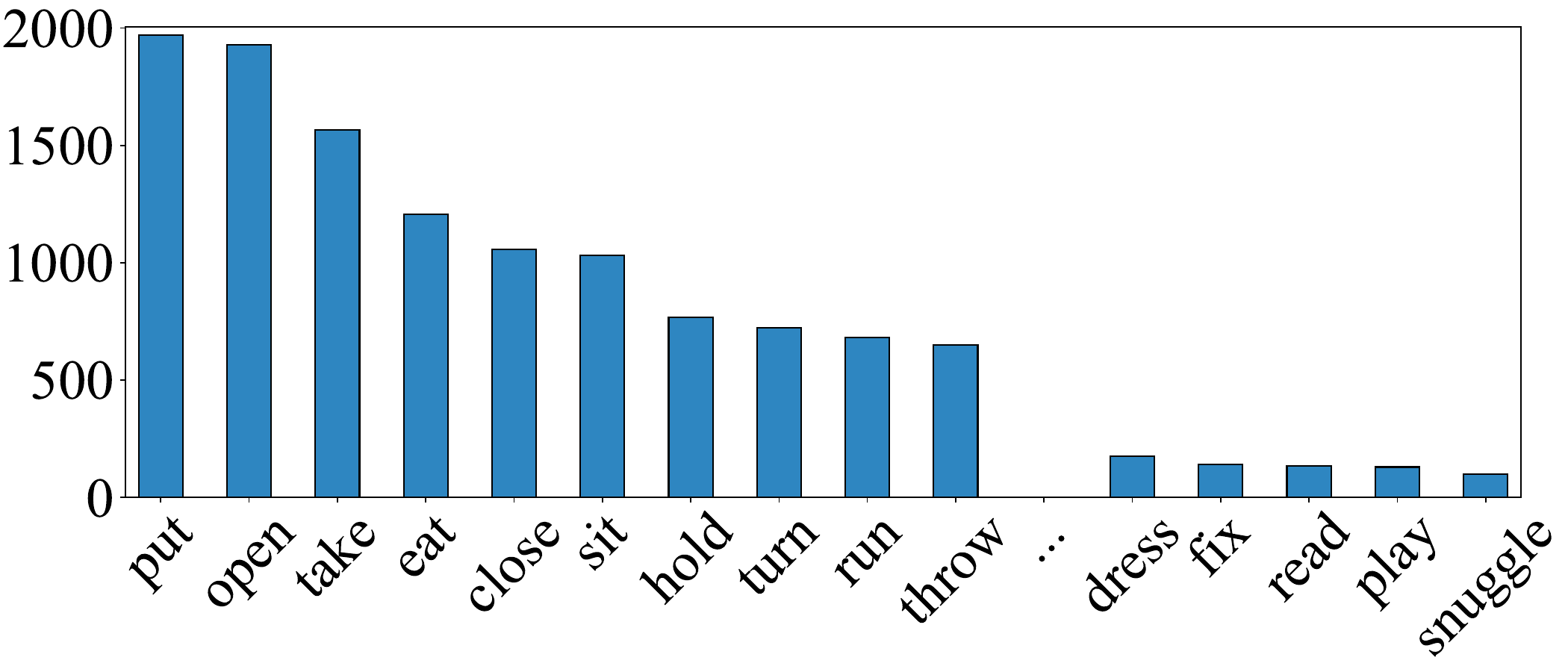}
	}
	\subfigure[ActivityNet Captions]
	{
	    \label{fig:activitynet_action_stat}	
	    \includegraphics[trim={0cm 0cm 0cm 0cm},clip, width=0.85\linewidth]{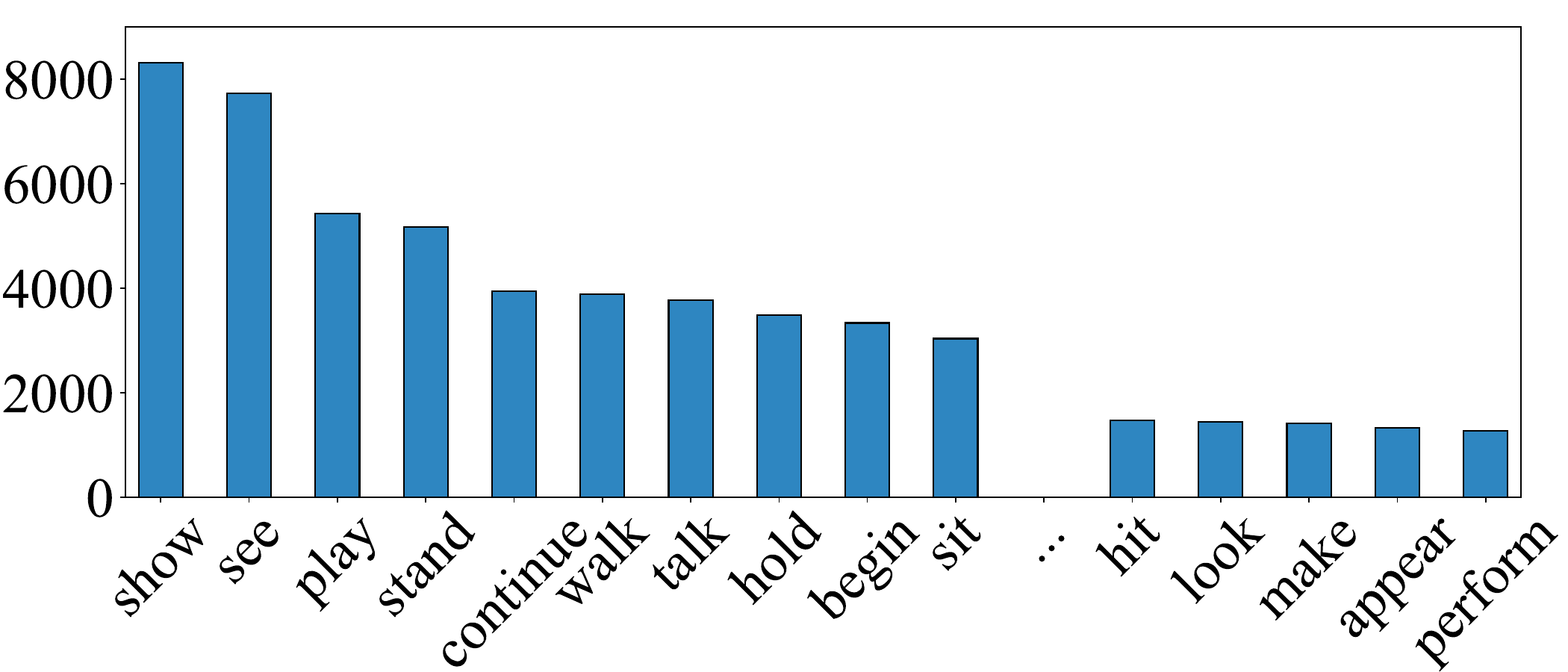}
	}
	\caption{The top-$30$ frequent actions in Charades-STA and ActivityNet Captions datasets.}
	\label{fig:action_stat}
\end{figure}

\begin{figure}[t]
    \centering
	\subfigure[Charades-STA]
	{
	    \label{fig:charades_sta_moment_dist}	
	    \includegraphics[trim={0cm 0cm 0cm 0cm},clip,width=0.7\linewidth]{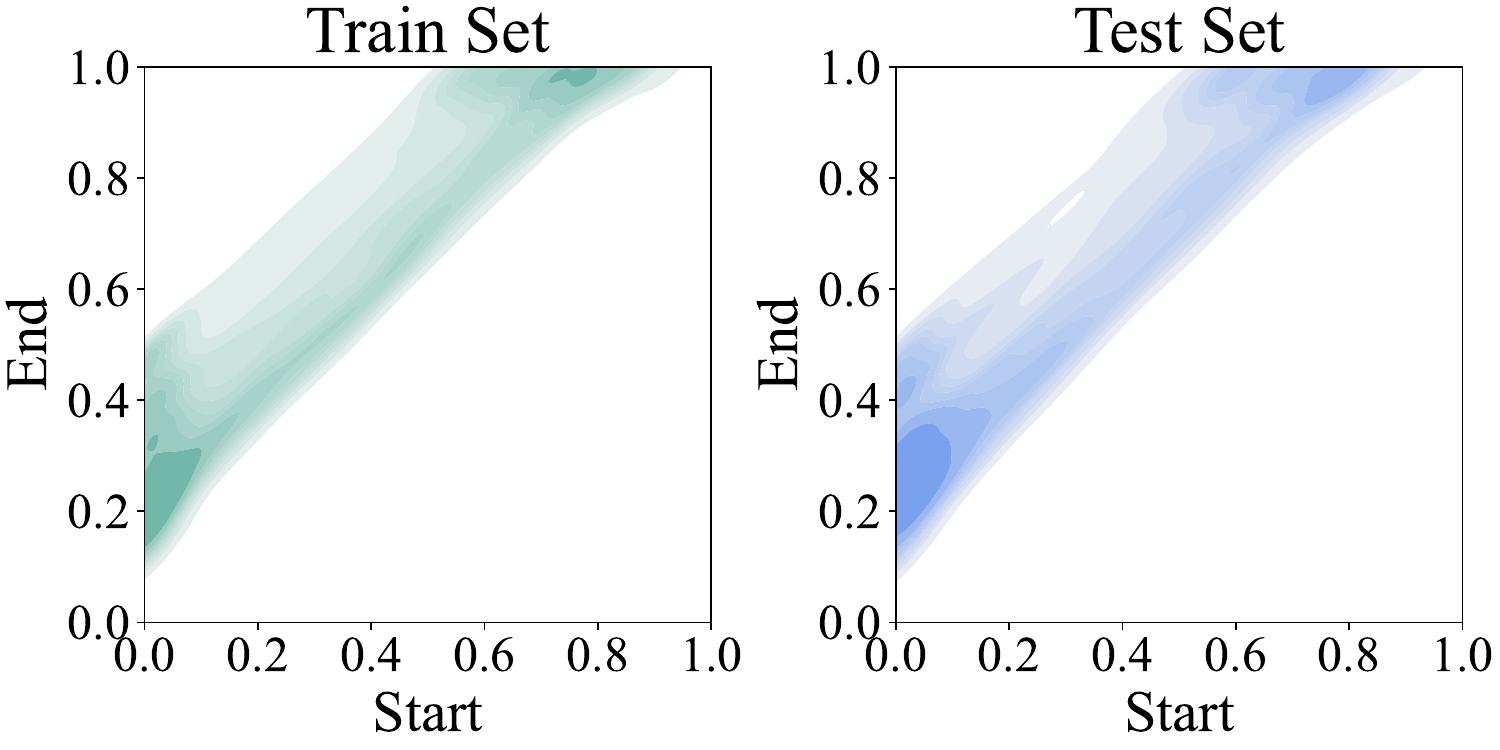}
	}
	\subfigure[ActivityNet Captions]
	{
	    \label{fig:activitynet_cap_moment_dist}	
	    \includegraphics[trim={0cm 0cm 0cm 0cm},clip, width=0.7\linewidth]{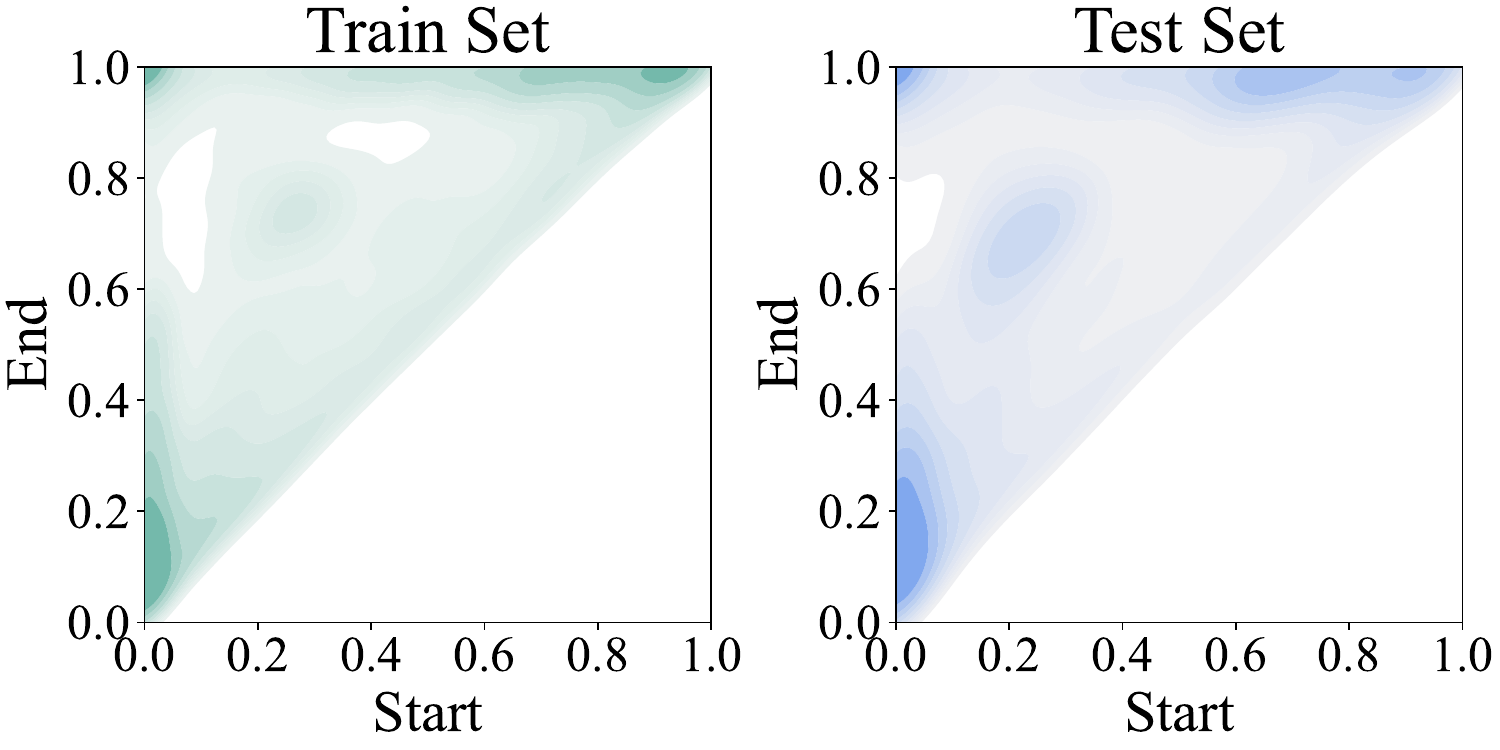}
	}
	\caption{An illustration of moment distributions for Charades-STA and ActivityNet Captions, where ``Start'' and ``End'' axes represent the normalized start and end timestamps, respectively. The deeper the color, the larger density (\ie more annotations) in the dataset.}
	\label{fig:org_data_dist}
\end{figure}

\smallskip \noindent \textbf{Data Bias.} Otani \etal~\cite{otani2020challengesmr} and Yuan \etal~\cite{yuan2021closer} conduct analysis on Charades-STA and ActivityNet Captions. They count frequent actions in queries and visualize joint distributions of the start and end timestamps of the ground-truth moment. As shown in Fig.~\ref{fig:action_stat}, a few frequent action verbs cover most of the actions in the dataset, \ie a long-tail distribution exists in both datasets. A large number of queries describe some common events, while only a few queries cover the remaining actions.  Fig.~\ref{fig:org_data_dist} shows that moment distributions are identical in train and test sets with a distinct distributional bias for both datasets. Because of the distributional bias, a TSGV model could make a good guess of the target moment, even without taking into consideration the input video and query~\cite{otani2020challengesmr}. For instance, Otani \etal~\cite{otani2020challengesmr} modify 2D-TAN~\cite{zhang2020learning} to build a Blind-TAN model by removing the video feature extractor and replacing the map of visual features with a learnable map in the same shape. By training Blind-TAN solely with query sentences, the learnable map may acquire some ideas on when certain actions are likely to happen. Experiments on benchmark datasets show that, without accessing the video content, Blind-TAN  achieves comparable performance with  state-of-the-art methods. This result demonstrates the severe distributional bias in existing TSGV benchmarks.

\begin{figure}[t]
    \centering
    \includegraphics[trim={0cm 0cm 0cm 0cm},clip,width=0.9\linewidth]{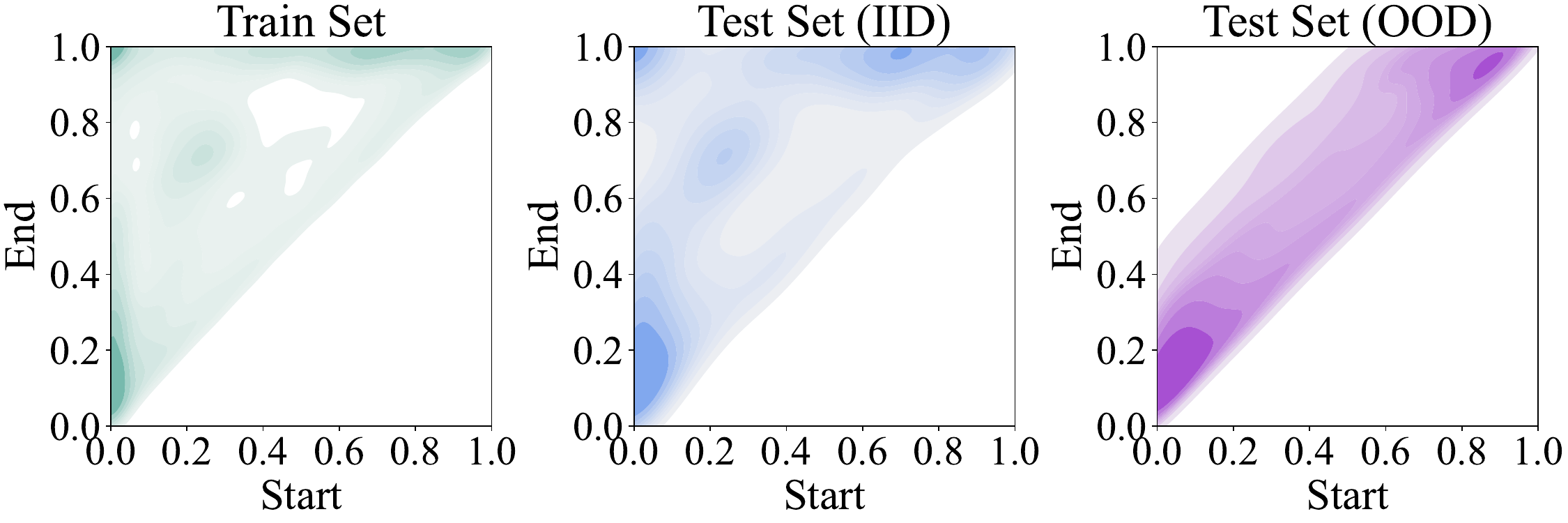}
    \caption{Illustration of moment distributions of ActivityNet-CD dataset.}
	\label{fig:activitynet_cd_dist}
\end{figure}

To investigate the effects of distributional bias among existing TSGV methods, Yuan \etal~\cite{yuan2021closer} further re-organize the two benchmark datasets and develop Charades-CD and ActivityNet-CD datasets. Each dataset contains two test sets, \ie the independent-and-identical distribution (iid) test set, and the out-of-distribution (ood) test set (see Fig.~\ref{fig:activitynet_cd_dist}). Then, Yuan \etal~\cite{yuan2021closer} collects a set of SOTA TSGV baselines and evaluates them on the reorganized benchmark datasets. Results show that baselines generally achieve impressive performance on the iid test set, but fail to generalize to the ood test set. It is worth noting that weakly-supervised methods are naturally immune to  distributional bias since they do not require annotated samples for training.

Recently, several solutions are proposed to alleviate the distributional bias. Yang \etal~\cite{yang2021deconfounded} develop a deconfounded cross-modal matching method to remove distributional bias by leveraging the structured casual mechanism~\cite{pearl2016causal}. Luo \etal~\cite{luo2021self} devise a self-supervised method to solve TSGV with pseudo label generation. Zhang \etal~\cite{zhang2021towards} disentangles bias from  the TSGV model by adjusting the losses to compensate for biases dynamically. Liu \etal~\cite{liu2022reducingtv} propose a Debiasing-TSG model to filter and remove the negative biases in both vision and language modalities via feature distillation and contrastive sample generation. Hao \etal~\cite{hao2022cansv} propose to use shuffled videos to address distributional bias without losing grounding accuracy. Specifically, they introduce two auxiliary tasks, \ie cross-modal matching and temporal order discrimination, to promote the grounding model training. The cross-modal matching task leverages the content consistency between shuffled and original videos, to force the grounding model to mine visual contents to semantically match queries. The temporal order discrimination task leverages the difference in temporal order to strengthen the understanding of long-term temporal contexts. Although solutions are developed to address moment distributional bias \eg debias strategies and dataset reorganization, it remains unclear if the current benchmarks provide the right setup to evaluate TSGV methods. Meanwhile, the long tail distribution of action verbs in queries has not been well explored. 

Recently, because of the inevitable limitations of current benchmark datasets, Soldan \etal~\cite{soldan2021mad} present the MAD dataset. MAD comprises long-form videos, highly descriptive sentences, and a large diversity in vocabulary. Most importantly, the timestamps of moments are uniformly distributed in the video. Lei \etal~\cite{lei2021qvhighlights} also develop a new benchmark dataset termed QVHighlights to avert data bias of existing TSGV datasets.


\subsection{Future Directions}
\label{ssec:future_direction}

\subsubsection{Effective Feature Extractor(s)}
\label{sssec:pre_training}
Feature quality directly affects TSGV performance. Illustrated in the common pipeline (Fig.~\ref{fig:video_inputs}), the existing solutions mainly extract visual and textual features independently using the corresponding pre-trained visual (\eg C3D~\cite{tran2015learning} and I3D~\cite{carreira2017quo}) and textual extractors (\eg GloVe~\cite{pennington2014glove}, BERT~\cite{devlin2019bert} and RoBERTa~\cite{liu2019roberta}). Thus, there is a large gap between the extracted visual and textual features in different feature spaces. Although TSGV methods attempt to project them into the same feature space, the natural gap between them is hard to be eliminated. There may also be differences between TSGV datasets and the datasets used for pre-training the feature extractors, which leads to information loss or inaccurate representations.

Recently, Zhang \etal~\cite{zhang2021multi} develop a single-stream feature extraction framework for TSGV, following BERT~\cite{devlin2019bert}. Visual and textual features are concatenated and jointly encoded with stacked transformer blocks. Similarly, LocFormer~\cite{rodriguezOpazo2021locformer} also concatenates  BERT-based query features and I3D-based video features, and feed them into a transformer-based localization module for video grounding. However, the visual and textual features remain separately generated by different pre-trained extractors. Xu \etal~\cite{xu2021boundary} propose a pre-training strategy for TSGV by constructing a large-scale synthesized dataset with TSGV annotations. Inspired by ViT~\cite{dosovitskiy2021an}, Cao \etal~\cite{cao2021pursuit} develop a video cubic embedding module to extract 3D visual tokens and learn video content from scratch without reliance on pre-trained visual feature extractors. Although they adopt GloVe~\cite{pennington2014glove} embeddings for queries, the issues of feature gap are not well alleviated. Xu \etal~\cite{xu2022contrastivela} indicate that, in TSGV,  the video encoder is usually fixed during fine-tuning. Thus, it cannot learn temporal boundaries and unseen classes, causing a domain gap with respect to the downstream task. To eliminate this issue, they propose a post-pre-training approach without freezing the video encoder and introduce a masked contrastive learning loss to capture visio-linguistic relations between activities, background video clips, and language queries. Inspired by DETR~\cite{carion2020end}, Liu \etal~\cite{liu2022umt} further design a unified multi-modal transformer for jointly optimizing moment retrieval and highlight detection, as well as mitigating the gap between visual and textual features. In general, despite many attempts like utilizing more powerful feature extractors (ViT~\cite{dosovitskiy2021an}, BERT~\cite{devlin2019bert}, etc.) or designing unified frameworks to better align video and text features, the semantic discrepancy between visual and textual features remains a key challenge.

With the recent success of  pre-training techniques for natural language processing~\cite{devlin2019bert,liu2019roberta} and image-linguistic~\cite{Su2020VLBERT,Radford2021LearningTV} tasks,  video-linguistic pre-training (VLP) is being developed to improve video-text related downstream tasks. For instance, ActBERT~\cite{zhu2020actbert} proposes to encode joint video-text representations from unlabeled data via self-supervised learning, which leverages global action information to catalyze mutual interactions between text and local regional objects. By designing a set of self-supervised objectives, \eg masked language modeling, masked action/object classification, and cross-modal matching, ActBERT not only learns the fine-grained video-text representations but also enforces the video and text representations being encoded in the same feature space. Similarly, other video-based vision-language pre-training methods (\eg BVET~\cite{wang2021bevt}, ClipBERT~\cite{lei2021less}, VideoCLIP~\cite{xu2021videoclip}, VLM~\cite{Xu2021VLMTV}, MERIOT~\cite{Zellers2021MERLOTMN}, etc.) also aim to learn better joint video-text representations by designing more sophisticated objectives or training strategies. After pre-training, these VLP models could be applied to various downstream video-and-language tasks, including text-video clip retrieval, video captioning, video question answering, moment localization, etc.

From the perspective of TSGV,  VLP models are good choices for feature extractors. Compared to the traditional visual and textual feature extractors, VLP models well mitigate the gaps between visual and textual features. More importantly, the features extracted by VLP models contain more cross-modal knowledge, which may further boost TSGV performance. However, applying off-the-shelf VLP models for TSGV tasks is yet well explored. On the other hand, the success of VLP also encourages  specific pre-training strategies for the TSGV task. Cao \etal~\cite{cao2022locvtp} indicate that almost all existing video-text pre-training methods are limited to retrieval-based downstream tasks. Their transfer potentials to localization-based tasks are underexplored. Based on the observation that current VLP methods are incompatible with localization tasks, they propose a localization-oriented video-text pre-training framework. Zeng \etal~\cite{zeng2022point} further design a point prompt tuning paradigm for the TSGV task. In general, localization-oriented VTP is a promising direction for TSGV. 

\subsubsection{TSGV with Multiple Answers}
\label{sssec:tsgv_multi_ans}
Existing TSGV benchmark datasets generally hold an implicit assumption that, for a query, there is only one ground truth moment exists in the input video. In reality, a query may describe multiple disjoint moments in a video. Based on the observation, Lei \etal~\cite{lei2021qvhighlights} present a unified benchmark dataset named QVHighlights for both TSGV and highlight detection tasks. In the dataset, each video is annotated with a human-written free-form language query, relevant moments in the video regarding the query, and five-point scale saliency scores for all query-relevant clips. This comprehensive annotation enables researchers to develop and evaluate systems that detect relevant moments as well as salient highlights for diverse and flexible user queries. Specifically, for the TSGV setting, for each language query, QVHighlights provides one or more disjoint moments in the video, enabling a more realistic evaluation of TSGV methods.

To solve TSGV with multiple answers,  a method is expected to deeply understand video contents and their semantic relationship to the language query. Inspired by DETR~\cite{carion2020end}, Lei \etal~\cite{lei2021qvhighlights} further propose Moment-DETR, an end-to-end transformer encoder-decoder architecture that views TSGV as a direct set prediction problem. The model takes the extracted video and query representations as inputs and predicts moment coordinates and saliency scores end-to-end, without any human prior, such as proposal generation or non-maximum suppression. From the query perspective, Bao \etal~\cite{bao2021dense} convert TSGV to dense events grounding task, which aims to jointly localize the multiple moments described in a paragraph \ie multiple queries. Then they devise a DepNet to adaptively aggregate the temporal and semantic information of dense events into a compact set, and selectively propagate the aggregated information to every single event with soft attention. Based on DepNet, Shi \etal~\cite{shi2021endtoend} further present an end-to-end parallel decoding paradigm by re-purposing a transformer-alike architecture from the perspective of TSGV, as language-conditioned regression. Jiang \etal~\cite{jiang2022gtlr} build a graph-based transformer with language reconstruction to jointly extract temporal moments and reconstruct queries with extracted moments for explainability. Liu \etal~\cite{liu2022umt} also apply a DETR-style framework and introduce pre-training with ASR captions for this task.

In general, jointly localizing multiple moments helps to alleviate the bias in TSGV. Joint localization may also help to improve overall accuracy as moments are semantically correlated and temporally coordinated by their order in a video. Multiple answers for a query is a novel task extended from the standard TSGV task, and this setting is more realistic and less biased.

\subsubsection{Spatio-Temporal Sentence Grounding in Videos}
\label{sssec:stsgv}
Spatio-temporal sentence grounding in videos (STSGV) is another extension of TSGV. The goal of TSGV is to extract a temporal moment, \ie detecting the start and end timestamps in a video for a language query. One step further, given a query, STSGV aims to sequentially localize the referring instances in a sequence of continuous frames in the video \ie a spatio-temporal tube.  Compared to TSGV, STSGV is more complicated since the task requires localizing not only the event's temporal boundaries but also the bounding boxes among frames in the video segment. Recently, a series of work~\cite{huang2018finding,shi2019not,chen2019weakly,chen2020activity,sadhu2020video,zhang2020does,zhang2020object,shen2020hierarchical,feng2021decoupled,tang2021human,tan2021look,su2021stvgbert,cao2022correspondence,li2022crossmodal,li2022endtoend,xiong2022gaussian,lin2022STVGFormer,yang2022tubedetr} has been proposed for this problem. A number of datasets are made available, including VID-sentence~\cite{chen2019weakly} which is based on ImageNet video object detection, ActivityNet-SRL~\cite{sadhu2020video} from existing caption and grounding datasets, VidSTG~\cite{zhang2020does}, and HC-STVG~\cite{tang2021human}. For instance, Lin \etal~\cite{lin2022STVGFormer} propose a STVGFormer with static-dynamic cross-modal understanding. In STVGFormer, a static branch learns to predict the spatial location according to static cues like human appearance; a dynamic branch learns to predict temporal boundaries according to dynamic cues like human action. Then a static-dynamic interaction block is designed to enable the two branches to query useful and complementary information from the opposite branch. Yang \etal~\cite{yang2022tubedetr} devise TubeDETR, inspired by the success of DETR-based architecture for text-conditioned object detection. TubeDETR jointly encodes text, appearance, and motion information, with the aim to predict the moment's temporal and spatial boundaries simultaneously.

Despite the availability of multiple datasets and methods, annotating spatio-temporal tubes in the video is more difficult and labor-intensive, compared to TSGV annotation. Thus, many methods~\cite{huang2018finding,shi2019not,chen2019weakly,chen2020activity,tan2021look} seek to solve STSGV under the weakly-supervised setting, which does not require fully annotated datasets. For instance, Chen \etal~\cite{chen2019weakly} utilize a pre-trained instance generator to produce spatio-temporal instances from video. They then adopt an attentive interactor to exploit the complicated relationships between instances and the sentence. The overall model is optimized through a multiple-instance learning strategy. Tan \etal~\cite{tan2021look} further design a self-supervised grounding mechanism with a contrastive multi-layer multi-modal attention module to locate spatial-temporal tubes in the video. Although some promising results are obtained, STSGV remains in its early stage.

\subsubsection{Multi-modal Temporal Grounding in Video}
\label{sssec:mtgv}
TSGV is a form of temporal video grounding using text as query, \ie language modality. Other modalities, such as audio, image, and short video clip, may also serve as queries for temporal video grounding. In fact, temporal video grounding with other modalities has also been studied in recent years, such as audio-visual event localization, image-to-video retrieval, and video re-localization. To be specific,  audio-visual event localization (AVEL)~\cite{tian2018audio,wu2019dual,xu2020crossmodal,xuan2020cross,duan2021audio,xuan2021discriminative,xue2021audio,xia2022videoguided} is to retrieve the synchronized video segment for a given audio from an untrimmed video. The task of image-to-video retrieval (IVR)~\cite{garcia2018asymmetric,zhang2019localizing,xu2020proposal,liu2021activity} is to localize video segments that contain similar activity as in the query image. Similarly, given a query video and a reference video, video re-localization (VRL)~\cite{feng2018video,feng2019spatio,huang2020weakly,jiang2021learning} aims to retrieve a segment in the reference video that semantically corresponds to the query video. Conceptually, the query is in the form of audio in AVEL, appearance vision (image) in IVR, and motion vision (video clip) in VRL, respectively. Despite the different query modalities used in AVEL, IVR, and VRL, their overall modeling process is similar to TSGV from the perspective of feature space. For instance, for AVEL, Xuan \etal~\cite{xuan2020cross} adopt a VGG network pre-trained on AudioSet~\cite{gemmeke2017audioset} to extract feature sequence for audio, and use ResNet to extract feature sequence for video. Then an attentive-based cross-modal network is applied to learn the multimodal interactions between video and audio to perform event localization. For IVR, Liu \etal~\cite{liu2021activity} extract the image query feature through a pre-trained VGG network, and the video feature sequence via R-C3D~\cite{xu2017r} model. Similarly, for VRL, Feng \etal~\cite{feng2018video} utilize the pre-trained C3D~\cite{tran2015learning} to extract feature sequences of both query and reference videos. To sum up, after feature extraction for query modality, the input formats of TSGV, AVEL, IVR, and VRL are almost the same. In principle, solutions to these tasks should be similar to each other, except for some subtle differences. However, compared to TSGV, temporal video grounding with such modalities is not widely studied.

Different query modalities could provide extra guidance to boost the performance for moment localization in videos. For instance, audio signals (\eg dog bark, noise in kitchen) offer auxiliary clues~\cite{chen2020learning,afouras2018deep} for precise localization. Audio transcription from video (if exists) using ASR~\cite{koenecke2020racial,huang2020deep} could provide relevant information for  cross-modal alignment between the video and query. For instance, Chen \etal~\cite{chen2020learning} introduce audio as an additional feature to TSGV and achieve better performance on several benchmark datasets. From the perspective of the query, different modalities of the query (\eg audio, sentence, and image) that describe the same event can be used for cross-validation of the retrieved results. Although TSGV, AVEL, IVR, and VRL accept different query modalities as a whole, there is a lack of a unified framework, which is suitable for all settings.

\subsubsection{Video Corpus Moment Retrieval}
\label{sssec:vcmr}
Video corpus moment retrieval (VCMR) extends video sources from a single video in TSGV to a large collection of videos. That is, VCMR aims to retrieve a matching moment to a query from a collection of untrimmed and unsegmented videos, \ie a video corpus. VCMR poses challenges to efficiently identify the relevant videos and to localize the relevant moments in the identified videos. Escorcia \etal~\cite{escorcia2019temporal} first extend TSGV to VCMR by modifying the existing TSGV benchmark datasets (\ie DiDeMo, Charades-STA, and ActivityNet Captions) to fit the VCMR setting. Then, they devise a clip-query alignment model, which learns to align the features of a natural language query to a sequence of short video clips that compose a candidate moment in a video. Lei \etal~\cite{lei2020tvr} construct the TVR dataset, where the videos come with associated textual subtitles and each query is associated with a tight temporal window in the corresponding video. This dataset is specifically designed for both TSGV and VCMR tasks, where it contains more than $100k$ queries collected on $21.8k$ videos from $6$ TV shows of diverse genres. Based on TVR, Lei \etal~\cite{lei2021mtvr} further extend it to a multilingual version named mTVR, which contains both English and Chinese queries. One of the purposes of mTVR is to investigate the generalization ability of VCMR models from one language to another language.

A number of methods~\cite{li2020hero,zhang2020hierarchical,zhang2021video,maeoki2021video,paul2021textbased,hou2021conquer,gao2021coarsetf,yoon2022cascaded,liu2022crosslingual,kim2022semantic,sun2022vsrnet} has been developed for VCMR. Li \etal~\cite{li2020hero} design a hierarchical transformer-based model for video-language omni-representation
learning and fine-tuning on TVR dataset. Zhang \etal~\cite{zhang2020hierarchical} develop a hierarchical multi-modal encoder to learn multimodal interactions at both coarse- and fine-grained granularities. Zhang \etal~\cite{zhang2021video} introduce contrastive learning to replace the time-consuming multimodal interaction strategy in VCMR to achieve a balance between efficiency and retrieval accuracy. Hou \etal~\cite{hou2021conquer} develop a two-step multimodal fusion for precise and efficient moment retrieval. Paul \etal~\cite{paul2021textbased} propose a hierarchical moment alignment network to effectively learn a joint embedding space to align the corresponding video moments and sentences. Liu \etal~\cite{liu2022crosslingual} further explore to solve the multilingual VCMR problem by devising a cross-lingual cross-modal consolidation strategy.

In general, VCMR contains two sub-tasks, \ie video retrieval, and moment localization. If a TSGV model is directly adapted, the query needs to interact with every video in the corpus, which is infeasible. However, VCMR is closer to practical scenarios as videos are ubiquitous. Although existing solutions to VCMR achieve consecutive improvements, the performance of VCMR is still inferior for real-world applications.

\section{Conclusion}
\label{sec:conclusion}
Many techniques are available to learn dense representations of various types of data \eg text, video, and audio. Through multimodal interaction,  cross-modal applications like TSGV become feasible. In this survey, we start with how to extract features from text and video,  then focus on the interaction between the two types of features for TSGV. Although TSGV has a short history, we have seen the trend of development from sliding window methods to proposal-based and proposal-free methods, then different views of the task with solutions from reinforcement learning and weakly-supervised learning. At the same time, we also see the challenges in this field; hence the results obtained on benchmark datasets may not necessarily reflect a model's performance in reality. Addressing these challenges would certainly bring improvements to current solutions. Furthermore, as a fundamental task, solutions to TSGV directly benefit many more  related applications like spatio-temporal sentence grounding in videos and video corpus moment retrieval. We hope this survey could serve as a good reference for researchers working on these interesting problems.

\bibliographystyle{IEEEtran}
\bibliography{main}

\appendices

\section{Supplementary Materials}
We present the following content as supplementary material: (i) the efficiency comparison among different method categories, and (ii) the comparison of TSGV and other video-language tasks.

\subsection{Efficiency Comparison}
\label{appd:ssec:efficiency}
In addition to the reported performance overview, we also provide an empirical  efficiency comparison among different categories of methods, such as training and test time. It is infeasible to compile and conduct efficiency evaluation for all TSGV models due to the computation resource and time constraints. Meanwhile, some methods do not make their codes publicly available. Hence, we select one or two representative models from each category for efficiency comparison. The purpose is to provide a glimpse of the efficiency of the different \textit{categories} of methods, rather than a detailed comparison among all methods.

To be specific, we choose CTRL~\cite{Gao2017TALLTA} to represent sliding window-based method (SW), LPNet~\cite{xiao2021natural} for proposal-generated method (PG), SCDM~\cite{yuan2019semantic} for the standard anchor-based method (AN), 2D-TAN~\cite{zhang2020learning} for 2D-Map anchor-based method (2D), ABLR~\cite{yuan2019to} for regression-based method (RG), VSLNet~\cite{zhang2020vslnet} for span-based method (SN), and RWM-RL~\cite{he2019read} to represent reinforcement learning-based method (RL).
Note that for the PG category, we choose  LPNet~\cite{xiao2021natural} instead of the two early PG methods QSPN~\cite{Xu2019MultilevelLA} and SAP~\cite{chen2019semantic}. QSPN is implemented in Caffe~\cite{jia2014caffeca}, which is difficult for us to compile under our preset environment. SAP does not release its source codes.

The ideal setting for model efficiency evaluation is to run all selected models on a benchmark dataset. However, we observe that the models process data in very different ways, such as different video/text features or different feature sequence lengths. In fact, there is no one common benchmark dataset on which all the selected models have been evaluated. For efficiency comparison, we choose to mock feature inputs to the models via random initialization \ie not relying on an existing benchmark dataset. Specifically, we set the batch size to be $16$, video sequence length $256$, video feature dimension $1,024$, word sequence length  $30$, and word feature dimension $300$. Then, the sizes of virtual video and text feature inputs for each batch are $16\times 256\times 4096$ and $16\times 30\times 300$, respectively. We follow the hyperparameters listed in the corresponding paper or code repository for other model settings. It is worth noting that the feature pre-load time is the same for all these methods due to the same mock features being utilized, so we do not count the data processing time in the evaluation. We run $200$ steps for model training and testing to calculate the execution time. The hyperparameters of the mocked inputs and train/test steps are summarized in Table~\ref{tab:efficiency_param}. For a fair comparison, all experiments are conducted on a single NVIDIA Tesla V100 GPU with 32GB memory.

\begin{table}
   \small
    \caption{\small The hyperparameters of feature input simulation for TSGV models, where $L_{\text{video}}$ is video sequence length, $d_{\text{video}}$ is video feature dimension, $L_{\text{text}}$ is word sequence length, and $d_{\text{text}}$ is  word feature dimension.}
	\centering
	\begin{tabular}{c | c | c | c | c | c}
		\specialrule{.1em}{.05em}{.05em}
		Batch Size & $L_{\text{video}}$ & $d_{\text{video}}$ & $L_{\text{text}}$ & $d_{\text{text}}$ & Train/Test Steps \\
        \hline
        $16$ & $256$ & $1,024$ & $30$ & $300$ & $200$ \\
        \specialrule{.1em}{.05em}{.05em}
	\end{tabular}
	\label{tab:efficiency_param}
\end{table}

The results of the efficiency comparison are summarized in Table~\ref{tab:efficiency_result}. Under the same environment and feature input settings, the SN category achieves the highest efficiency, SW and RL categories are the least efficient. PG, RG, 2D, and AN categories are in between these two extremes. For the PG category, LPNet utilizes VSLNet as the backbone to generate proposals and then selects a few top-ranked proposals for refinement to get the final prediction. Thus, LPNet is much faster than the SW-based method as well as the early PG-based solutions. For CTRL, SCDM, and RWM-RL, we observe their testing time is longer than the training time because these models only accept a single video as input each time by architecture design. In summary, the efficiency results of different method categories generally support our discussion on model efficiency in the main paper.

\begin{table}
    \small
    \caption{\small Efficiency comparison among the select TSGV models from different method categories, where $T_{\text{train}}$ and $T_{\text{test}}$ represent the total training and testing time in seconds.}
	\centering
	\begin{tabular}{@{}c | c | c | r | r }
		\specialrule{.1em}{.05em}{.05em}
		Category & Model & Backend & $T_{\text{train}}$ & $T_{\text{test}}$ \\
        \hline
        SW & CTRL~\cite{Gao2017TALLTA} & TensorFlow & 1732.51 & 3013.42 \\
        PG & LPNet~\cite{xiao2021natural} & TensorFlow & 113.45 & 42.98 \\
        AN & SCDM~\cite{yuan2019semantic} & TensorFlow & 61.99 & 279.01 \\
        2D & 2D-TAN~\cite{zhang2020learning} & PyTorch & 356.10 & 148.31 \\
        RG & ABLR~\cite{yuan2019to} & TensorFlow & 142.63 & 81.00 \\
        SN & VSLNet~\cite{zhang2020vslnet} & TensorFlow & 35.45 & 32.38 \\
        RL & RWM-RL~\cite{he2019read} & PyTorch & 2214.92 & 3296.86 \\
        \specialrule{.1em}{.05em}{.05em}
	\end{tabular}
	\label{tab:efficiency_result}
\end{table}

\subsection{Comparison between TSGV and Other VL Tasks}
\label{appd:ssec:tsgv_vs_others}

\begin{figure}[t]
    \centering
    \includegraphics[trim={0cm 0cm 0cm 0cm},clip,width=0.9\linewidth]{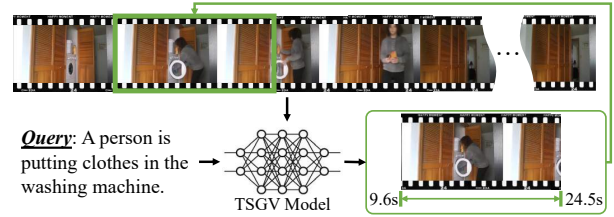}
    \caption{An illustration of temporal sentence grounding in videos (TSGV).}
	\label{fig:example_appd}
\end{figure}

In this section, we briefly discuss the relationships between TSGV and other vision-language (VL) tasks. Our goal is to better understand TSGV from different perspectives and the similarities and dissimilarities between TSGV and other VL tasks. We begin with the definition of TSGV: given an untrimmed video, TSGV is to retrieve a video segment, also known as a temporal moment, that semantically corresponds to a query in natural language \ie sentence. Fig.~\ref{fig:example_appd} provides an illustration.

\subsubsection{TSGV versus Visual Grounding}
\begin{figure}[t]
    \centering
    \includegraphics[trim={0cm 0cm 0cm 0cm},clip,width=0.9\linewidth]{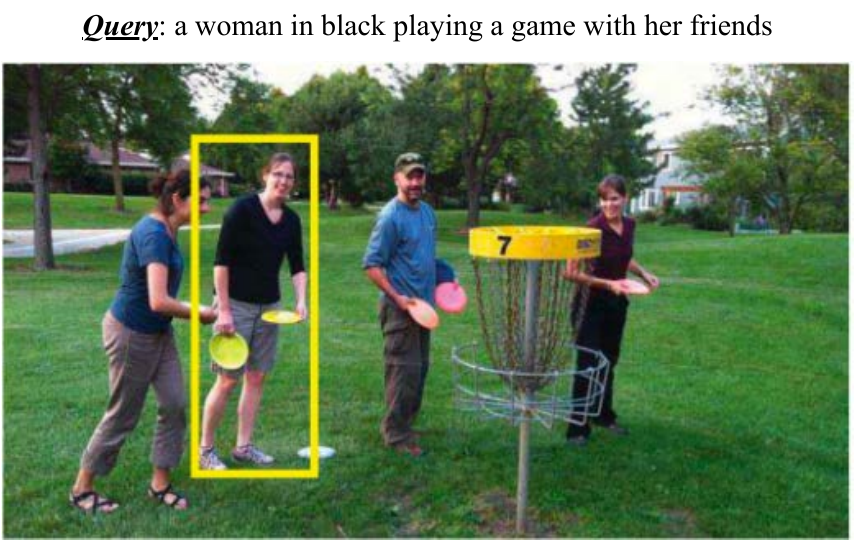}
    \caption{An illustration of visual grounding (VG). Given a language query and an image, VG aims to localize the referential object (yellow box) from the image, where the answer bounding box contains the object described by the query.}
	\label{fig:vg}
\end{figure}

Visual grounding (VG)~\cite{deng2018visual,Yang2019DynamicGA,deng2021transvg} aims to locate the most relevant object or region in an image, based on a natural language query. The query can be a phrase, a sentence, or even a multi-round dialogue. Illustrated in Fig.~\ref{fig:vg}, for a language query ``a woman in black playing a game with her friends'', VG needs to return the bounding box of the corresponding object from the image as the answer.

Based on the definitions of TSGV and VG, the only difference between the two tasks is the reference, \ie a video in TSGV and an image in VG. Solutions to VG~\cite{deng2018visual,Yang2019DynamicGA,deng2021transvg} usually apply a pre-trained CNN model to extract features from the whole image and cropped regions, resulting in a sequence of visual features. The language query is encoded by pre-trained word embeddings or language models. After that, various cross-modal learning frameworks are designed to encode the multimodal interactions between image and query and to predict the target bounding boxes. In general, the overall processes of TSGV and VG are the same. However, TSGV focuses on learning the relationships between language query and the visual contents in a dynamic temporal sequence \ie detecting the temporal boundaries. In contrast, VG focuses on learning the relationships between the query and the visual contents in a static spatial region \ie detecting spatial bounding boxes. Moreover, as discussed in Section~\ref{sssec:stsgv}, STSGV is a task to sequentially localize the referring instances from a sequence of continuous frames in a video. In this case, STSGV could be regarded as a combined task of TSGV and VG. In summary, the relationships among TSGV, VG, and STSGV suggest that some VL tasks are related, and solutions could be shared among them to some extent.

\subsubsection{TSGV versus Video Retrieval}
\begin{figure}[t]
    \centering
    \includegraphics[trim={0cm 0cm 0cm 0cm},clip,width=0.95\linewidth]{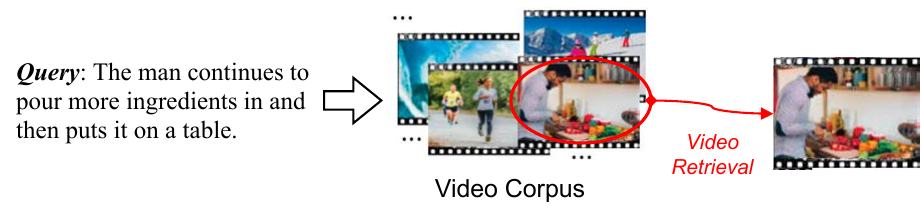}
    \caption{An illustration of text-based video retrieval (VR). Given a text query and a set of candidate videos, video retrieval (VR)  retrieves and ranks candidate videos by their relevance to the query.}
	\label{fig:vr}
\end{figure}

Given a query and a set of candidate videos, video retrieval (VR)~\cite{pan2016jointly,li2021sease,wang2022multiquery,wang2022align} is a task to retrieve and rank candidate videos by their relevance to the query. In general, queries for VR are not limited to text. Here we only consider the text-video retrieval scenario for its relevance to TSGV. As depicted in Fig.~\ref{fig:vr}, given a language query ``The man continues to pour more ingredients in and then puts it on a table.'', VR retrieves the videos whose content matches the query description. The general procedure of VR is to conduct cross-modal reasoning between text query and video candidates and project them into a joint embedding space. Within the joint space, VR aims to reduce the distance of the matching video-query pairs and increase the distance of non-matching pairs.

In our comparison, VR is to retrieve a video from a set of videos, and TSGV is to localize a temporal segment within a single video. To some extent, VR is coarse-grained retrieval while TSGV is fine-grained retrieval. Thus, VR focuses more on the overall semantic knowledge of the query as well as the overall information of candidate videos. TSGV is expected to understand the fine-grained query information and the representations and relationships of different events in a video. For a query, VR usually retrieves target videos from thousands of video candidates, while TSGV only considers the interaction between the query and a single video. Thus, the fine-grained cross-modal reasoning between language query and video in TSGV could be inefficient or even infeasible for VR. However, if TSGV adopts sliding window-based or proposal generation methods, then the video is first decomposed into a set of proposal candidates. Treating the proposal candidates as a set of short videos, TSGV and VR become similar since both tasks aim to rank the best matching candidates among multiple proposals/videos, for a given language query. In this case, solutions to VR may be  applicable to TSGV to some extent. Besides, VCMR (discussed in Section~\ref{sssec:vcmr}) is to retrieve a matching moment to a query from a collection of untrimmed and unsegmented videos; this task can be regarded as a combination of TSGV and VR.

\subsubsection{TSGV versus Video Question Answering}
\begin{figure}[t]
    \centering
    \includegraphics[trim={0cm 0cm 0cm 0cm},clip,width=0.95\linewidth]{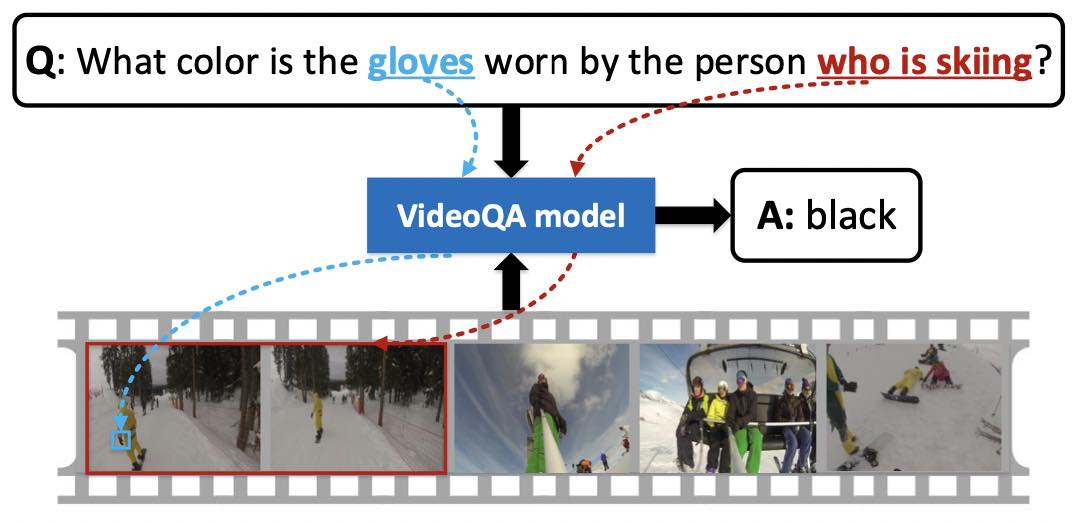}
    \caption{An example of video question answering (VideoQA) from Yu \etal~\cite{yu2019activitynetqa}. Given a text question, VideoQA needs to fully understand the fine-grained semantics of the question (\eg keywords) and to perform cross-modal reasoning on the visual contents (frames in the red border and objects in the blue box) to answer the question.}
	\label{fig:videoqa}
\end{figure}

Video question answering (VideoQA) ~\cite{jang2017tgifqa,lei2018tvqa,Gao2018MotionAppearanceCN,liang2019focalvisual,yu2019activitynetqa} is to answer a question in text form, based on the events/objects contained in an input video. As shown in Fig.~\ref{fig:videoqa}, given a question ``What color are the gloves worn by the person who is skiing?'', the VideoQA model needs to understand the key components of the question (\eg ``gloves'' and ``skiing''), and to interact with the video to ground the events and/or objects mentioned in the question. Then, the model predicts the answer based on the retrieved events/objects.

\begin{figure}[t]
    \centering
    \includegraphics[trim={0cm 0cm 0cm 0cm},clip,width=0.95\linewidth]{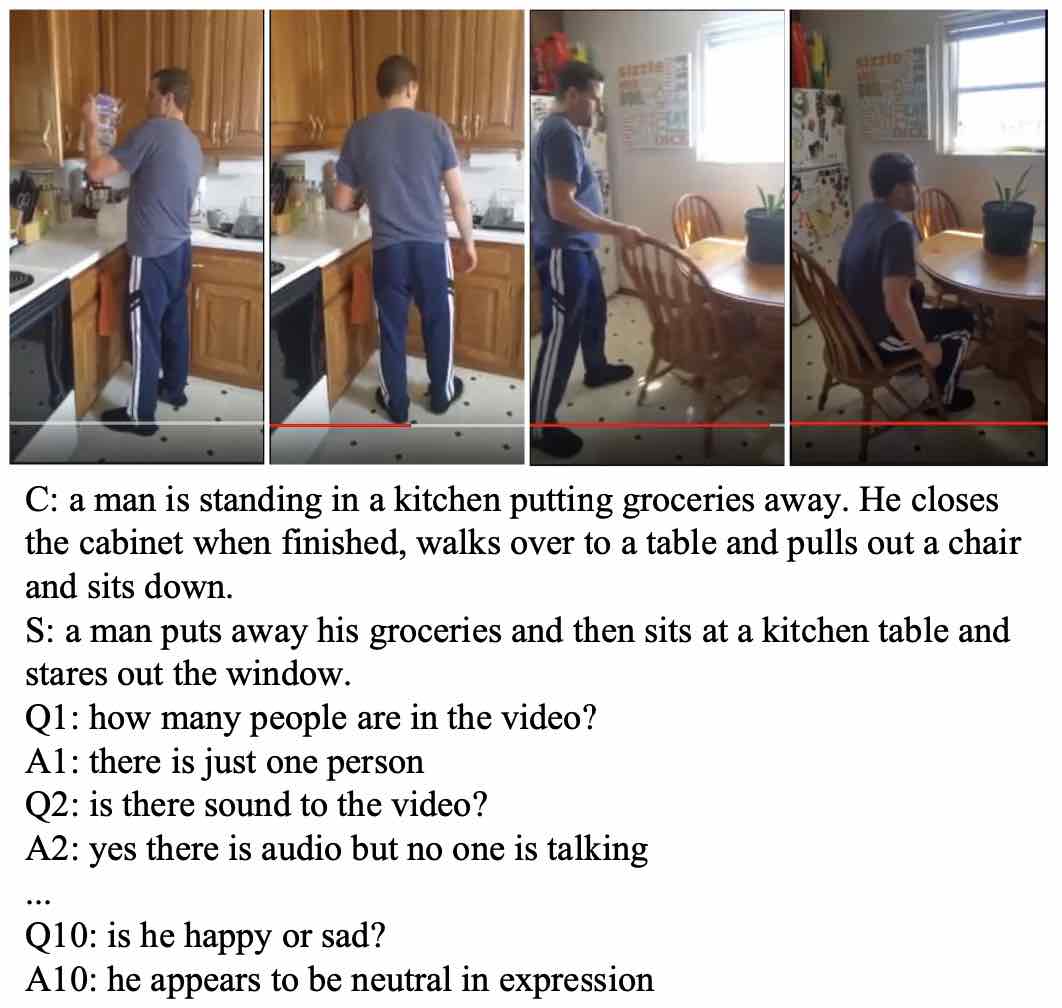}
    \caption{An example of video grounded dialogue (VideoDial) from Le \etal~\cite{le2019multimodal}, where $C$ denotes video caption, $S$ represents video summary, $Q_i$ denotes the $i$-th turn question and $A_i$ represents the $i$-th tun answer. The definition of VideoDial is that, given a video, a dialogue is conducted based on the visual and audio aspects of the given video.}
	\label{fig:videodial}
\end{figure}

VideoQA contains two reasoning steps.  The first is to localize the contents relevant to the given question from the video. The second is to infer the answer based on the grounded contents. Since  temporal grounding in  the video is an indispensable component, TSGV could serve as an intermediate step in VideoQA. As object grounding is also required for VideoQA in answer prediction, both visual grounding (VG) and STSGV can be applied here. In fact, several work~\cite{lei2020tvqaplus,kim2020modality,kim2021self} applies TSGV as an auxiliary component in VideoQA. For instance, Lei \etal~\cite{lei2020tvqaplus} propose a spatio-temporal answerer with grounded evidence. They design a video-text fusion module followed by a span predictor to localize the temporal boundaries of moments relevant to the answer. Similarly, Kim \etal~\cite{kim2020modality} deploy a moment proposal network to localize the required temporal moment of interest for question answering.

\subsubsection{TSGV versus Video Grounded Dialogue}
Video grounded dialogue (VideoDial)~\cite{pasunuru2018game,pasunuru2019dstc7,le2019multimodal,le2022vgnmn} is to conduct a multi-turn conversation, based on the visual and audio aspects of a given video. Similar to VideoQA, VideoDial also requires moment localization in the video as an intermediate step to support  answer generation. However, there are several differences between VideoQA and VideoDial. First, VideoQA is usually formulated as a multiple-choice problem, while VideoDial is a generation task. Second, VideoQA is a single-turn task while VideoDial consists of a multi-turn conversation. Thus, VideoDial is considered more challenging and it needs an in-depth understanding of the visual and/or audio contents. Meanwhile, VideoDial also requires continuous moment localization based on both the current utterance and conversation history. In general, TSGV is an indispensable component in VideoDial.

\end{document}